\documentclass{article} %
\usepackage{iclr2027_conference,times}

\usepackage{amsmath,amsfonts,bm}

\def\eqref#1{equation~\ref{#1}}

\def\1{\bm{1}}

\DeclareMathAlphabet{\mathsfit}{\encodingdefault}{\sfdefault}{m}{sl}
\SetMathAlphabet{\mathsfit}{bold}{\encodingdefault}{\sfdefault}{bx}{n}

\usepackage{enumitem}
\usepackage{booktabs}
\usepackage{mathtools}
\usepackage{multirow}
\usepackage{graphicx}
\usepackage{url}
\usepackage[table]{xcolor}
\usepackage{tabularx}
\usepackage{soul}
\usepackage{tcolorbox}
\usepackage{makecell}

\usepackage{hyperref}
\definecolor{darkblue}{rgb}{0, 0, 0.5}
\hypersetup{
    colorlinks=true,
    citecolor=darkblue,
    linkcolor=darkblue,
    urlcolor=darkblue
}

\usepackage{wrapfig}

\usepackage{tikz}
\usepackage{pgfplots}
\pgfplotsset{compat=1.18}
\usetikzlibrary{plotmarks}
\usepgfplotslibrary{groupplots}

\usepackage{arydshln}
\makeatletter
\def\adl@drawiv#1#2#3{%
        \hskip.5\tabcolsep
        \xleaders#3{#2.5\@tempdimb #1{1}#2.5\@tempdimb}%
                #2\z@ plus1fil minus1fil\relax
        \hskip.5\tabcolsep}
\newcommand{\cdashlinelr}[1]{%
  \noalign{\vskip\aboverulesep
           \global\let\@dashdrawstore\adl@draw
           \global\let\adl@draw\adl@drawiv}
  \cdashline{#1}
  \noalign{\global\let\adl@draw\@dashdrawstore
           \vskip\belowrulesep}}
\makeatother

\makeatletter
\newcommand{\localcaptiontype}[1]{\def\@captype{#1}}
\makeatother

\title{GSLM2: Multi-Stream Speech Representation Learning and Generative Spoken Language Modeling}

\author{%
Yosuke Higuchi$^{1,*, \dagger}$, Tom Fizycki$^{1,2,*, \dagger}$, Mahi Luthra$^{1,\dagger}$, Charles-\'Eric Saint-James$^{1, \dagger}$,\\
\textbf{Azeez Syed$^{1, \dagger}$, Angel Villar-Corrales$^{1}$, Mido Assran$^{3,\ddagger}$, Nicolas Ballas$^{3,\ddagger}$, Jiayi Shen$^{1}$,}\\
\textbf{Dongyan Lin$^{1}$, Phillip Rust$^{1}$, Surya Parimi$^{1}$, Vanessa Stark$^{1}$, Alexandre Allauzen$^{2}$,}\\
\textbf{Bal\'azs Kegl$^{4, \ddagger}$, Juan Pino$^{1}$, Emmanuel Dupoux$^{1,5}$}\\
$^1$Meta Superintelligence Labs, $^2$Universit\'e Paris Dauphine-PSL, $^3$AMI Labs, $^4$Doctolib, \\
$^5$Ecole des Hautes \'Etudes en Sciences Sociales \\
$^*$Equal contribution, $^\dagger$Core contributors, $^\ddagger$Work done while at Meta \\
\texttt{\{yosukeh,tomfizycki\}@meta.com}
}

\iclrfinalcopy %
\begin{document}

\maketitle

\begin{abstract}
Generative spoken language models (SLMs) offer a route to learning linguistic structure purely from speech,
yet they lag far behind text-based models in both capability and data scaling efficiency.
We identify two bottlenecks underlying this gap:
the limited linguistic quality of existing speech units
and the reliance on a single stream of discrete units to represent speech.
Multiple codebooks are widely used to encode acoustic features but rarely to represent speech semantics.
We present GSLM2, which learns and models complementary properties of speech at multiple levels of abstraction.
Its encoder, \textbf{SpidR 2.0}, uses self-supervised learning with differentiable prediction heads to induce high-quality discrete units at multiple levels.
A \textbf{multi-stream SLM} then jointly models these units using a shared temporal backbone, stream-specific predictors, and an autoregressive head that captures cross-stream dependencies.
Across training data sizes and model scales, up to 6M hours and 6.5B parameters,
GSLM2 consistently improves linguistic competence while scaling more efficiently with data.
As a result, it substantially narrows the gap between spoken and text-based LMs;
across read and conversational speech domains, we shrink the gap by about two orders of magnitude on sWuggy, sBLIMP, and tSC,
while also outperforming competitive speech-text systems on all three tasks.
We open-source the training code and model checkpoints at
\url{https://github.com/facebookresearch/gslm2}.
\end{abstract}

\section{Introduction}
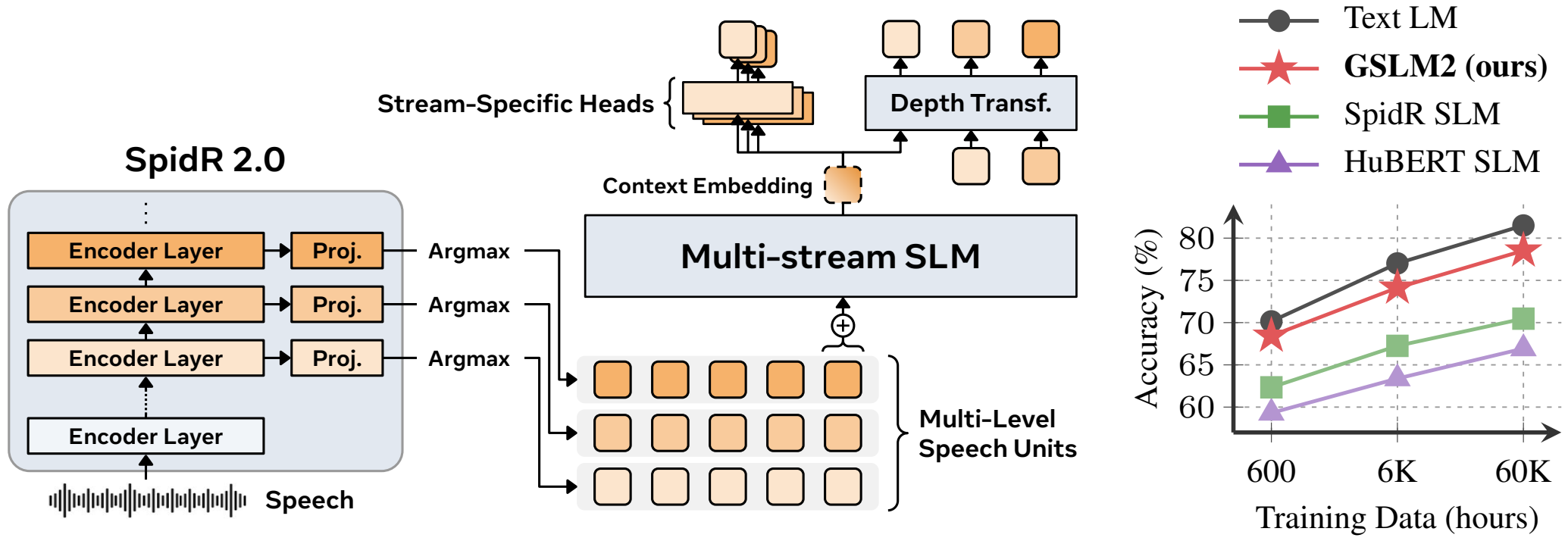
\begin{figure}[h]
    \vspace{-0.7cm}
    \centering

    \begin{minipage}[c]{0.70\columnwidth}
        \centering
        \includegraphics[width=0.96\linewidth]{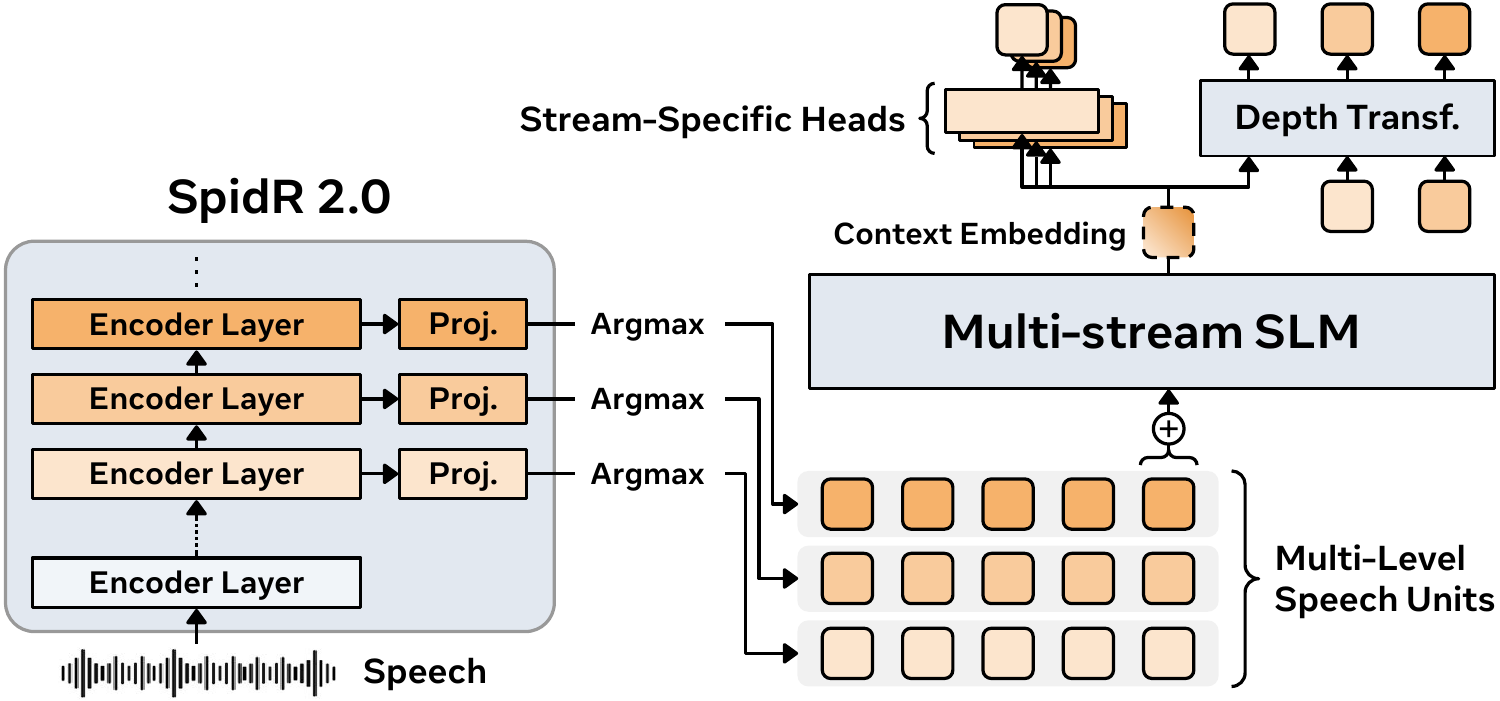}
    \end{minipage}
    \hfill
    \begin{minipage}[c]{0.29\columnwidth}
        \centering
        \resizebox{\linewidth}{!}{%
            \begin{tikzpicture}
    \definecolor{hubertpurple}{RGB}{148,103,189}
    \definecolor{spidrgreen}{RGB}{89,161,79}
    \definecolor{gslm2red}{RGB}{225,87,89}
    \definecolor{textgray}{RGB}{80,80,80}

    \pgfdeclareplotmark{filled star}{%
        \pgfpathmoveto{\pgfqpoint{0pt}{\pgfplotmarksize}}%
        \pgfpathlineto{\pgfqpoint{.224514\pgfplotmarksize}{.309017\pgfplotmarksize}}%
        \pgfpathlineto{\pgfqpoint{.951057\pgfplotmarksize}{.309017\pgfplotmarksize}}%
        \pgfpathlineto{\pgfqpoint{.363271\pgfplotmarksize}{-.118034\pgfplotmarksize}}%
        \pgfpathlineto{\pgfqpoint{.587785\pgfplotmarksize}{-.809017\pgfplotmarksize}}%
        \pgfpathlineto{\pgfqpoint{0pt}{-.381966\pgfplotmarksize}}%
        \pgfpathlineto{\pgfqpoint{-.587785\pgfplotmarksize}{-.809017\pgfplotmarksize}}%
        \pgfpathlineto{\pgfqpoint{-.363271\pgfplotmarksize}{-.118034\pgfplotmarksize}}%
        \pgfpathlineto{\pgfqpoint{-.951057\pgfplotmarksize}{.309017\pgfplotmarksize}}%
        \pgfpathlineto{\pgfqpoint{-.224514\pgfplotmarksize}{.309017\pgfplotmarksize}}%
        \pgfpathclose
        \pgfusepath{fill,stroke}%
    }
    
    \begin{axis}[
        width=\linewidth,
        height=3.3cm,
        xmode=log,
        log basis x=10,
        xmin=300,
        xmax=120000,
        ymin=57,
        ymax=84,
        xtick={600,6000,60000},
        ytick={60,65,70,75,80},
        xticklabels={600, 6K, 60K},
        xlabel={Training Data (hours)},
        ylabel={Accuracy (\%)},
        xticklabel style={font=\scriptsize},
        yticklabel style={
            font=\scriptsize,
            xshift=2pt
        },
        xlabel style={
            font=\scriptsize,
            yshift=2pt
        },
        ylabel style={
            font=\scriptsize,
            yshift=-4pt
        },
        tick align=outside,
        tick style={black!70},
        axis lines=left,
        axis line style={black!80, line width=0.9pt},
        ymajorgrids=true,
        xmajorgrids=true,
        major grid style={
            draw=black!40,
            dash pattern=on 1pt off 1.4pt,
            line width=0.3pt
        },
        line width=0.8pt,
        mark size=2.0pt,
        every axis plot/.append style={solid},
        legend entries={
            {Text LM},
            {\textbf{GSLM2 (ours)}},
            {SpidR SLM},
            {HuBERT SLM},
        },
        legend style={
            legend columns=1,
            legend style={
                font=\scriptsize,
                draw=none,
                fill=none,
                at={(0.5,1.02)},
                anchor=south,
                column sep=3pt,
            },
            legend cell align={left},
        },
        legend cell align={left},
        clip=false,
    ]

        \addlegendimage{
            color=textgray,
            mark=*,
            mark options={solid, fill=textgray}
        }

        \addlegendimage{
            color=gslm2red,
            mark=filled star,
            mark size=3pt,
            mark options={solid, fill=gslm2red, draw=gslm2red}
        }

        \addlegendimage{
            color=spidrgreen,
            mark=square*,
            mark options={solid, fill=spidrgreen}
        }

        \addlegendimage{
            color=hubertpurple,
            mark=triangle*,
            mark size=2.5pt,
            mark options={solid, fill=hubertpurple}
        }

        \addplot[
            color=hubertpurple!70,
            mark=triangle*,
            mark size=2.5pt,
            mark options={solid, fill=hubertpurple!70},
            forget plot
        ] coordinates {
            (600, 59.31)
            (6000, 63.38)
            (60000, 66.93)
        };

        \addplot[
            color=spidrgreen!70,
            mark=square*,
            mark options={solid, fill=spidrgreen!70},
            forget plot
        ] coordinates {
            (600, 62.34)
            (6000, 67.25)
            (60000, 70.47)
        };

        \addplot[
            color=textgray,
            mark=*,
            mark options={solid, fill=textgray},
            forget plot
        ] coordinates {
            (600, 70.15)
            (6000, 77.01)
            (60000, 81.50)
        };
        
        \addplot[
            color=gslm2red,
            mark=filled star,
            mark size=3pt,
            mark options={solid, fill=gslm2red, draw=gslm2red},
            forget plot
        ] coordinates {
            (600, 68.44)
            (6000, 74.13)
            (60000, 78.52)
        };
    
    \end{axis}
\end{tikzpicture}%
        }
    \end{minipage}
    \vspace{-0.4cm}
    \caption{\textbf{Overview of GSLM2, comprising SpidR 2.0 and multi-stream SLM.} \textit{Left}: SpidR 2.0 extracts discrete speech units from multiple encoder layers, which are jointly modeled by the multi-stream SLM. \textit{Right}: Scaling performance of SLMs and a text LM trained on varying amounts of Libri-Light data, measured by the average accuracy across thes Wuggy, sBLIMP, and tSC tasks.}
    \label{fig:gslm2}
\end{figure}

Generative spoken language modeling (GSLM)~\citep{lakhotia2021generative} seeks to learn
the structure of language directly from speech, typically by modeling
sequences of discrete speech units.
This textless approach aligns with embodied and cognitive developmental perspectives of intelligence,
which view language as building on representations of the world learned through
perception and action~\citep{lecun2022path, dupoux2026ai}.
A central challenge is to discover this structure in continuous, high-dimensional signals without symbolic supervision---a problem human learners demonstrably solve.
Models that learn from speech alone are not restricted to languages with a
written form and they acquire speech-native representations that retain both
linguistic content and speech-specific information such as phonetic variation.
These representations are particularly valuable for speech-first applications such as spoken dialogue modeling~\citep{nguyen2023generative}.
However, a major obstacle to textless modeling remains:
spoken language models (SLMs) can require up to four orders of magnitude more training data than text LMs to achieve comparable linguistic performance~\citep{cuervo2024scaling, poli2025spidr,maimon2025scaling},
as illustrated by HuBERT- and SpidR-based SLMs in Figure~\ref{fig:gslm2}.
Bridging this data-efficiency gap is the goal of this paper.

The gap may partly stem from the high sampling rate and low information density of speech representations:
the same content requires many more speech units than text tokens~\citep{cuervo2024scaling}.
Recent work has explored representations with lower sampling rates,
e.g., word-level~\citep{algayres2023generative}, syllable-like~\citep{cho2025sylber,baade2025syllablelm,visser2026zerosyl}, and latent patch-based units~\citep{lu2026latent}.
By shortening sequences, these approaches expand the effective context for next-unit prediction and improve linguistic abilities,
yet they still lag behind text LMs.
Despite their differing granularities,
they retain the text-inspired assumption that speech should be modeled as a single stream of discrete units,
seeking more efficient substitutes for text tokens.
Compression alone, however, does not resolve which distinctions these units should preserve.
Coarser units can improve efficiency and performance on sentence-level semantic tasks,
but may lose finer phonetic and lexical distinctions~\citep{kando2025exploring,visser2025spoken}.
These trade-offs suggest that the most useful level of detail may vary with the task and context,
motivating flexible access to different levels of abstraction.
Choosing a single unit type fixes the balance between compactness and detail,
potentially restricting available information as these demands change.

We argue that this reflects a broader representational bottleneck.
SLMs typically rely on speech representations or discrete units derived from a single layer of a self-supervised speech encoder,
despite evidence that different layers capture complementary phonetic and linguistic information~\citep{pasad2021layer,shen2023wave,pasad2024self}.
Consequently, a single representation is expected both to preserve fine-grained phonetic distinctions and to expose the high-level semantic regularities required for efficient language modeling. 
Satisfying these competing objectives within one sequence may unnecessarily burden the model and limit its ability to learn linguistic structure.

To this end, we introduce \textbf{GSLM2},
a framework for representing and modeling speech at multiple levels of abstraction.
Its encoder, \textbf{SpidR 2.0}, uses self-supervised student-teacher pretraining with layer-wise differentiable tokenization.
Rather than relying on a post-hoc $k$-means quantizer,
SpidR 2.0 learns multiple discrete unit spaces jointly during pretraining.
Shared prediction heads map student and teacher representations into common, stable unit spaces,
yielding high-quality 
discrete units at multiple encoder depths.
A \textbf{multi-stream SLM} then jointly models the unit streams extracted from these layers.
Its shared LM backbone captures temporal dependencies across sequence positions,
stream-specific heads model transitions within each stream,
and a generative head captures cross-stream dependencies.
By distributing speech properties across complementary streams,
GSLM2 preserves fine-grained distinctions while exposing higher-level regularities for efficient language modeling,
thereby narrowing the performance gap with text LMs (e.g., Figure~\ref{fig:gslm2}).
Empirically, GSLM2 shows improved scaling at model sizes up to 6.5B parameters with 6M hours of speech data
and outperforms competing systems, including speech-text models, across several key metrics.
These advances provide a foundation for speech-first, textless applications that model both linguistic and phonetic content directly from speech.
We make our models and code publicly available to support further research and application development.\footnote{\url{https://github.com/facebookresearch/gslm2}}

\section{GSLM2}
\label{sec:gslm2}

GSLM~\citep{lakhotia2021generative} aims to model the distribution of speech using discrete speech units.
A standard GSLM pipeline consists of three separately trained components:
a speech tokenizer, an SLM, and a unit vocoder.
The speech tokenizer maps a continuous speech signal into a sequence of discrete units
$Q = (q_1,\ldots,q_U)$,
typically by applying $k$-means clustering to hidden representations from a self-supervised speech encoder~\citep{hsu2021hubert}.
As in conventional language modeling,
the SLM autoregressively predicts the next speech unit conditioned on the preceding units.
Finally, the unit vocoder~\citep{polyak2021speech} converts speech units back into a waveform.
We focus here on the standard GSLM formulation;
Appendix~\ref{app:related_work} discusses related work
on speech tokenization, multi-level speech representations, and multi-stream sequence modeling.

The tokenizer is central to GSLM because it determines which speech properties are exposed to the SLM.
Standard tokenizers generally quantize representations from a single encoder layer,
restricting the resulting units to one level of abstraction.
However, self-supervised speech encoders distribute complementary phonetic and semantic information across layers~\citep{pasad2021layer,shen2023wave}.
Standard GSLMs therefore leave much of this multi-level information unexploited.

We propose GSLM2,
an extension of GSLM comprising two components,
as illustrated in Figure~\ref{fig:gslm2}.
First, we introduce \textbf{SpidR 2.0}, an encoder pretraining method that learns structured and stable code distributions throughout the encoder.
Second, we propose a \textbf{multi-stream SLM} that jointly models unit streams extracted from different encoder layers.
Together, these components enable GSLM2 to combine complementary speech information at multiple levels of representation.

\subsection{Encoder Pretraining: SpidR 2.0}
\label{ssec:spidr2}

SpidR 2.0 builds on the DinoSR and SpidR family ~\citep{liu2023dinosr, poli2025spidr}
within a student-teacher framework~\citep{grill2020bootstrap}.
We extend SpidR with differentiable tokenization and entropy regularization to learn stable, informative code distributions across multiple encoder layers.
The student and teacher are $L$-layer Transformer encoders.
The student receives a masked input sequence~\citep{hsu2021hubert},
while the teacher processes its unmasked counterpart.
At each selected layer, the student predicts targets derived from the corresponding teacher layer.
The teacher encoder parameters are updated as an exponential moving average (EMA) of the student parameters.

\textbf{Layer-wise differentiable tokenization.}
DinoSR and SpidR construct discrete targets by assigning teacher representations to the nearest entries in separate codebooks updated by EMA.
SpidR 2.0 instead uses \textit{shared prediction heads} to map teacher and student representations into a common code space at each selected layer,
inspired by~\citet{zhou2022image}.
These heads produce soft code assignments,
making tokenization fully differentiable during pretraining and eliminating the need to maintain separate online codebooks~\citep{liu2023dinosr}.

We attach a layer-specific prediction head to each encoder layer in a selected set
$\mathcal{K} \subseteq \{1,\ldots,L\}$ (as ``Proj.'' in Figure~\ref{fig:gslm2}).
For every $l \in \mathcal{K}$,
the same head is applied to the corresponding teacher and student representations and
emits a distribution over a vocabulary of $C$ codes.
During pretraining, these distributions serve as soft code assignments;
at inference, they are converted by argmax into discrete units for SLMs.
For each selected layer $l$, the prediction head
$\phi_{l}: \mathbb{R}^{d_{\mathsf{enc}}} \rightarrow \mathbb{R}^{C}$
consists of a layer normalization across channels, without scale or bias parameters, 
followed by a linear projection to $C$-dimensional logits. 
The projection matrix at layer~$l$ has ${C}$ rows,
with row $i$ given by $\mathbf{c}_{l,i}^{\top}$,
where $\mathbf{c}_{l,i} \in \mathbb{R}^{d_{\mathsf{enc}}}$ is the embedding of the $i$-th codeword.
We parametrize the rows of each projection layer $\mathbf{c}_{l,i}$ to have unit $L_2$ norm, hence pushing the codewords onto the unit sphere $\mathcal{U}(\mathbb{R}^{d_{\mathsf{enc}}})$.
The backward pass differentiates through this parametrization.
Let $\mathbf{h}_{l,t},\tilde{\mathbf{h}}_{l,t} \in \mathbb{R}^{d_{\mathsf{enc}}}$ denote the teacher and student representations at layer $l$ and frame $t$, respectively.
Using the shared head,
the teacher target and student prediction are computed as
\begin{equation}
    \mathbf{p}_{l,t} = \mathrm{Softmax} \Bigl(
        \phi_{l} \bigl(
            \mathrm{IN} (\mathbf{h}_{l,t})
        \bigr)
    \Bigr),\qquad
    \tilde{\mathbf{p}}_{l,t} = \mathrm{Softmax} \bigl(
        \phi_{l}(\tilde{\mathbf{h}}_{l,t})
    \bigr),
\label{eq:teacher-student}
\end{equation}
where $\mathrm{IN}(\cdot)$ indicates instance normalization on the time dimension,
and $\mathbf{p}_{l,t},\tilde{\mathbf{p}}_{l,t}\in\Delta^{C-1}$ are distributions over the shared code vocabulary.
Each selected layer therefore is associated with a differentiable tokenizer whose code space is optimized directly by distillation rather than through independently maintained online codebooks.
The same code space can be used directly for speech unit extraction,
eliminating the need for post-hoc, non-differentiable $k$-means on continuous representations~\citep{lakhotia2021generative},
while also improving performance (see Appendix~\ref{app:sssec:codes-vs-kmeans}).

The self-supervised objective is the cross-entropy between the detached teacher targets and student predictions,
averaged over the selected layers $\mathcal{K}$ and masked frames $\mathcal{M}$:
\begin{equation}
    \mathcal{L}_{\mathsf{ssl}} =
    -\frac{1}{|\mathcal{K}|} \sum_{l\in\mathcal{K}} 
    \frac{1}{|\mathcal{M}|}\sum_{t \in \mathcal{M}}
    \sum_{c=1}^{C}\mathrm{sg}(
        \mathbf{p}_{l,t}[c]
    ) \,
    \log \tilde{\mathbf{p}}_{l,t}[c],
    \label{eq:L_ssl}
\end{equation}
where $\mathrm{sg}(\cdot)$ denotes the stop-gradient operator.
Gradients are propagated only through the student branch,
while the teacher encoder parameters are updated using an EMA of the student parameters.

\textbf{Entropy regularization.}
To stabilize the shared prediction heads,
we regularize the teacher distributions to avoid both codebook collapse and overly diffuse assignments.
Following the ME-MAX principle for equipartition-constrained clustering~\citep{assran2022masked},
we combine a \textit{conditional-entropy} term $\mathbb{E}_t[H_c(\mathbf{p}_{l,t})]$ with a \textit{marginal-entropy} term $H_c(\mathbb{E}_t[\mathbf{p}_{l,t}])$ as
\begin{equation}
    \mathcal{L}_{\mathsf{reg}} =
    \frac{1}{|\mathcal{K}|}\sum_{l\in\mathcal{K}} \Bigl[
        \mu \, \mathbb{E}_t \bigl[
            H_c(\mathbf{p}_{l,t})
        \bigr] -
        \lambda \, H_c \bigl(
            \mathbb{E}_t[\mathbf{p}_{l,t}]
        \bigr)
    \Bigr],
    \label{eq:L_reg}
\end{equation}
where $\mu$ and $\lambda$ control the strengths of the two terms.
Minimizing the conditional entropy encourages confident code assignments for individual frames,
while maximizing the marginal entropy promotes broad code usage across frames.
Together, these objectives push for stable and informative discrete representations across multiple encoder layers,
whereas SpidR produces its highest-quality units primarily at a single layer (see Section~\ref{ssec:enc_eval}).
Applying this loss on the teacher distribution only ensures that the gradients from this regularization loss do not backpropagate to the encoder, but only constrain the unit space.
A study on these regularization choices is provided in Appendix \ref{app:sssec:regularization}.

\textbf{Training objective.}
The overall objective for encoder pretraining combines the self-supervised distillation loss from Eq.~(\ref{eq:L_ssl}) with the entropy regularization term in Eq.~(\ref{eq:L_reg}) as
$\mathcal{L}_{\mathsf{enc}} = \mathcal{L}_{\mathsf{ssl}} + \mathcal{L}_{\mathsf{reg}}$.

\subsection{Multi-Stream Spoken Language Modeling}
\label{ssec:ms_slm}
The multi-stream SLM jointly models speech units extracted from multiple encoder layers.
Let $\mathcal{J} \subseteq \mathcal{K}$ be the set of layers used as unit streams.
For each $l \in \mathcal{J}$,
we run the student encoder without input masking and extract the most likely code at each frame: $z_{l,t} = \arg\max_{c}\,\tilde{\mathbf{p}}_{l,t}[c]$.
This gives a raw unit sequence $Z_{l} = (z_{l,1},\ldots,z_{l,T})$ for each stream.
To shorten the sequence while preserving changes in any stream,
we apply \textit{cross-stream deduplication}:
for $t > 1$, frame $t$ is removed only if $z_{l,t} = z_{l,t-1}$ for all $l \in \mathcal{J}$, and is retained otherwise.
The retained frames thus form the union of the change points across all streams.
As a result, a stream contains repeated units at consecutive positions when another stream changes,
unlike independent single-stream deduplication.
The resulting multi-stream sequence is $\mathcal{Q} = (\mathbf{q}_1,\ldots,\mathbf{q}_U)$,
where $\mathbf{q}_u = (q_{l,u} \mid l\in\mathcal{J})$ is the tuple of stream units at the $u$-th retained frame,
and $U$ is the deduplicated sequence length shared by all streams.
The multi-stream SLM models the joint distribution over $\mathcal{Q}$ as
$p(\mathcal{Q}) = \prod_{u=1}^{U} p(\mathbf{q}_u \mid \mathbf{q}_{1},\ldots,\mathbf{q}_{u-1})$.

\textbf{Architecture.}
Each stream has its own embedding layer,
$\mathrm{Emb}_{l}: \{1,\ldots,C\} \rightarrow \mathbb{R}^{d_{\mathsf{slm}}}$,
which maps a discrete unit to the $d_{\mathsf{slm}}$-dimensional SLM embedding space.
At position $u$, the stream units are embedded independently and averaged:
$\mathbf{x}_u = \frac{1}{|\mathcal{J}|}\sum_{l\in\mathcal{J}} \mathrm{Emb}_{l}(q_{l,u}) \in \mathbb{R}^{d_{\mathsf{slm}}}$.
A shared decoder-only Transformer processes the preceding embeddings $(\mathbf{x}_{1},\ldots,\mathbf{x}_{u-1})$ and produces a context embedding $\mathbf{y}_u \in \mathbb{R}^{d_{\mathsf{slm}}}$.
This shared temporal backbone captures dependencies across sequence positions.

The context embedding is then fed to two types of output heads:
\textit{stream-specific heads} and a \textit{generative head},
as illustrated in Figure~\ref{fig:gslm2}.
For each stream $l \in \mathcal{J}$,
the \textit{stream-specific head} predicts the next unit from the shared context $\mathbf{y}_u$
(or equivalently from the preceding tuple sequence) as
\begin{equation}
    p_{l}^{\mathsf{ms}} (\cdot \mid \mathbf{q}_{1},\ldots,\mathbf{q}_{u-1}) =
    \mathrm{Softmax} \bigl(
        \mathrm{Linear}_{l}(\mathbf{y}_u)
    \bigr),
    \label{eq:stream_head}
\end{equation}
where $\mathrm{Linear}_l: \mathbb{R}^{d_{\mathsf{slm}}} \rightarrow \mathbb{R}^{C}$.
These heads provide explicit per-stream supervision for the unit transition dynamics,
encouraging the shared backbone to preserve structure useful for joint prediction.

The stream-specific heads predict current units conditioned only on the shared multi-stream history.
Sampling independently from these predicted distributions therefore imposes conditional independence among current units.
To model dependencies within each tuple,
we introduce a \textit{generative head} based on the depth Transformer architecture~\citep{lee2022autoregressive}.
Let $\pi: \{1,\ldots,|\mathcal{J}|\} \rightarrow \mathcal{J}$ be a fixed bijection specifying an ordering of the streams in $\mathcal{J}$.
Starting from the context embedding $\mathbf{y}_u$, the generative head predicts the units in $\mathbf{q}_u$ autoregressively according to the ordering $\pi$ as
\begin{equation}
    p ( \mathbf{q}_u \mid \mathbf{q}_{1},\ldots,\mathbf{q}_{u-1}) =
    \prod_{j=1}^{|\mathcal{J}|} p \bigl(
        q_{\pi(j),u} \mid \mathbf{y}_u,q_{\pi(1),u},\ldots,q_{\pi(j-1),u}
    \bigr).
    \label{eq:generative_head}
\end{equation}
Thus, the shared Transformer models temporal dependencies across sequence positions,
while the generative head captures dependencies among streams within each position.
During generation, the units in each tuple are sampled sequentially according to the ordering $\pi$.

\textbf{Training.}
The multi-stream SLM is trained with a combined objective consisting of
a single-stream loss, a transition-level self-distillation loss, and a multi-stream generative loss.

\textit{Single-stream loss.}
For each training example, we sample a stream $l \sim \mathrm{Unif}(\mathcal{J})$ and
independently deduplicate its frame-level sequence $Z_l$.
This yields $\bar{Q}_l = (\bar{q}_{l,1},\ldots,\bar{q}_{l,\bar{U}_l})$,
where adjacent units are always distinct, and $\bar{U}_l$ is the deduplicated sequence length.
At position $u$,
the model takes the stream-specific embedding $\mathrm{Emb}_{l}(\bar{q}_{l,u})$ as input.
The resulting sequence is processed by the shared temporal Transformer,
and next-unit predictions are produced using the same stream-specific head $\mathrm{Linear}_l$ as in multi-stream mode (cf. Eq.~(\ref{eq:stream_head})).
The single-stream loss is defined as
\begin{equation}
    \mathcal{L}_{\mathsf{single}} =
    - \mathbb{E}_{l \sim \mathrm{Unif}(\mathcal{J})} \! \left[
        \frac{1}{\bar{U}_l}
        \sum_{u=1}^{\bar{U}_l} \log p_{l}^{\mathsf{ss}} (
            \bar{q}_{l,u} \mid \bar{q}_{l,1},\ldots,\bar{q}_{l,u-1}
        )
    \right].
\end{equation}
Because consecutive repetitions have been removed,
this objective emphasizes transitions between distinct units within each stream.

\textit{Transition-level self-distillation.}
The cross-stream-deduplicated sequence $\mathcal{Q}$ contains additional positions introduced by changes in other streams.
To transfer the transition behavior learned in single-stream mode to these multi-stream contexts,
we align the multi-stream predictive distribution with a detached single-stream reference distribution.
Let $\smash{\mathcal{A}_{l} = (a_{l,1},\ldots,a_{l,\bar{U}_{l}})}$ denote the positions in $\mathcal{Q}$ corresponding to the successive units in $\bar{Q}_{l}$.
Thus, $q_{l, a_{l,u}} = \bar{q}_{l,u}$ and,
for $u>1$, $a_{l,u}$ is a retained position at which stream $l$ changes.
At each aligned position, we define the self-distillation loss as
\begin{equation}
    \mathcal{L}_{\mathsf{distill}} =
    \mathbb{E}_{l \sim \mathrm{Unif}(\mathcal{J})} \! \left[
    \frac{1}{\bar{U}_l}
    \sum_{u=1}^{\bar{U}_l}
        \mathrm{KL} \Bigl(
            \mathrm{sg} \bigl(
                p_{l}^{\mathsf{ss}} (
                    \cdot \mid \bar{q}_{l,1},\ldots,\bar{q}_{l,u-1} 
                ) 
            \bigr)
        \,\|\,
        p_{l}^{\mathsf{ms}} \bigl(
                \cdot \mid \mathbf{q}_1,\ldots,\mathbf{q}_{a_{l,u}-1} 
            \bigr) 
        \Bigr)
    \right].
\end{equation}
The single-stream distribution provides rich supervision focused on transitions between distinct units.
Matching this distribution at aligned positions encourages the shared temporal backbone to capture per-stream transition structure efficiently from the complex multi-stream context.

\textit{Multi-stream generative loss.}
The generative head is trained on the cross-stream-deduplicated sequence $\mathcal{Q}$.
Using the factorization in Eq.~(\ref{eq:generative_head}),
the multi-stream generative loss is
\begin{equation}
    \mathcal{L}_{\mathsf{gen}} = 
    -\frac{1}{U|\mathcal{J}|}
    \sum_{u=1}^{U}
    \sum_{j=1}^{|\mathcal{J}|}
    \log p (
        q_{\pi(j),u} \mid \mathbf{y}_u,q_{\pi(1),u},\ldots,q_{\pi(j-1),u}
    ).
\end{equation}
This objective trains the model to predict a coherent tuple at each position
by accounting for both the preceding sequence and units generated earlier within the current tuple.

The final multi-stream SLM objective is
$\mathcal{L}_{\mathsf{slm}} = (1 - \alpha) \left[(1 - \beta) \mathcal{L}_{\mathsf{single}} + \beta \mathcal{L}_{\mathsf{distill}}\right] + \alpha \mathcal{L}_{\mathsf{gen}}$,
where $\alpha$ controls the contribution of multi-stream generation and $\beta$ balances single-stream modeling against transition-level self-distillation.
The two types of heads play complementary roles in training the shared temporal backbone.
The stream-specific heads provide supervision on transitions between distinct units,
while the generative head models the joint distribution over complete tuples,
including per-stream unit persistence and within-tuple dependencies (see Appendix~\ref{app:sssec:slm_ablation_studies} for ablation studies).

\section{Experiments}
\label{sec:experiments}

\subsection{Experimental Setup}
\label{ssec:experimental_setup}

\textbf{Models.}
We compare four models:
\textbf{SpidR SLM}, trained on discrete units extracted by the original SpidR encoder~\citep{poli2025spidr};
\textbf{SpidR 2.0 SLM}, trained on a single unit stream extracted by our SpidR 2.0 encoder;
\textbf{GSLM2}, our multi-stream model trained on unit streams extracted from multiple SpidR 2.0 layers; and
\textbf{Text LM}, trained on text transcriptions as a topline.
Where applicable, we match model size and training data across systems to enable controlled comparisons.

\textbf{Data.} We mainly use the English Libri-Light corpus~\citep{kahn2020libri}.
The speech encoders are pretrained on the 960-hour LibriSpeech corpus~\citep{panayotov2015librispeech}, 
while the SLMs are trained on the 600-, 6K-, and 60K-hour Libri-Light subsets.
The corresponding Text LMs are trained on Libriheavy transcriptions~\citep{kang2024libriheavy}.
To study model scaling beyond Libri-Light,
we additionally use a 6M-hour corpus of English conversational speech sourced from the internet.
We train SLMs on progressively larger subsets, ranging from 600 hours to the full 6M hours,
and train the corresponding Text LMs on transcripts generated by automatic speech recognition (ASR).
For each setting,
we train a byte-pair encoding (BPE)~\citep{sennrich2016neural} tokenizer with a vocabulary of 4,096 tokens,
using the corresponding text data.
Full dataset details are provided in Appendix~\ref{app:ssec:datasets_and_preprocessing}.

\textbf{Architecture.}
For both SpidR and SpidR 2.0,
the student and teacher encoders each comprise $L=12$ Transformer encoder layers and contain 90M parameters.
For SpidR 2.0, the shared prediction heads are applied to layers $\mathcal{K} = \{5,6,\dots,12\}$,
each projecting to a vocabulary of $C = 256$ codes.
The SLMs and Text LMs use decoder-only Transformers instantiated at five parameter scales:
90M, 300M, 1.2B, 2.5B, and 6.5B parameters.
Full architecture details are provided in Appendix~\ref{app:ssec:modeling}.

\textbf{Training.}
Encoder pretraining largely follows the setup used for SpidR.
For SpidR 2.0,
the conditional-entropy weight is fixed at $\mu=0.1$, while the marginal-entropy weight is initialized at $\lambda=0.1$ and reduced to $0.001$ after 20k steps.
SpidR SLM uses units from layer~6, while SpidR 2.0 SLM uses layer~7.
GSLM2 uses six streams extracted from layers $\mathcal{J}=\{5,6,\ldots,10\}$ 
and generates them in ascending order of layer-wise bitrate, following $(10,5,6,9,8,7)$.
Layer selections are based on development-set performance.
We set both loss coefficients, $\alpha$ and $\beta$, in $\mathcal{L}_{\mathsf{slm}}$ to $0.1$.
Appendix~\ref{app:ssec:training} provides full training details and motivates our hyperparameter choices.

\begin{table}[t]
    \centering
    \vspace{-0.1cm}
    \caption{Spoken-language evaluation suite. Metrics described with ``vs.'' use likelihood-based paired comparisons, while the others assess prompted generation.}
    \label{tb:benchmarks}

    \scriptsize
    \setlength{\tabcolsep}{2.5pt}
    \renewcommand{\arraystretch}{1.08}
    \begin{tabular}{llll}
    \toprule
    \textbf{Benchmark} & \textbf{Metric} & \textbf{Capability} & \textbf{Evaluation} \\
    \midrule
    \multirow{2}{*}[0pt]{ZeroSpeech'21}
    & \textbf{sWuggy(-all)} & Lexical & Real word vs.\ phonotactically matched nonword. \\
    & \textbf{sBLIMP} & Syntax & Grammatical vs.\ minimally different ungrammatical sentence.\\
    \cmidrule(l{0.2em}r{0.2em}){1-4}
    \multirow{2}{*}[0pt]{StoryCloze (SC)}
    & \textbf{Topic SC (tSC)} & Semantics & Correct ending vs.\ random ending from another story. \\
    & \textbf{Spoken SC (sSC)} & Semantics & Correct ending vs.\ adversarial, superficially plausible ending. \\
    \cmidrule(l{0.2em}r{0.2em}){1-4}
    \multirow{3}{*}[0pt]{LongTail-Swap}
    & \textbf{WordSwap (WS)} & Semantics & Context-appropriate word vs.\ frequency- and POS-matched substitution. \\
    & \textbf{InflectionSwap (IS)} & Morphosyntax & Context-appropriate inflection vs.\ another form of the same lemma. \\
    & \textbf{AgreementSwap (AS)} & Syntax & Matching vs.\ mismatching grammatical agreement. \\
    \cmidrule(l{0.2em}r{0.2em}){1-4}
    \multirow{2}{*}[0pt]{Continuation}
    & \textbf{ASR-PPL} & Fluency & LM perplexity of an ASR-transcribed continuation at a calibrated temperature. \\
    & \textbf{LLM Judge} & Quality & Rubric-based relevance, coherence, and fluency of the transcribed continuation. \\
    \bottomrule
    \end{tabular}
    \vspace{-0.1cm}
\end{table}

\textbf{Evaluation.}
We evaluate complementary linguistic capabilities using the metrics summarized in Table~\ref{tb:benchmarks}.
For the paired-stimulus tasks from
ZeroSpeech'21 (ZS21)~\citep{nguyen2020zero},
StoryCloze (SC)~\citep{mostafazadeh2016corpus,hassid2023textually}, and
LongTail-Swap (LT-Swap)~\citep{algayres2025longtail},
SLMs score spoken stimuli and Text LMs score their textual counterparts;
accuracy is the proportion of pairs for which the correct item receives the higher score.
The single-stream models use log-likelihoods averaged over units or tokens;
for sWuggy, Text LMs instead use unnormalized log-likelihoods~\citep{poli2025spidr}.
GSLM2 uses the stream-specific heads,
averaging log probabilities over unit-change positions within each stream and then across streams.
This evaluates distinct-unit transitions from temporal context,
following the scoring convention of the independently deduplicated single-stream SLMs.
The resulting transition score differs from the joint sequence likelihood computed with the generative head,
which also accounts for per-stream repetitions and within-tuple dependencies.
Further analysis of scoring choices is provided in Appendix~\ref{app:sssec:scoring_choices}.
Our primary metrics are \textbf{sWuggy}, \textbf{sBLIMP}, and \textbf{tSC},
widely used measures of lexical, syntactic, and semantic competence, respectively.
We also report sSC,
a more challenging variant of tSC that requires greater story-level contextual understanding.
We additionally develop a spoken adaptation of LT-Swap that evaluates performance across word-frequency bins using WordSwap (WS), InflectionSwap (IS), and AgreementSwap (AS),
which probe semantic, morphosyntactic, and syntactic abilities, respectively.
We report the macro-average of the per-bin accuracies.
Alongside the paired-stimulus tasks,
we assess generation quality by producing prompt-conditioned continuations with top-$p$ sampling,
synthesize them using a duration predictor and unit vocoder,
and transcribe the resulting speech via ASR.
For GSLM2, the generative head samples each tuple autoregressively.
For each model, we calibrate the decoding temperature on a held-out set using ground-truth repetition statistics measured by uni/bigram auto-BLEU~\citep{lakhotia2021generative}.
We then compute perplexity (PPL) over the ASR transcriptions (ASR-PPL) using Llama3~\citep{grattafiori2024llama}
and assess the transcribed continuations using a large language model (LLM) judgment score (LLMJ)
assigned by Prometheus~2~\citep{kim2024prometheus}.
For speech encoders,
we evaluate their representations using zero-shot metrics, including
triphone-level ABX discrimination~\citep{schatz2016abx}
and word-level mean average precision (MAP)~\citep{carlin2011rapid}.
Full evaluation details are provided in Appendix~\ref{app:ssec:evaluation}.

\subsection{SLM Evaluation}
\label{ssec:slm_eval}

\begin{figure*}[t]
\centering
\vspace{-0.1cm}
\begin{minipage}[t]{0.52\textwidth}
    \vspace{-0.2cm}
    \centering

    \localcaptiontype{table}
    \caption{Main results on ZeroSpeech'21 (ZS21), StoryCloze (SC), and LongTail-Swap (LT-Swap) for SLMs trained on Libri-Light at three scales. Scale labels indicate ``train data hours / parameter count''. At each scale, the Text LM serves as a topline. Scores are accuracies (\%; chance level: 50\%).}
    \label{tab:main_results}

    \scriptsize
    \setlength{\tabcolsep}{3pt}
    \renewcommand{\arraystretch}{1.08}

    \begin{tabular}{llccccccc}
    \toprule
    & & \multicolumn{2}{c}{\textbf{ZS21} ($\uparrow$)} & \multicolumn{2}{c}{\textbf{SC} ($\uparrow$)} & \multicolumn{3}{c}{\textbf{LT-Swap} ($\uparrow$)} \\
    \cmidrule(l{0.3em}r{0.3em}){3-4}
    \cmidrule(l{0.3em}r{0.3em}){5-6}
    \cmidrule(l{0.3em}r{0.3em}){7-9}
    \multicolumn{2}{l}{\textbf{Model}} & sWug. & sBLI. & tSC & sSC & WS & IS & AS \\
    \midrule

    \multirow{4}{*}[-3pt]{\rotatebox{90}{600\,h / 90M}}
    & Text LM & 73.4 & 61.7 & 75.4 & 53.4 & 73.5 & 80.6 & 62.2 \\
    \cmidrule(l{0.2em}r{0.2em}){2-9}
    & SpidR SLM & 67.4 & 54.2 & 65.4 & 52.4 & 54.3 & 57.9 & 54.2 \\
    & SpidR 2.0 SLM & 69.2 & 57.7 & 68.1 & \textbf{52.7} & 58.0 & 63.5 & 55.0 \\
    & GSLM2 & \textbf{71.5} & \textbf{59.8} & \textbf{74.1} & 52.0 & \textbf{62.4} & \textbf{68.2} & \textbf{59.1} \\
    \cmidrule(l{0.2em}r{0.2em}){1-9}

    \multirow{4}{*}[-2pt]{\rotatebox{90}{6K\,h / 300M}}
    & Text LM & 79.5 & 66.8 & 84.8 & 52.2 & 80.1 & 90.9 & 79.1 \\
    \cmidrule(l{0.2em}r{0.2em}){2-9}
    & SpidR SLM & 71.1 & 59.3 & 71.4 & 50.7 & 62.0 & 63.0 & 58.5 \\
    & SpidR 2.0 SLM & 73.9 & 65.1 & 74.7 & 52.2 & 67.5 & 69.0 & 64.9 \\
    & GSLM2 & \textbf{76.3} & \textbf{67.2} & \textbf{78.9} & \textbf{52.5} & \textbf{75.6} & \textbf{75.5} & \textbf{69.8} \\
    \cmidrule(l{0.2em}r{0.2em}){1-9}

    \multirow{4}{*}[-2pt]{\rotatebox{90}{60K\,h / 1.2B}}
    & Text LM & 83.8 & 71.5 & 89.2 & 54.6 & 87.3 & 94.2 & 85.4 \\
    \cmidrule(l{0.2em}r{0.2em}){2-9}
    & SpidR SLM & 73.9 & 62.4 & 75.1 & 52.8 & 66.4 & 65.5 & 60.7 \\
    & SpidR 2.0 SLM & 76.7 & 69.6 & 81.0 & 56.2 & 74.8 & 72.5 & 66.6 \\
    & GSLM2 & \textbf{79.3} & \textbf{71.8} & \textbf{84.5} & \textbf{57.6} & \textbf{79.5} & \textbf{78.5} & \textbf{69.9} \\
    \bottomrule
    \end{tabular}
\end{minipage}
\hfill
\begin{minipage}[t]{0.46\textwidth}
    \vspace{0pt}
    \centering

    \begin{tikzpicture}
\definecolor{spidrgreen}{RGB}{89,161,79}
\definecolor{spidr2blue}{RGB}{78,121,167}
\definecolor{gslm2red}{RGB}{225,87,89}
\definecolor{textgray}{RGB}{80,80,80}

\pgfdeclareplotmark{filled star}{%
    \pgfpathmoveto{\pgfqpoint{0pt}{\pgfplotmarksize}}%
    \pgfpathlineto{\pgfqpoint{.224514\pgfplotmarksize}{.309017\pgfplotmarksize}}%
    \pgfpathlineto{\pgfqpoint{.951057\pgfplotmarksize}{.309017\pgfplotmarksize}}%
    \pgfpathlineto{\pgfqpoint{.363271\pgfplotmarksize}{-.118034\pgfplotmarksize}}%
    \pgfpathlineto{\pgfqpoint{.587785\pgfplotmarksize}{-.809017\pgfplotmarksize}}%
    \pgfpathlineto{\pgfqpoint{0pt}{-.381966\pgfplotmarksize}}%
    \pgfpathlineto{\pgfqpoint{-.587785\pgfplotmarksize}{-.809017\pgfplotmarksize}}%
    \pgfpathlineto{\pgfqpoint{-.363271\pgfplotmarksize}{-.118034\pgfplotmarksize}}%
    \pgfpathlineto{\pgfqpoint{-.951057\pgfplotmarksize}{.309017\pgfplotmarksize}}%
    \pgfpathlineto{\pgfqpoint{-.224514\pgfplotmarksize}{.309017\pgfplotmarksize}}%
    \pgfpathclose
    \pgfusepath{fill,stroke}%
}

\pgfplotsset{
    results axis/.style={
        width=\linewidth,
        height=3.1cm,
        xmode=log,
        log basis x=10,
        xmin=400,
        xmax=9000000,
        xtick={600,6000,60000,600000,6000000},
        ymin=57,
        ymax=84,
        ytick={60,65,70,75,80},
        xticklabels={},
        ylabel={Accuracy (\%)},
        xticklabel style={font=\scriptsize},
        yticklabel style={
            font=\scriptsize,
            xshift=2pt
        },
        xlabel style={
            font=\scriptsize,
            yshift=2pt
        },
        ylabel style={
            font=\scriptsize,
            yshift=-4pt
        },
        tick align=outside,
        tick style={black!70},
        axis lines=left,
        axis line style={
            black!80,
            line width=0.9pt
        },
        ymajorgrids=true,
        xmajorgrids=true,
        major grid style={
            draw=black!40,
            dash pattern=on 1pt off 1.4pt,
            line width=0.3pt
        },
        line width=0.8pt,
        mark size=1.5pt,
        every axis plot/.append style={solid},
        clip=false
    }
}

\begin{groupplot}[
    results axis,
    group style={
        group size=1 by 3,
        vertical sep=0.25cm
    }
]

\nextgroupplot[
    ymin=63,
    ymax=88,
    ytick={70,80,90},
    minor ytick={75,85},
    yminorgrids=true,
    minor grid style={
        draw=black!40,
        dash pattern=on 1pt off 1.4pt,
        line width=0.3pt
    },
    legend entries={
        {Text LM},
        {GSLM2},
        {SpidR 2.0 SLM},
        {SpidR SLM}
    },
    legend style={
        legend columns=2,
        font=\scriptsize,
        draw=black!50,
        rounded corners=2pt,
        line width=0.8pt,
        fill=none,
        at={(0.5,1.10)},
        anchor=south,
        column sep=3pt
    },
    legend cell align=left
]

\addlegendimage{
    color=textgray,
    mark=*,
    mark options={solid,fill=textgray}
}
\addlegendimage{
    color=gslm2red,
    mark=filled star,
    mark size=2.0pt,
    mark options={
        solid,
        sharp corners,
        fill=gslm2red,
        draw=gslm2red,
    }
}
\addlegendimage{
    color=spidr2blue,
    mark=square*,
    mark options={solid,fill=spidr2blue}
}
\addlegendimage{
    color=spidrgreen,
    mark=triangle*,
    mark size=2.0pt,
    mark options={solid,fill=spidrgreen}
}

\addplot[
    color=spidrgreen,
    mark=triangle*,
    mark size=2.0pt,
    mark options={
        solid,
        fill=spidrgreen
    },
    forget plot
] coordinates {
    (600,64.97)
    (6000,69.40)
    (60000,72.59)
    (600000,73.72)
    (6000000,74.73)
};

\addplot[
    color=spidr2blue,
    mark=square*,
    mark options={
        solid,
        fill=spidr2blue,
    },
    forget plot
] coordinates {
    (600,68.38)
    (6000,72.59)
    (60000,75.98)
    (600000,77.19)
    (6000000,77.59)
};

\addplot[
    color=textgray,
    mark=*,
    mark options={
        solid,
        fill=textgray
    },
    forget plot
] coordinates {
    (600,69.12)
    (6000,76.28)
    (60000,81.05)
    (600000,83.64)
    (6000000,83.74)
};

\addplot[
    color=gslm2red,
    mark=filled star,
    mark size=2.0pt,
    mark options={
        solid,
        fill=gslm2red,
        draw=gslm2red
    },
    forget plot
] coordinates {
    (600,70.73)
    (6000,76.11)
    (60000,78.70)
    (600000,80.01)
    (6000000,80.74)
};

\node[
    anchor=north west,
    font=\scriptsize
] at (rel axis cs:0.02,0.98) {\textbf{sWuggy}};

\nextgroupplot[
    ymin=51,
    ymax=77,
    ytick={60,70},
    minor ytick={55,65,75},
    yminorgrids=true,
    minor grid style={
        draw=black!40,
        dash pattern=on 1pt off 1.4pt,
        line width=0.3pt
    },
]

\addplot[
    color=spidrgreen,
    mark=triangle*,
    mark size=2.0pt,
    mark options={
        solid,
        fill=spidrgreen,
    },
    forget plot
] coordinates {
    (600,53.22)
    (6000,58.82)
    (60000,61.37)
    (600000,62.26)
    (6000000,62.59)
};

\addplot[
    color=spidr2blue,
    mark=square*,
    mark options={
        solid,
        fill=spidr2blue,
    },
    forget plot
] coordinates {
    (600,56.38)
    (6000,63.62)
    (60000,67.90)
    (600000,69.45)
    (6000000,70.03)
};

\addplot[
    color=textgray,
    mark=*,
    mark options={
        solid,
        fill=textgray
    },
    forget plot
] coordinates {
    (600,64.72)
    (6000,70.93)
    (60000,72.76)
    (600000,74.38)
    (6000000,74.31)
};

\addplot[
    color=gslm2red,
    mark=filled star,
    mark size=2.0pt,
    mark options={
        solid,
        fill=gslm2red,
        draw=gslm2red
    },
    forget plot
] coordinates {
    (600,57.98)
    (6000,66.40)
    (60000,71.05)
    (600000,71.98)
    (6000000,72.68)
};

\node[
    anchor=north west,
    font=\scriptsize
] at (rel axis cs:0.02,0.98) {\textbf{sBLIMP}};

\nextgroupplot[
    xlabel={Training Data (hours)},
    xticklabels={600,6K,60K,600K,6M},
    ymin=62,
    ymax=97,
    ytick={70,80,90},
    minor ytick={75,85,95},
    yminorgrids=true,
    minor grid style={
        draw=black!40,
        dash pattern=on 1pt off 1.4pt,
        line width=0.3pt
    },
]

\addplot[
    color=spidrgreen,
    mark=triangle*,
    mark size=2.0pt,
    mark options={
        solid,
        fill=spidrgreen,
    },
    forget plot
] coordinates {
    (600,65.17)
    (6000,71.69)
    (60000,76.01)
    (600000,80.30)
    (6000000,81.77)
};

\addplot[
    color=spidr2blue,
    mark=square*,
    mark options={
        solid,
        fill=spidr2blue,
    },
    forget plot
] coordinates {
    (600,70.83)
    (6000,74.89)
    (60000,81.57)
    (600000,86.60)
    (6000000,88.45)
};

\addplot[
    color=textgray,
    mark=*,
    mark options={
        solid,
        fill=textgray
    },
    forget plot
] coordinates {
    (600,79.43)
    (6000,86.16)
    (60000,92.73)
    (600000,96.29)
    (6000000,96.98)
};

\addplot[
    color=gslm2red,
    mark=filled star,
    mark size=2.0pt,
    mark options={
        solid,
        fill=gslm2red,
        draw=gslm2red
    },
    forget plot
] coordinates {
    (600,74.42)
    (6000,79.45)
    (60000,85.12)
    (600000,89.43)
    (6000000,90.49)
};

\node[
    anchor=north west,
    font=\scriptsize
] at (rel axis cs:0.02,0.98) {\textbf{tSC}};

\end{groupplot}
\end{tikzpicture}

    \localcaptiontype{figure}
    \vspace{-0.4cm}
    \caption{Scaling trends for SLMs and Text LM trained on increasing amounts of internet-sourced conversational speech, up to 6M\,h.}
    \label{fig:scaling}
\end{minipage}
\vspace{-0.2cm}
\end{figure*}

\textbf{Main results.}
Table~\ref{tab:main_results} compares the SLM variants across the three Libri-Light scales,
with Text LMs serving as reference toplines.
On sWuggy, sBLIMP, and tSC,
SpidR 2.0 SLM consistently outperforms SpidR SLM at every scale,
suggesting that its units better preserve linguistically relevant information,
which is examined in Section~\ref{ssec:enc_eval}.
Multi-stream modeling provides substantial further gains:
GSLM2 achieves the best overall SLM performance at every scale,
considerably narrowing the gap with Text LM.
Although sSC scores generally remain close to chance,
both SpidR 2.0 SLM and GSLM2 improve at 60K hours, notably outperforming Text LM.
On LT-Swap, GSLM2 consistently outperforms both single-stream SLMs,
but large gaps to Text LM remain.
The gaps on IS and AS, which probe syntactic abilities,
widen with scale and are particularly pronounced at 60K hours.
Because LT-Swap is constructed from the training data at each scale, however,
these comparisons do not provide a controlled assessment of scalability.
Yet, the remaining gaps suggest that LT-Swap reveals limitations in spoken language modeling that are less apparent on ZS21 and SC,
highlighting the need for broader evaluation coverage.
Appendix~\ref{app:sssec:lt_swap} further analyzes the LT-Swap results.

\textbf{Scalability study.}
Results so far cover scaling on Libri-Light up to 60K hours (Table~\ref{tab:main_results}, Figure~\ref{fig:gslm2}). To probe scaling further, we turn to conversational speech from the internet, reaching 6M hours.
Unlike the read audiobook speech of Libri-Light, this is spontaneous conversational speech, letting us test whether our conclusions hold across domains (see Appendix~\ref{app:ssec:datasets_and_preprocessing} for details). Figure~\ref{fig:scaling} reports sWuggy, sBLIMP, and tSC as we scale models to 2.5B and 6.5B parameters, trained on 600K and 6M hours respectively. Relative to SpidR SLM, GSLM2 substantially narrows the gap to the Text LM toplines at every scale and on every metric. 
At 6M hours, the additional data an SLM would need to match the Text LM falls from roughly four orders of magnitude for SpidR SLM, to two orders for GSLM2 on the average metric.
Our two modeling improvements contribute differently: the gain from SpidR 2.0 units compounds with scale,
whereas multi-stream modeling contributes a roughly constant error reduction at every scale (see Appendix~\ref{app:sssec:scaling_exponents} for detailed scaling analyses).

\begin{table}[t]
    \centering
    \vspace{-0.3cm}
    \caption{Comparison of GSLM2 with prior SLMs under different training regimes. Prior results are from cited papers unless marked $\ddagger$ (our evaluation). Bold indicates the best overall result; underlining indicates the best prior baseline in each regime. $^\dagger$The amount of speech data used for training.}
    \label{tab:slm_comparison}
    
    \scriptsize
    \setlength{\tabcolsep}{6pt}
    \begin{tabular}{llcccccc}
        \toprule
        \textbf{Training Regime}
        & \textbf{Model}
        & \textbf{\#Params}
        & \textbf{\#Data}$^\dagger$\textbf{[h]}
        & \textbf{sWuggy}
        & \textbf{sBLIMP}
        & \textbf{tSC}
        & \textbf{sSC} \\
        \midrule

        \multirow{8}{*}[-3pt]{Speech-only training}
        & \multirow{2}{*}{GSLM2 (ours; Libri-Light)}
        & 300M & 6K  & 76.3 & 67.2 & 78.9 & 52.5 \\
        & & 1.2B & 60K & 79.3 & 71.8 & 84.5 & 57.6 \\
        \cdashlinelr{2-8}
        & GSLM2 (ours; Conversational Speech) & 6.5B & 6M & \textbf{80.7} & \textbf{72.9} & \textbf{90.5} & \textbf{63.2} \\
        \cmidrule(l{0.2em}r{0.2em}){2-8}
    
        & GSLM~\citep{lakhotia2021generative} %
        & 100M & 6K & 64.8 & 54.2 & 66.6 & 53.3 \\
        & AudioLM~\citep{borsos2023audiolm}
        & 300M & 60K & 71.5 & \underline{64.7} & -- & -- \\
        & \citet{cuervo2024scaling}
        & 823M & 160K & -- & 61.3 & 75.4 & -- \\
        & Moshi~\citep{defossez2024moshi}
        & 7B & 7M & \underline{74.8} & 59.9 & \underline{80.9} & \underline{56.9} \\
        & SyllableLM~\citep{baade2025syllablelm}
        & 300M & 60K & 72.2 & 63.7 & 75.4 & -- \\

        \cmidrule(l{0.2em}r{0.2em}){1-8}

        \multirow{4}{*}[0pt]{\shortstack[l]{Speech-only training\\\ \ + text-LM init.}}
        & TWIST~\citep{hassid2023textually}
        & 13B & 150K & \underline{74.5} & 59.2 & 76.4 & 55.4 \\
        & Moshi~\citep{defossez2024moshi}
        & 7B & 7M & 74.3 & 58.9 & \underline{81.8} & \underline{58.7} \\
        & Flow-SLM~\citep{chou2025flow}
        & 7B & 45K & 69.8 & \underline{60.0} & -- & -- \\
        & Llama-Mimi~\citep{sugiura2025llama}
        & 8B & 240K & 68.8 & 55.1 & 67.6 & -- \\

        \cmidrule(l{0.2em}r{0.2em}){1-8}

        \multirow{7}{*}[0pt]{\shortstack[l]{Speech-text training\\\ \ + text-LM init.}}
        & VoxtLM~\citep{maiti2024voxtlm}
        & 1.3B & 60K & 66.1 & 57.1 & -- & -- \\
        & Moshi~\citep{defossez2024moshi}
        & 7B & 7M & 72.6 & 58.8 & 83.0 & 60.8 \\
        & SpiRit-LM~\citep{nguyen2025spirit}
        & 7B & 458K & 69.0 & 58.3 & 82.9 & 61.0 \\
        & Align-SLM~\citep{lin2025align}
        & 7B & 4K & \underline{77.9} & \underline{62.3} & \underline{86.8} & 61.1 \\
        & BSLM~\citep{luo2026bslm}
        & 4B & 220K & 74.1 & 60.2 & 84.1 & \underline{61.3} \\ 
        & LST~\citep{lu2026latent}
        & 1.8B & 164K & 72.8 & 59.0 & 79.5 & 60.8 \\
        & PersonaPlex$^{\ddagger}$\citep{roy2026personaplex}
        & 7B & 2K & 65.4 & 54.9 & 80.1 & 58.9 \\
        \bottomrule
    \end{tabular}%
    \vspace{-0.1cm}
\end{table}

\textbf{Comparison against previous SLMs.}
Table~\ref{tab:slm_comparison} compares GSLM2 with previous SLMs trained under different regimes.
Among models trained solely on speech,
GSLM2 achieves the best overall performance across model and data scales.
Despite using a similar architecture,
our 1.2B model outperforms the 7B Moshi on all four metrics,
highlighting strong parameter and data efficiency.
Scaling GSLM2 to 6.5B parameters yields further gains.
GSLM2 also remains competitive with models that use pretrained LMs or text data.
Such models perform well on semantic tasks (i.e., tSC and sSC),
but tend to lag on sWuggy and sBLIMP, which can require finer lexical and syntactic distinctions.
GSLM2 leads on the latter benchmarks while remaining competitive on semantic tasks,
suggesting that its multi-level units support broad and balanced linguistic competence.
Appendix~\ref{app:sssec:slm_eval_speech_llm} provides additional results for speech-capable LLMs trained at much larger scales.

\begin{wraptable}{r}{0.45\textwidth}
    \centering
    \vspace{-0.5cm}
    \caption{Prompt-conditioned continuation results. $\hat{\tau}$ is the decoding temperature calibrated with ground-truth repetition statistics.}
    \label{tab:gen_eval}

    \scriptsize
    \setlength{\tabcolsep}{4pt}
    \renewcommand{\arraystretch}{1.08}

    \begin{tabular}{llccc}
    \toprule
    \multicolumn{2}{l}{\textbf{Model}} & $\hat{\tau}$ & \textbf{ASR-PPL} ($\downarrow$) & \textbf{LLMJ} ($\uparrow$) \\
    \midrule

    \multirow{4}{*}[-2pt]{\rotatebox{90}{600\,h / 90M}}
    & Text LM & 1.22 & 690.0\,$\pm$\,12.9 & 1.88\,$\pm$\,0.02 \\
    \cmidrule(l{0.2em}r{0.2em}){2-5}
    & SpidR SLM & 1.31 & 670.5\,$\pm$\,\phantom{0}9.4 & 1.73\,$\pm$\,0.02 \\
    & SpidR 2.0 SLM & 1.34 & 626.5\,$\pm$\,14.3 & 1.78\,$\pm$\,0.02 \\
    & GSLM2 & 1.49 & \textbf{583.5}\,$\pm$\,12.1 & \textbf{1.80}\,$\pm$\,0.02 \\
    \cmidrule(l{0.2em}r{0.2em}){1-5}

    \multirow{4}{*}[-2pt]{\rotatebox{90}{6K\,h / 300M}}
    & Text LM & 1.21 & 331.0\,$\pm$\,\phantom{0}6.5 & 2.30\,$\pm$\,0.03 \\
    \cmidrule(l{0.2em}r{0.2em}){2-5}
    & SpidR SLM & 1.39 & 407.2\,$\pm$\,\phantom{0}9.7 & 2.20\,$\pm$\,0.02\\
    & SpidR 2.0 SLM & 1.31 & 367.5\,$\pm$\,\phantom{0}8.0 & 2.27\,$\pm$\,0.03 \\
    & GSLM2 & 1.58 & \textbf{326.9}\,$\pm$\,11.3 & \textbf{2.43}\,$\pm$\,0.03 \\
    \cmidrule(l{0.2em}r{0.2em}){1-5}

    \multirow{4}{*}[-2pt]{\rotatebox{90}{60K\,h / 1.2B}}
    & Text LM & 1.20 & 172.7\,$\pm$\,\phantom{0}6.8 & 2.87\,$\pm$\,0.04 \\
    \cmidrule(l{0.2em}r{0.2em}){2-5}
    & SpidR SLM & 1.50 & 259.4\,$\pm$\,\phantom{0}7.3 & 2.78\,$\pm$\,0.03 \\
    & SpidR 2.0 SLM & 1.45 & 247.4\,$\pm$\,\phantom{0}3.5 & 2.89\,$\pm$\,0.05 \\
    & GSLM2 & 1.59 & \textbf{189.1}\,$\pm$\,\phantom{0}5.5 & \textbf{3.05}\,$\pm$\,0.04 \\
    \bottomrule
    \end{tabular}
    \vspace{-0.1cm}
\end{wraptable}
\textbf{Generative evaluation.}
Table~\ref{tab:gen_eval} reports prompt-conditioned continuation results on ASR-PPL and LLMJ.
We also report the decoding temperature $\hat{\tau}$,
calibrated per model against ground-truth repetition statistics (see Appendix~\ref{app:sssec:asr_ppl} for details).
Among the SLMs, SpidR 2.0 consistently improves over SpidR, while GSLM2 provides further gains, 
suggesting that the advantages in paired-stimulus evaluation transfer to generation.
Text LM shows a different trend:
with more training data, its PPL improves, but its LLMJ score worsens relative to SLMs.
This may reflect a granularity mismatch between fine-grained speech units and coarser BPE tokens,
with SLMs better preserving the local lexical coherence favored by LLMJ.
In the 60K-hour setting, for example,
GSLM2 produces 33\% more continuations rated 4 or 5 than Text LM.
The LLM judge explicitly describes nearly all of these high-scoring continuations as coherent in its rationales.
GSLM2 also outperforms Text LM on ASR-PPL in the 600- and 6K-hour settings,
though these comparisons are not strictly fair because SLM outputs are evaluated through ASR.
The calibrated temperatures are also noticeably higher than those typically used for LM decoding.
Appendix~\ref{app:sssec:calibration_curves} details the temperature calibration process,
while Appendix~\ref{app:sssec:gen_eval_fixed_temp} examines the effects of evaluation protocol differences and presents fixed-temperature comparisons.

\subsection{Evaluation of Encoder Representations}

\label{ssec:enc_eval}
\paragraph{Layer-wise evaluations.}

\begin{wraptable}{r}{0.39\textwidth}
\vspace{-0.5cm}
\caption{ABX and MAP results for DinoSR, SpidR, and SpidR 2.0, reported at the best layer $l^*$ for each metric.}
\label{tab:encoder-eval}
\centering
\scriptsize
\setlength{\tabcolsep}{3pt}
\begin{tabular}{l c cc c c c}
\toprule
& \multicolumn{4}{c}{\textbf{ABX} ($\downarrow$)} & \multicolumn{2}{c}{\textbf{MAP} ($\uparrow$)} \\
\cmidrule(l{0.3em}r{0.3em}){2-5}
\cmidrule(l{0.3em}r{0.3em}){6-7}
\textbf{Model} & $l^*$ & within & across & Avg. & $l^*$ & Score \\
\midrule
DinoSR    & 5  & 4.58 & 6.02 & 5.30 & 11 & 69.49 \\
SpidR     & 6  & 3.53 & 4.31 & 3.92 & 5  & 63.42 \\
\midrule
SpidR 2.0 & 10 & \textbf{2.54} & \textbf{3.13} & \textbf{2.83} & 7 & \textbf{69.58} \\
\bottomrule
\end{tabular}
\vspace{-0.2cm}
\end{wraptable}

To examine how complementary information is distributed across the SpidR 2.0 encoder layers,
we evaluate representations from each layer using ABX and MAP.
Table~\ref{tab:encoder-eval} compares SpidR~2.0 with official checkpoints of SpidR and DinoSR.
To our knowledge, SpidR achieves the best previously reported ABX performance among self-supervised speech encoders.
SpidR 2.0 outperforms both encoders on ABX and matches DinoSR on MAP,
with phoneme-/word-level specialization at layers~10 and~7, respectively.
Appendix~\ref{app:ssec:literature-encoder-evals} extends this comparison to a broader set of speech encoders,
further supporting the effectiveness of SpidR 2.0 on both metrics.

\input{figures/layerwise_all_tikz}

Figure~\ref{fig:layerwise} compares the layer-wise ABX and MAP performance across the three speech encoders.
For SpidR and SpidR 2.0,
we report results from eight encoders initialized and trained with different random seeds.
SpidR 2.0 maintains strong ABX and MAP scores from layer~5 onward,
whereas SpidR and DinoSR peak sharply at a single layer.
To further probe layer-wise specialization,
we train single-stream SLMs on units associated with each layer and evaluate them on sWuggy, sBLIMP, and tSC.
We use the best-performing discretization method for each model at layer~$l$:
SpidR applies $k$-means to representations from student layer~$l$;
DinoSR predicts codebook assignments from teacher layer~$l$ using the corresponding prediction head at the final student layer; and
SpidR 2.0 uses codebook prediction from student layer~$l$.
For SpidR 2.0, sWuggy and sBLIMP scores are highest with units from shallower layers,
whereas tSC benefits more from units extracted from deeper layers.
Every evaluated SpidR 2.0 layer outperforms SpidR and DinoSR.
These findings suggest that SpidR 2.0 learns consistently high-quality representations that capture complementary linguistic information across layers,
providing a strong foundation for multi-stream modeling.
Appendices~\ref{app:sssec:codebook-metrics},~\ref{app:sssec:superb}, and~\ref{app:sssec:stream-similarities} further study these differences through analyzing codebook properties, probing representations on supervised downstream tasks, and assessing stream similarities, respectively.

\textbf{Layer-wise stability across training runs.}
Figure~\ref{fig:layerwise} shows that SpidR 2.0 achieves more stable unit quality across training runs than SpidR.
At the best-performing layer,
SpidR 2.0 reduces the standard deviation of ABX across training seeds by a factor of ten relative to SpidR.
The difference is most pronounced at layers~7-12,
where SpidR exhibits substantial variation on ABX and MAP scores across seeds,
while SpidR 2.0 consistently preserves high-quality representations.
This stability enables more reliable unit extraction for subsequent SLM training.
Ablations in Appendix~\ref{app:sssec:spidr2_ablation} show how each design change from SpidR to SpidR 2.0 affects representation quality and stability.

\section{Conclusion}
We presented GSLM2,
a framework for efficient spoken language modeling at multiple levels of abstraction.
SpidR 2.0 learns stable, complementary discrete unit spaces across multiple encoder layers, 
while the multi-stream SLM captures dependencies within and across the resulting streams.
Across data scales, model sizes, and domains,
GSLM2 improved lexical, syntactic, semantic, and generation performance,
narrowing the data-efficiency gap with text LMs.
These findings suggest multi-level speech representations as a promising foundation for efficient speech-native modeling.

\textbf{Limitations.}
Here, \S~points to related appendix sections.
This work primarily focuses on modeling linguistic and phonetic content from speech without explicitly preserving acoustic properties such as prosody,
which are important for tasks including speech resynthesis (\S\ref{app:ssec:speech_resynthesis}).
Multi-stream modeling also increases computational cost, particularly during training (\S\ref{app:ssec:training}).
Our analyses are limited to empirical comparisons and probing,
providing only a partial understanding of how individual streams contribute to GSLM2 and relate to one another (\S\ref{app:sssec:layer_diversity} and \ref{app:sssec:stream-similarities}).
Finally, our evaluation relies on synthetic benchmarks and indirect generation metrics that, despite their widespread use, may conflate language modeling errors with those introduced by TTS, vocoding, or ASR (\S\ref{app:sssec:lt_swap}, \ref{app:sssec:gen_eval_fixed_temp}, and \ref{app:sssec:cascaded_asr_slm}).

\subsubsection*{Acknowledgments}
We would like to thank Hongyu Gong, Zehai Tu, and Maxime Poli for helpful discussions and support on this project.
ED in his EHESS role was supported in part by the Agence Nationale pour la Recherche (ANR17-EURE0017 Frontcog, ANR10-IDEX-0001-02 PSL*) and an ERC grant (InfantSimulator).

\subsection*{AI use statement}

In this work, we used generative AI systems to construct portions of our spoken-language benchmarks,
including translating source material, filtering samples for suitability for speech synthesis, and synthesizing speech using text-to-speech models (see Appendices~\ref{app:sssec:zerospeech21}-\ref{app:sssec:longtail_swap}).
Generative AI also provided feedback on experimental design and assisted in implementing some portions of the method and its associated training and evaluation procedures.
We further used LLMs as automated judges during generative evaluation (see Appendix~\ref{app:sssec:llm_judge}).
We did not use generative AI to develop
theoretical models or conceptual frameworks,
formulate mathematical claims,
provide arguments for or assist in writing mathematical proofs,
propose or refine research hypotheses,
or interpret experimental results.

Additionally, we used generative AI to suggest experimental parameters;
generate, edit, and debug code;
produce ASR-generated transcriptions for model training and generative evaluation (see Appendices~\ref{app:ssec:datasets_and_preprocessing} and~\ref{app:ssec:evaluation});
draft and improve the readability of manuscript text;
search for, identify, and summarize relevant literature;
and support brainstorming about potential experimental directions.
Apart from outputs produced as part of the explicitly described benchmark-construction and evaluation pipelines, 
AI outputs were treated as suggestions and incorporated only after author review.

AI-assisted code was manually inspected by at least one author,
compared against reference implementations where applicable,
and validated by reproducing the reported results.
The entire manuscript was reviewed by the authors.
Literature summaries, citations, and factual claims were checked against the original publications.
Potential overlap with prior work was assessed through a manual literature survey.
Experimental results were verified through repeated runs,
inspection of raw outputs,
statistical analyses,
and independent review by other authors.
The authors were responsible for interpreting the results and drawing scientific conclusions.
We take responsibility for the final content of this work,
including all text, claims, code, data, and other artifacts produced with the aid of generative AI.

\subsection*{Reproducibility statement}

We document our methods and experimental procedures throughout the main paper, appendices, and supplementary materials.
Section~\ref{sec:gslm2} presents the proposed method and training procedure.
Section~\ref{ssec:experimental_setup} provides a concise overview of the experimental settings,
while Appendices~\ref{app:ssec:modeling} and~\ref{app:ssec:training} provide further implementation details,
including model configurations, optimization settings, computational resources, and experimental hyperparameters.
Appendix~\ref{app:ssec:datasets_and_preprocessing} identifies the source datasets and explains the preprocessing procedures.
Appendix~\ref{app:ssec:evaluation} describes the public benchmarks used in our experiments,
details our benchmark-construction process (including translation, filtering, speech synthesis, and quality control), and
specifies the evaluation protocols, metrics, prompts, and model versions.
Appendices~\ref{app:ssec:supplementary_slm_evaluation} and~\ref{app:ssec:supplementary_encoder_evaluation} provide additional analyses supporting our hyperparameter choices across multiple components of the method.
To assess the statistical reliability of our findings,
Appendix~\ref{app:ssec:statistical_study} quantifies statistical uncertainty in model comparisons using confidence intervals
and assesses variability across encoder training runs with different random seeds.
The supplementary materials include anonymized source code, configuration files, prompts, and data-processing and evaluation scripts used in the main experiments.
We release code, model checkpoints, and derived benchmark artifacts,
as permitted by the applicable licenses.
For artifacts that cannot be redistributed,
we will provide scripts and instructions to reconstruct them.

\bibliography{mybib}
\bibliographystyle{iclr2027_conference}

\appendix

\section{Additional Related Work}
\label{app:related_work}

\paragraph{Speech tokenization for spoken language modeling.}
A central distinction among speech tokenizers is the type of information their units are optimized to preserve.
Neural codecs, such as SoundStream~\citep{zeghidour2021soundstream} and EnCodec~\citep{defossez2023encodec},
learn quantized representations optimized for waveform reconstruction.
Although this objective preserves the acoustic detail required for high-fidelity resynthesis,
the resulting units may retain variation irrelevant to linguistic prediction.
Thus, these units may not be linguistically invariant or readily predictable in language modeling~\citep{chang2025dc}.
SLMs therefore commonly rely on units derived from self-supervised speech encoders.\footnote{
Methods that use text supervision or pretrained LLMs to learn semantically aligned speech tokens~\citep{turetzky2024last,zhang2025mimo,tseng2026taste} are complementary to our approach;
here, we focus on textless GSLM.
}
A canonical example is HuBERT~\citep{hsu2021hubert},
which is trained with a self-supervised masked-prediction objective over discrete targets and learns hidden representations that capture phonetic and linguistic structure~\citep{wells2022phonetic,sicherman2023analysing}.
These representations are most often discretized by applying post-hoc $k$-means clustering to a selected encoder layer~\citep{lakhotia2021generative}.
DinoSR~\citep{liu2023dinosr} and SpidR~\citep{poli2025spidr} instead learn discrete targets online
during masked-prediction-based pretraining.
Nevertheless, online codes do not uniformly outperform post-hoc $k$-means units;
for SpidR in particular, post-hoc units perform better on several downstream SLM metrics.

SpidR 2.0 advances this line of work by learning differentiable unit spaces at multiple encoder layers through shared student-teacher prediction heads~\citep{zhou2022image,oquab2024dinov} and entropy regularization~\citep{assran2022masked}.
These techniques originate in self-supervised vision models and remain underexplored for discrete speech tokenization.
Adapting them to speech enables SpidR 2.0 to learn stable, linguistically informative unit spaces at multiple levels of abstraction during pretraining,
eliminating post-hoc $k$-means quantization and providing complementary unit streams for downstream spoken language modeling.

Complementary approaches use continuous speech representations directly for spoken language modeling~\citep{algayres2023generative,ramapuram2026scaling},
or jointly model discrete semantic units and continuous acoustic
or paralinguistic representations~\citep{chen2025variational,chou2025flow}.
Given our emphasis on linguistic competence over reconstruction
fidelity,
we focus on discrete units as a representational bottleneck
that can suppress linguistically irrelevant acoustic
variation~\citep{nguyen2022discrete}.

\paragraph{Multi-level speech representations.}
Prior work has represented speech using multiple discrete levels,
most often to separate high-level content from the acoustic detail required for waveform reconstruction.
AudioLM~\citep{borsos2023audiolm} cascades semantic, coarse acoustic, and fine acoustic tokens.
Related codec-based approaches realize this content-fidelity trade-off in different ways:
SpeechTokenizer~\citep{zhang2024speechtokenizer} distributes semantic and acoustic information hierarchically across residual vector-quantization (RVQ) layers;
SemantiCodec~\citep{liu2024semanticodec} uses separate semantic and acoustic encoders;
and X-Codec~\citep{ye2025codec} injects pretrained semantic features into an acoustic codec and introduces a semantic reconstruction objective.
HAC~\citep{khurana2025factorized} further factorizes a codec bottleneck into acoustic, 
phonetic, and lexical token sets using distillation from HuBERT and LaBSE~\citep{feng2022language}.
Despite their architectural differences, these methods organize speech representations around the trade-off between content modeling and reconstruction fidelity.

A complementary line of work exploits representations from different encoder depths.
MR-HuBERT~\citep{shi2024multi} adapts HuBERT's training procedure to learn units at multiple temporal resolutions from such representations.
More closely related to our representation scheme,
MMM~\citep{shi2024mmm} constructs multiple discrete streams by applying iterative residual $k$-means quantization to several layers of a self-supervised speech encoder,
and evaluates them on ASR, speech resynthesis, and text-to-speech (TTS).
GSLM2 likewise exploits representations across encoder depth,
but learns the corresponding unit spaces jointly during self-supervised pretraining through layer-wise student-teacher prediction.
It then treats the resulting units as parallel, complementary levels of abstraction for spoken language modeling
rather than as a hierarchy of residual refinements for waveform reconstruction.

\paragraph{Multi-stream sequence modeling.}
Multi-stream architectures differ not only in what their streams represent,
but also in how they factorize the joint distribution over those streams.
In textless GSLM,
pGSLM~\citep{kharitonov2022text} jointly models phonetic units, duration, and pitch,
whereas dGSLM~\citep{nguyen2023generative} jointly models the two speaker channels of spoken dialogue.
Codec-based LMs instead operate on multiple quantizer streams.
MusicGen~\citep{copet2023simple} and SoCodec~\citep{guo2024socodec} use delay patterns to express generation across time and codebook depth as a parallel autoregressive sequence.
Hierarchical audio models such as UniAudio~\citep{yang2023uniaudio} and GPST~\citep{zhu2024generative} adapt the RQ-Transformer principle~\citep{lee2022autoregressive},
using a large temporal or global Transformer to model dependencies across frames and a smaller local Transformer to autoregressively model codec tokens within each frame.
Moshi~\citep{defossez2024moshi} employs this factorization for spoken dialogue generation:
its temporal Transformer conditions on all codec streams from preceding frames,
while its depth Transformer autoregressively predicts the semantically distilled first codebook and the residual acoustic codebooks at the current frame.
Hibiki~\citep{labiausse2025high} extends this approach to simultaneous speech-to-speech translation.

Our multi-stream SLM repurposes this temporal-depth factorization for time-aligned units extracted from different layers of a speech encoder,
with each stream representing a distinct level of abstraction rather than a level of residual codec quantization.
The shared temporal backbone integrates the histories of all streams,
while the generative head autoregressively factorizes the joint next-step distribution over the layer-specific units,
thereby capturing dependencies across levels of abstraction.
In parallel, stream-specific heads directly predict the next unit in each stream, providing explicit objectives for modeling its temporal transition dynamics.

\section{Additional Experimental Details}
\label{app:additional_experimental_details}

This section details the experimental setup for both our main experiments (Section~\ref{sec:experiments}) and additional experiments (Appendix~\ref{app:additional_experiments}).
The material is organized as follows:
\begin{itemize}[leftmargin=1.5em]
\item Appendix~\ref{app:ssec:datasets_and_preprocessing}: Datasets and preprocessing
\item Appendix~\ref{app:ssec:modeling}: Model architecture and design choices
\item Appendix~\ref{app:ssec:training}: Training configurations
\item Appendix~\ref{app:ssec:evaluation}: Evaluation metrics, protocols, and configurations
\end{itemize}

\subsection{Datasets and Preprocessing}
\label{app:ssec:datasets_and_preprocessing}

\begin{table}[t]
    \centering
    \caption{Training corpora, languages, and data scales for GSLM components and text LMs.}
    \label{tab:datasets}

    \small
    \renewcommand{\arraystretch}{1.12}

    \begin{tabular}{llll}
    \toprule
    \textbf{Corpus} & \textbf{Language(s)} & \textbf{Component} & \textbf{Training data} \\
    \midrule

    LibriSpeech
    & English & Speech encoder & 960\,h \\
    \cmidrule(l{0.2em}r{0.2em}){1-4}

    \multirow{2}{*}{Libri-Light}
    & \multirow{2}{*}{English} & SLM & 600\,h / 6K\,h / 60K\,h \\
    & & Duration predictor & 600\,h \\
    \cmidrule(l{0.2em}r{0.2em}){1-4}

    Libriheavy
    & English & Text LM & Transcripts aligned with Libri-Light \\
    \cmidrule(l{0.2em}r{0.2em}){1-4}

    LJ Speech
    & English & Unit vocoder & 24\,h \\
    \cmidrule(l{0.2em}r{0.2em}){1-4}

    \multirow{3}{*}{VoxPopuli}
    & \multirow{3}{*}{En, Fr, De} & Speech encoder & 6K\,h per language \\
    & & SLM & 6K\,h per language \\
    & & Text LM & ASR transcripts of the same 6K\,h \\
    \cmidrule(l{0.2em}r{0.2em}){1-4}

    \multirow{2}{*}{\shortstack[l]{Conversational\\Speech}}
    & \multirow{2}{*}{English} & SLM & 600\,h / 6K\,h / 60K\,h / 600K\,h / 6M\,h \\
    & & Text LM & ASR transcripts of the same 600\,h to 6M\,h subsets \\
    \bottomrule
    \end{tabular}
\end{table}

Table~\ref{tab:datasets} summarizes the training data for the four GSLM components---the speech encoder, SLM, duration predictor, and unit vocoder---as well as the corresponding text LMs.
Unless otherwise noted, we use the released audio without applying corpus-specific filtering.

\paragraph{LibriSpeech.}
The speech encoders, including HuBERT, DinoSR, SpidR and SpidR 2.0, are trained on the full 960-hour LibriSpeech training set~\citep{panayotov2015librispeech},
comprising \texttt{train-clean-100}, \texttt{train-clean-360}, and \texttt{train-other-500} subsets of English audiobook speech.
Pretraining is fully self-supervised and uses only the speech waveforms, without access to the corresponding transcriptions.
We use \texttt{dev-clean} and \texttt{dev-other} for validation during encoder development.
All encoders operate on raw waveforms.

\paragraph{Libri-Light.}
The SLMs are trained on the \texttt{unlab-600}, \texttt{unlab-6k}, and \texttt{unlab-60k} subsets of Libri-Light~\citep{kahn2020libri}, 
which contain approximately 600, 6K, and 60K hours of untranscribed English audiobook speech, respectively.
The subsets are nested, such that each smaller training set is contained within the next larger one.
We use the official Libri-Light segmentation pipeline\footnote{\url{https://github.com/facebookresearch/libri-light}} to divide the raw recordings into segments of approximately 60 seconds.

To construct the SLM training data,
we pass the segmented Libri-Light waveforms through the corresponding pretrained speech encoder and extract sequences of discrete speech units.
For tokenizers requiring post-hoc quantization, i.e., HuBERT and SpidR,
we fit a $k$-means model to representations extracted from the \texttt{train-clean-100} subset of LibriSpeech,
where the number of clusters is set to 256, unless otherwise noted.
The DinoSR and SpidR 2.0 units are obtained directly from their learned prediction heads.
To resynthesize speech from deduplicated unit sequences,
we train the duration predictor on the unit sequences extracted from the \texttt{unlab-600} subset.

\paragraph{Libriheavy.}
For the Libri-Light experiments,
the text LMs are trained on Libriheavy~\citep{kang2024libriheavy},
which provides transcripts aligned with the Libri-Light recordings.
At each scale, we use the transcripts corresponding to the respective Libri-Light subset.
The text LM and SLM therefore use the same underlying audiobook data and approximately matched amounts of linguistic content, enabling a controlled comparison of their scaling behavior.

We use the official long-form resegmentation of Libriheavy,\footnote{\url{https://huggingface.co/datasets/pkufool/libriheavy_long}}
in which consecutive transcripts are grouped into multi-sentence segments to preserve greater cross-sentence continuity.
In preliminary experiments,
we found this longer context to be important for training effective text LMs,
particularly for sentence-level evaluations such as tSC and sSC.
The released transcripts are used without additional normalization or filtering.
For each training-data scale, we train a separate BPE tokenizer~\citep{sennrich2016neural} with a vocabulary of 4,096 tokens on the corresponding training transcripts.

\paragraph{LJ Speech.}
The unit HiFi-GAN vocoder~\citep{polyak2021speech} is trained on the 24-hour LJ Speech corpus~\citep{ito2017lj}.
We hold out 250 utterances for validation and testing, leaving approximately 23 hours for training.

\paragraph{VoxPopuli.}
For the multilingual experiments,
we use the English, French, and German portions of VoxPopuli~\citep{wang2021voxpopuli},
which contain approximately 24K, 23K, and 23K hours of speech respectively.
VoxPopuli consists of European Parliament recordings and exhibits substantially more heterogeneous acoustic conditions than Libri-Light,
including greater variation in speakers, recording environments, channels, and background noise.

For each language, we construct training examples from the original unsegmented VoxPopuli recordings\footnote{\url{https://github.com/facebookresearch/voxpopuli}}.
We first apply voice activity detection using pyannote~\citep{bredin2023pyannote}\footnote{\url{https://huggingface.co/pyannote/segmentation-3.0}} to identify contiguous speech regions.
We then perform language identification (LID) on each region using both Whisper large-v3~\citep{radford2023robust}\footnote{\url{https://huggingface.co/openai/whisper-large-v3}} and MMS-LID~\citep{pratap2024scaling}\footnote{\url{https://huggingface.co/facebook/mms-lid-1024}}.
We retain speech regions of at least 0.5 seconds and merge adjacent regions separated by less than 5 seconds of silence, provided that the merged interval does not contain a region rejected by LID.
Any resulting example longer than 60 seconds is recursively split at the longest interior silence until the maximum-duration constraint is satisfied.
Intervening nonspeech intervals are removed, resulting in speech-only examples between 0.5 and 60 seconds in duration.
After filtering and segmentation, the English, French, and German collections contain approximately 16K, 23K, and 23K hours of speech, with mean example durations of 21.6, 26.2, and 27.1 seconds, respectively.

For each language, we independently sample a 6K-hour subset and use it to train both the speech encoder and the SLM.
For the SpidR-based SLM, we fit a $k$-means model to representations extracted from a randomly sampled 100-hour subset of the corresponding training data,
with the number of clusters set to 256.
To construct the text LM training data,
we transcribe the same 6K-hour speech subsets using Whisper large-v3.
For each language, we train a separate BPE tokenizer with a vocabulary of 4,096 tokens on the resulting ASR transcriptions
and use the tokenized sequences to train the corresponding text LM.

\paragraph{Conversational Speech.}

To study scaling beyond Libri-Light, we use English conversational speech sourced from the internet, totaling approximately 6M hours. We combine the public Emilia corpus~\citep{he2024emilia} with off-the-shelf conversational audio licensed from two commercial data providers.
We construct subsets of 600, 6K, 60K, and 600K hours by randomly sampling segments from the 6M-hour superset, keeping the proportion of the three sources constant across subsets.

We use the English portion of Emilia as released, together with its published ASR transcripts.
The Emilia-Pipe preprocessing pipeline has already filtered the data for quality and segmented them into single-speaker utterances lasting 3-30 seconds.

For the licensed audio, we first apply quality-based filtering to the source recordings. We estimate signal-to-noise ratio (SNR) using brouhaha~\citep{lavechin2023brouhaha}, perform LID with Whisper large-v3, and apply audio event detection classification to discard recordings dominated by non-speech content. We then construct training examples from the retained recordings. We apply voice activity detection to identify contiguous speech regions, retaining regions of at least 0.5 seconds and merging adjacent regions separated by less than 5 seconds of silence. Any resulting example longer than 90 seconds is recursively split at the longest interior silence until the maximum-duration constraint is satisfied. Intervening nonspeech intervals are removed, yielding speech-only examples between 0.5 and 90 seconds in duration. To build the corresponding text corpus, we transcribe these examples with faster-whisper,\footnote{\url{https://github.com/SYSTRAN/faster-whisper}} using the large-v3 checkpoint.

SLM training data is constructed as per Libri-Light: we pass the segmented waveforms through the pretrained speech encoder and extract sequences of discrete speech units. For text LM training, we train a separate BPE tokenizer with a vocabulary of 4{,}096 tokens on the ASR transcriptions.

\subsection{Model Architecture and Design Choices}
\label{app:ssec:modeling}

\paragraph{Speech encoder.}
All encoders used in our experiments, including HuBERT, DinoSR, SpidR and SpidR 2.0,
use the same Transformer backbone architecture,
based on \textsc{Base} wav2vec~2.0~\citep{baevski2020wav2vec}.
The Transformer configuration is summarized in Table~\ref{tab:transformer_params}.
A convolutional feature extractor downsamples the 16\,kHz input waveform to a sequence of 50\,Hz features,
which are processed by the Transformer layers.
DinoSR and SpidR associate each of the top eight teacher layers with an online codebook containing 256 codewords.
SpidR 2.0 replaces the separate online codebooks with layer-specific heads shared between the student and teacher at the same eight layers,
each producing a softmax distribution over 256 codes.

\begin{table}[t]
    \centering
    \caption{Transformer backbone configurations for the speech encoders and spoken/text LMs. Scale labels denote nominal backbone sizes, and Table~\ref{tab:model_params} reports the exact total parameter counts for the LMs.}
    \label{tab:transformer_params}
    
    \small
    \renewcommand{\arraystretch}{1.12}
    
    \begin{tabular}{lcccccc}
        \toprule
        & \multicolumn{1}{c}{\textbf{Encoder}}
        & \multicolumn{5}{c}{\textbf{Language model}} \\
        \cmidrule(l{0.3em}r{0.3em}){2-2}
        \cmidrule(l{0.3em}r{0.3em}){3-7}
        \textbf{Hyperparameter}
        & 90M
        & 90M
        & 300M
        & 1.2B
        & 2.5B
        & 6.5B \\
        \midrule
    
        Transformer layers
        & 12 & 12 & 24 & 24 & 32 & 32 \\
    
        Hidden dimension
        & 768 & 768 & 1,024 & 2,048 & 2,560 & 4,096 \\
    
        FFN dimension
        & 3,072 & 3,072 & 4,096 & 8,192 & 10,240 & 16,384 \\
    
        Attention heads
        & 12 & 12 & 16 & 32 & 32 & 32 \\
        \bottomrule
    \end{tabular}
\end{table}

\begin{table}[t]
    \centering
    \caption{Total parameter counts for the LMs in the main comparison (Table~\ref{tab:main_results} and Figure~\ref{fig:scaling}). Each count includes the Transformer backbone and its associated input and output modules.}
    \label{tab:model_params}
    
    \small
    \renewcommand{\arraystretch}{1.12}

    \begin{tabular}{lccccc}
    \toprule
    & \multicolumn{5}{c}{\textbf{LM backbone scale}} \\
    \cmidrule(l{0.3em}r{0.3em}){2-6} \textbf{Model} & 90M & 300M & 1.2B & 2.5B & 6.5B \\
    \midrule
    Text LM & 92.9M & 312.8M & 1.23B & 2.54B & 6.49B \\
    SpidR SLM & 87.0M & 304.9M & 1.21B & 2.52B & 6.45B \\
    SpidR 2.0 SLM & 87.0M & 304.9M & 1.21B & 2.52B & 6.45B \\
    GSLM2 & 96.3M & 315.5M & 1.23B & 2.54B & 6.48B \\
    \bottomrule
    \end{tabular}
\end{table}

\paragraph{Language model.}
We instantiate the SLMs and text LMs at several parameter scales using decoder-only Transformer configurations based on OPT~\citep{zhang2022opt},
as summarized in Table~\ref{tab:transformer_params}.
We retain the standard OPT architecture to facilitate controlled comparisons across model scales and with prior SLM work.\footnote{
In preliminary experiments, replacing OPT components with more modern alternatives,
such as rotary positional embeddings and RMS normalization,
did not yield consistent improvements.
We therefore use the basic OPT-style architecture in all reported experiments.
}
The models are implemented using the official SLM training and evaluation pipeline released with SpidR.\footnote{\url{https://github.com/facebookresearch/spoken-lm}}

For GSLM2, the temporal Transformer uses the same backbone configuration as the single-stream SLMs.
The generative head is implemented as a six-layer depth Transformer with a hidden dimension of 256, four attention heads, and a feed-forward network (FFN) dimension of 1,024.
The generative head contains approximately 6M parameters and uses the same configuration at every backbone scale.
GSLM2 additionally introduces stream-specific input embeddings and output heads,
whose parameter counts grow slightly with the hidden dimension of the temporal backbone.
Table~\ref{tab:model_params} reports the total parameter counts for the models compared in Table~\ref{tab:main_results} and Figure~\ref{fig:scaling}.

\paragraph{Duration predictor.}
Because the SLMs operate on deduplicated unit sequences,
we train a separate duration predictor to restore the number of 20\,ms frames associated with each generated unit.
Each input unit is mapped to a learned embedding and augmented with sinusoidal positional encoding.
The resulting sequence is processed by a non-autoregressive Conformer~\citep{gulati2020conformer} followed by a two-layer classification head over 50 duration classes, corresponding to durations of 1-50 frames.
For single-stream SLMs, the Conformer has six layers, a hidden dimension of 128, four attention heads, a convolution kernel size of 17, and an FFN dimension of 512.
For GSLM2, we increase its capacity to eight layers, a hidden dimension of 192, and an FFN dimension of 768,
while this larger configuration was ineffective for single-stream SLMs.
The per-stream embeddings are concatenated and linearly projected to the 192-dimensional Conformer input. 
We then predict a single duration for each cross-stream-deduplicated tuple and share it across all streams.

\paragraph{Unit vocoder.}
To convert SLM-generated unit sequences into audio,
we use a HiFi-GAN unit vocoder~\citep{kong2020hifi,polyak2021speech}
that maps discrete units at 20\,ms frame intervals to a 16\,kHz waveform.
For single-stream models, each unit is mapped to a learned 128-dimensional embedding,
and the resulting sequence conditions the HiFi-GAN generator.
The generator upsamples this sequence through a cascade of transposed convolutions with strides $(5, 4, 4, 2, 2)$,
yielding a total upsampling factor of 320.
This matches the 320 samples corresponding to each 20\,ms frame at 16\,kHz.
Each transposed convolution is followed by a multi-receptive-field fusion module with kernel sizes $(3, 7, 11)$ and dilations $(1, 3, 5)$.
For GSLM2, the vocoder is jointly conditioned on all streams:
each stream has its own 64-dimensional unit embedding,
and the per-stream embeddings are concatenated and linearly projected to the 128-dimensional generator input.

\subsection{Training Configurations}
\label{app:ssec:training}

\paragraph{Speech encoder.}
Unless otherwise specified, we pretrain SpidR and SpidR 2.0.
Pretraining largely follows the protocols established for DinoSR~\citep{liu2023dinosr} and SpidR~\citep{poli2025spidr}.
We train each encoder on LibriSpeech for 400k steps using 16 NVIDIA H200 GPUs and a global batch size corresponding to 63 minutes of audio.
Training takes approximately 18 hours.
Optimization uses AdamW with $(\beta_1,\beta_2)=(0.9,0.95)$ and a weight decay of 0.01.
The learning rate follows a three-stage schedule consisting of a 12k-step warmup, a 188k-step constant phase, and a 200k-step decay.
During warmup, the learning rate increases from $5\times10^{-6}$ to $5\times10^{-4}$ for SpidR,
and from $2\times10^{-6}$ to $2\times10^{-4}$ for SpidR 2.0.
We freeze the convolutional feature extractor and the teacher encoder after 200k steps.
For SpidR 2.0, the conditional-entropy weight is fixed at $\mu=0.1$,
while the marginal-entropy weight is initialized at $\lambda=0.1$ and reduced to $0.001$ after 20k steps.
This regularization setup is selected to maximize development-set performance
while minimizing bitrate, as detailed in Appendix~\ref{app:sssec:regularization}.
For encoders trained with this setup,
we use the final checkpoint at step 400k for evaluation and unit extraction.
We also use the official DinoSR\footnote{\url{https://github.com/Alexander-H-Liu/dinosr}} and SpidR\footnote{\url{https://github.com/facebookresearch/spidr}} checkpoints,
along with the pretrained HuBERT \textsc{Base} model available through TorchAudio~\citep{hwang2023torchaudio}.
All three downloaded models are pretrained on LibriSpeech and use similar architectures.

\paragraph{Language model.}
\begin{table}[t]
    \centering
    \caption{Training configurations for text LMs, single-stream SLMs, and GSLM2 across data and model scales. Parenthesized values correspond to the parenthesized model size.}
    \label{tab:lm_training_params}

    \scriptsize
    \renewcommand{\arraystretch}{1.12}

    \begin{tabular}{rlcccccc}
    \toprule
    \textbf{Data} & \textbf{Model} & \textbf{\#Params} & \textbf{\shortstack{Peak\\learning rate}} & \textbf{Steps} & \textbf{\shortstack{Global batch\\(units/tokens)}} & \textbf{\shortstack{Context\\length}} & \textbf{\#GPUs} \\
    \midrule

    \multirow{3}{*}[0pt]{600\,h}
    & Text LM & 90M & $10^{-2}$ & 10k & $81{,}920$ & $2{,}048$ & 8 DDP \\
    & SLM & 90M & $10^{-2}$ & 15k & $655{,}360$ & $2{,}048$ & 8 DDP \\
    & GSLM2 & 90M & $10^{-2}$ & 15k & $1{,}310{,}720$ & $4{,}096$ & 32 DDP \\
    \cmidrule(l{0.2em}r{0.2em}){1-8}

    \multirow{3}{*}[0pt]{6K\,h}
    & Text LM & 300M & $5.0\!\times\!10^{-3}$ & 20k & $491{,}520$ & $2{,}048$ & 8 DDP \\
    & SLM & 300M (90M) & $5.0\!\times\!10^{-3}\;(10^{-2})$ & 25k & $655{,}360$  & $2{,}048$ & 8 DDP \\
    & GSLM2 & 300M (90M) & $5.0\!\times\!10^{-3}\;(10^{-2})$ & 25k & $1{,}310{,}720$ & $4{,}096$ & 32 DDP \\
    \cmidrule(l{0.2em}r{0.2em}){1-8}

    \multirow{3}{*}[0pt]{60K\,h}
    & Text LM & 1.2B & $10^{-3}$ & 100k & $491{,}520$ & $2{,}048$ & 16 DDP \\
    & SLM & 1.2B & $10^{-3}$ & 150k & $655{,}360$ & $2{,}048$ & 16 DDP \\
    & GSLM2 & 1.2B & $10^{-3}$ & 150k & $1{,}310{,}720$ & $4{,}096$ & 64 DDP \\
    \cmidrule(l{0.2em}r{0.2em}){1-8}

    \multirow{3}{*}[0pt]{600K\,h}
    & Text LM & 2.5B & $8.0 \times 10^{-4}$ & 250k & $688{,}128$ & $2{,}048$ & 32 FSDP \\
    & SLM & 2.5B & $8.0 \times 10^{-4}$ & 350k & $917{,}504$ & $2{,}048$ & 32 FSDP \\
    & GSLM2 & 2.5B & $8.0 \times 10^{-4}$ & 350k & $1{,}835{,}008$ & $4{,}096$ & 64 FSDP \\
    \cmidrule(l{0.2em}r{0.2em}){1-8}

    \multirow{3}{*}[0pt]{6M\,h}
    & Text LM & 6.5B & $3.0 \times 10^{-4}$ & 600k & $688{,}128$ & $2{,}048$ & 32 FSDP \\
    & SLM & 6.5B & $3.0 \times 10^{-4}$ & 750k & $917{,}504$ & $2{,}048$ & 32 FSDP \\
    & GSLM2 & 6.5B & $3.0 \times 10^{-4}$ & 750k & $1{,}835{,}008$ & $4{,}096$ & 112 FSDP \\
    \bottomrule
    \end{tabular}
\end{table}

\begin{table}[t]
    \centering
    \caption{Training and inference compute across model scales. Training FLOPs include both forward and backward passes, whereas inference FLOPs correspond to a forward pass per second of audio. Inference FLOPs of text LMs assume 4.45 BPE tokens/s, measured on Libriheavy transcriptions.}
    \label{tab:flops}

    \small
    \renewcommand{\arraystretch}{1.12}

    \begin{tabular}{llcc}
    \toprule
    & & \multicolumn{2}{c}{\textbf{Compute (FLOPs)}} \\
    \cmidrule(l{0.3em}r{0.3em}){3-4}
    \textbf{Scale} & \textbf{Model} & Training & Inference / audio s \\
    \midrule

    \multirow{3}{*}[0pt]{600\,h / 90M}
    & Text LM & $5.26\times10^{17}$ & $9.52\times10^{8}$ \\
    & SpidR 2.0 SLM & $6.13\times10^{18}$ & $3.22\times10^{9}$ \\
    & GSLM2 & $2.34\times10^{19}$ & $1.11\times10^{10}$ \\
    \cmidrule(l{0.2em}r{0.2em}){1-4}

    \multirow{3}{*}[0pt]{6K\,h / 300M}
    & Text LM & $2.10\times10^{19}$ & $3.17\times10^{9}$ \\
    & SpidR 2.0 SLM & $3.47\times10^{19}$ & $1.09\times10^{10}$ \\
    & GSLM2 & $1.16\times10^{20}$ & $3.14\times10^{10}$ \\
    \cmidrule(l{0.2em}r{0.2em}){1-4}

    \multirow{3}{*}[0pt]{60K\,h / 1.2B}
    & Text LM & $3.88\times10^{20}$ & $1.17\times10^{10}$ \\
    & SpidR 2.0 SLM & $7.72\times10^{20}$ & $4.06\times10^{10}$ \\
    & GSLM2 & $2.39\times10^{21}$ & $1.04\times10^{11}$ \\
    \cmidrule(l{0.2em}r{0.2em}){1-4}

    \multirow{3}{*}[0pt]{600K\,h / 2.5B}
    & Text LM & $2.78\times10^{21}$ & $2.40\times10^{10}$ \\
    & SpidR 2.0 SLM & $5.17\times10^{21}$ & $8.32\times10^{10}$ \\
    & GSLM2 & $1.57\times10^{22}$ & $2.09\times10^{11}$ \\
    \cmidrule(l{0.2em}r{0.2em}){1-4}

    \multirow{3}{*}[0pt]{6M\,h / 6.5B}
    & Text LM & $1.67\times10^{22}$ & $5.99\times10^{10}$ \\
    & SpidR 2.0 SLM & $2.77\times10^{22}$ & $2.08\times10^{11}$ \\
    & GSLM2 & $8.28\times10^{22}$ & $5.08\times10^{11}$ \\
    \bottomrule
    \end{tabular}
\end{table}

We train SLMs and text LMs across multiple model scales and training-data sizes.
Table~\ref{tab:lm_training_params} summarizes the key configurations for each setting,
including model size, peak learning rate, total training steps, global batch size, context length, number of NVIDIA H200 GPUs, and parallelization strategy.
In the table, SLM denotes a single-stream model trained on units extracted by HuBERT, DinoSR, SpidR, or SpidR 2.0.
Table~\ref{tab:flops} reports the training and inference compute requirements for the text LM, SpidR 2.0-based SLM, and GSLM2 at each model scale.

The HuBERT- and SpidR-based SLMs use units obtained by applying $k$-means quantization to representations from encoder layers~11 and~6, respectively, following~\citet{poli2025spidr}.
In DinoSR, all prediction heads operate on representations from student encoder layer~12,
with each head trained to predict codebook assignments derived from a corresponding teacher layer in the range 5-12.
The DinoSR-based SLM uses units predicted by the head corresponding to teacher layer~5.
The SpidR 2.0-based SLM uses units predicted from layer~7 of the SpidR 2.0 encoder,
whereas GSLM2 uses units predicted from layers~5-10.
Appendix~\ref{app:sssec:layer_selection} describes the layer-selection criteria for the proposed models. 
For GSLM2, streams are generated in ascending order of bitrate,
yielding the layer order $(10,5,6,9,8,7)$ for $\pi$.
Figure~\ref{fig:codebook-layerwise} reports the layer-wise bitrates,
and Appendix~\ref{app:sssec:effect_of_layer_order} compares alternative generation orders.

All models are optimized using AdamW with $(\beta_1,\beta_2)=(0.9,0.95)$.
Weight decay is set to 0.1 for all models except text LMs, which use 0.01.
Models with 90M, 300M, 1.2B, or 2.5B parameters use a cosine-annealing learning-rate schedule with a 1k-step linear warmup.
The 6.5B-parameter SLM instead uses a tri-stage schedule that allocates 10\%, 40\%, 50\% of the total training steps to warmup, hold, and cosine decay, respectively,
corresponding to 75k, 300k, and 375k steps.
The 6.5B-parameter text LM follows the same allocation, with 60k, 240k, and 300k steps for the respective stages.
For GSLM2, we increase the global batch size and context length to account for the reduced compression of cross-stream deduplication.
Because a position is retained whenever any stream changes,
individual streams may contain consecutive repeated units.
With five or more streams, the resulting sequences are approximately twice as long as SpidR 2.0 sequences,
as shown in Figure~\ref{fig:gslm2_dev_curve}.
Larger batches and longer contexts provide each stream with comparable coverage of non-repeated tokens. 
GSLM2 retains its performance advantage even when the batch size and context length are matched, as reported in Appendix~\ref{app:sssec:slm_ablation_studies}.
For each language model, we select the checkpoint with the lowest validation loss.

\paragraph{Duration predictor.}
We train the duration predictor on deduplicated unit sequences extracted from the Libri-Light \texttt{unlab-600} subset.
Training runs for 25k steps on a single NVIDIA H100 GPU with a batch size of 64.
Optimization uses AdamW with $(\beta_1,\beta_2)=(0.9,0.999)$ and a weight decay of $10^{-5}$.
The learning rate increases linearly to $3\times10^{-3}$ over the first 1k steps and then decays linearly.
We select the checkpoint with the lowest validation loss.

\paragraph{Unit vocoder.}
We train the unit HiFi-GAN vocoder on non-deduplicated unit sequences extracted from LJ Speech,
following the HiFi-GAN training procedure~\citep{kong2020hifi}.
Training runs up to 2k epochs on randomly sampled 0.56-second segments with a global batch size of 512 across eight NVIDIA H100 GPUs.
Optimization uses AdamW with $(\beta_1,\beta_2)=(0.8,0.99)$ and a weight decay of 0.01.
The learning rate is initialized to $2\times10^{-4}$ and decays exponentially by a factor of 0.999 after each epoch.
Adversarial training uses a multi-period discriminator with periods $\{2,3,5,7,11\}$ and a multi-scale discriminator operating at three scales,
both trained with a least-squares GAN objective.
The generator objective combines a least-squares adversarial loss, a feature-matching loss, and an $L_1$ log-mel reconstruction loss weighted by 45.
We use the final checkpoint for evaluation.

\subsection{Evaluation Metrics, Protocols, and Configurations}
\label{app:ssec:evaluation}

This section details the metrics and protocols used to evaluate our SLMs and encoder representations.
We first describe the data splits, adaptations, and scoring procedures for three linguistic benchmarks:
ZeroSpeech'21 (sWuggy and sBLIMP; Appendix~\ref{app:sssec:zerospeech21}),
StoryCloze (tSC and sSC; Appendix~\ref{app:sssec:storycloze}),
and LongTail-Swap (Appendix~\ref{app:sssec:longtail_swap}).
We then detail the evaluation of generated continuations using ASR-PPL and LLMJ,
together with the decoding-temperature calibration procedure (Appendices~\ref{app:sssec:asr_ppl} and~\ref{app:sssec:llm_judge}).
Finally, we describe ABX and MAP evaluations for assessing encoder representation quality (Appendix~\ref{app:sssec:abx_map}).

\subsubsection{ZeroSpeech'21}
\label{app:sssec:zerospeech21}
We evaluate lexical and syntactic competence of our SLMs using the scoring-based tasks from
ZeroSpeech'21~\citep{nguyen2020zero}.

\paragraph{Tasks.}
We consider two tasks:
\begin{itemize}[leftmargin=1.5em,topsep=0pt]
\item \textbf{sWuggy (lexical)}:
A spoken spot-the-word task in which each example pairs a real English word with a nonword (generated using Wuggy~\citep{keuleers2010wuggy}) and matched in phonotactics and syllabic structure (e.g., ``blick'' vs.\ ``brick'').
The task comprises \textit{in-vocabulary} and \textit{out-of-vocabulary} subsets,
defined by whether the real word appears in the LibriSpeech training vocabulary.
We report sWuggy-all, the mean accuracy across the two subsets.

\item \textbf{sBLIMP (syntax)}:
The spoken counterpart of BLiMP~\citep{warstadt2020blimp}, designed to evaluate syntactic acceptability.
Each example pairs a grammatical sentence with a minimally different ungrammatical sentence
(e.g., ``dogs eat meat'' vs.\ ``dogs eats meat'').
The task comprises 63 subtasks grouped into 12 linguistic categories.
We report mean accuracy across all sentence pairs.

\end{itemize}

\paragraph{Data splits.}
For both tasks, we use the official development sets for model selection and the official test sets for final evaluation.

\paragraph{Multilingual adaptations.}
For the VoxPopuli experiments in Appendix~\ref{app:sssec:slm_eval_voxpopuli},
we construct French and German sBLIMP from the corresponding multilingual grammaticality pairs from MultiBLiMP~\citep{jumelet2026multiblimp}.
For English, we instead use the original sBLIMP benchmark from ZeroSpeech'21 rather than the English split of MultiBLiMP,
as the two are not identical.

Because some written grammatical contrasts are neutralized in speech,
we apply an audibility filter before synthesis.
Specifically, we prompt Llama 3.1 405B\footnote{\url{https://huggingface.co/meta-llama/Llama-3.1-405B}}
to act as a phonetics expert and identify pairs whose grammatical contrast is not recoverable from speech
(e.g., French ``aurais'' vs.\ ``aurait'', which are homophones).
To validate the filtering procedure,
authors who are native speakers of the respective languages review a random sample of 50 pairs per language.
After filtering, the French and German benchmarks contain 1,464 and 2,249 grammaticality pairs, respectively.
We synthesize the retained pairs using Gemini 2.5 Pro TTS
with six voices:
Aoede, Charon, Fenrir, Kore, Puck, and Zephyr.\footnote{\url{https://ai.google.dev/gemini-api/docs/models/gemini-2.5-pro-preview-tts}}
All parameters are set to their default values except temperature, which we set to 0.0 (i.e., greedy decoding).

\paragraph{Scoring.}
For each stimulus pair,
a prediction is counted as correct if the model assigns a higher score to the real word in sWuggy or to the grammatical sentence in sBLIMP.
We report pairwise accuracy; chance performance is 50\%.
For French and German sBLIMP, we average accuracy across the six voices.

\subsubsection{StoryCloze}
\label{app:sssec:storycloze}
We use the Spoken StoryCloze benchmark developed by~\citet{hassid2023textually},
which is derived from the original Story Cloze Test~\citep{mostafazadeh2016corpus}.
Each example consists of a four-sentence commonsense story and two candidate fifth sentences,
one of which is the reference ending and the other a distractor.

\paragraph{Tasks.}
We consider two variants:
\begin{itemize}[leftmargin=1.5em,topsep=0pt]
\item \textbf{Topic StoryCloze (tSC; semantics)}:
The distractor is a correct ending randomly sampled from a different story.
This contrast primarily tests whether the model recognizes topical and global semantic coherence between a story and its ending.

\item \textbf{Spoken StoryCloze (sSC; semantics)}:
The distractor is the adversarial ending from the original Story Cloze Test.
Because these endings are generally fluent and superficially plausible,
sSC provides a more challenging test of causal, temporal, and discourse-level understanding.
\end{itemize}

\paragraph{Data splits.}
We evaluate both tasks on the 1,871 stories in the official spoken test set.
Because Spoken StoryCloze does not provide an official development split,
we construct a tSC development set for model selection and hyperparameter tuning.
We draw its stories from the ``test'' split of the original textual Story Cloze Test benchmark,
whereas the released spoken test set is derived from the original ``development'' split;
the resulting development and test sets are therefore story-disjoint.

To match the released spoken benchmark,
we synthesize each sentence independently using FastSpeech 2~\citep{ren2021fastspeech}\footnote{\url{https://github.com/facebookresearch/fairseq/tree/main/examples/speech_synthesis}} and a HiFi-GAN vocoder~\citep{kong2020hifi}\footnote{\url{https://github.com/jik876/hifi-gan}} trained on LJ Speech,
then concatenate consecutive sentences with 100\,ms of intervening silence.
Preliminary experiments showed substantial variation in tSC accuracy across independently trained model seeds when only one distractor was used per story.
We therefore sample ten distractors from other stories for each of the 1,871 development stories,
yielding ten correct-incorrect pairs per story and 18,710 pairs in total.
This expanded set provides a more stable validation signal.

\paragraph{Multilingual adaptations.}
For the VoxPopuli experiments in Appendix~\ref{app:sssec:slm_eval_voxpopuli},
we construct French and German adaptations of tSC from the original English text benchmark~\citep{mostafazadeh2016corpus}.
We translate the stories and candidate endings sentence by sentence
using Llama 3.3 70B\footnote{\url{https://huggingface.co/meta-llama/Llama-3.3-70B-Instruct}} with structured prompts.
To ensure that the translated task remains well formed and solvable,
authors who are native speakers of the respective target languages review a random sample of 50 pairs per language,
and we correct any issues they identify.
We synthesize the translated pairs using the same six-voice Gemini 2.5 Pro TTS configuration as for multilingual sBLIMP.

\paragraph{Scoring.}
For each stimulus pair, the model scores the two complete five-sentence stories.
A prediction is counted as correct if the story containing the reference ending receives the higher score.
We report pairwise accuracy; chance performance is 50\%.
For French and German tSC, we average accuracy across the six voices.

\subsubsection{LongTail-Swap}
\label{app:sssec:longtail_swap}
We develop a spoken adaptation of LT-Swap~\citep{algayres2025longtail},
a corpus-dependent benchmark that evaluates linguistic knowledge across the word-frequency distribution.

\paragraph{Tasks.}
LT-Swap comprises three complementary tasks.
Each example is a quadruplet $(S_1,S_2,S_1^{*},S_2^{*})$,
where $S_1$ and $S_2$ contain two target forms in their intended contexts,
and $S_1^{*}$ and $S_2^{*}$ are obtained by exchanging those forms.

\begin{itemize}[leftmargin=1.5em,topsep=0pt]
\item \textbf{WordSwap (WS; semantics)}:
Two target words from the same part of speech and nonzero frequency bin are placed in separate contexts in which they are semantically appropriate.
Exchanging the targets yields sentences that remain syntactically well formed but become semantically anomalous. 
For example,
\begin{flushleft}
\(\begin{aligned}
    &\qquad S_1: \text{``The \textit{cat} is sleeping on the mat.''}
    && S_1^{*}: \text{``The \textit{boat} is sleeping on the mat.''} \\
    &\qquad S_2: \text{``The \textit{boat} is sailing on the sea.''}
    && S_2^{*}: \text{``The \textit{cat} is sailing on the sea.''}
\end{aligned}\)
\end{flushleft}

\item \textbf{InflectionSwap (IS; morphosyntax)}:
Two inflected forms of the same lemma,
such as the bare and present-participle forms of a verb,
are placed in contexts that license their respective forms.
Exchanging them creates morphosyntactic violations while holding the underlying lemma fixed.
For example,
\begin{flushleft}
\(\begin{aligned}
    &\qquad S_1: \text{``He couldn't \textit{sleep} last night.''}
    && S_1^{*}: \text{``He couldn't \textit{sleeping} last night.''} \\
    &\qquad S_2: \text{``The baby was \textit{sleeping} peacefully.''}
    && S_2^{*}: \text{``The baby was \textit{sleep} peacefully.''}
\end{aligned}\)
\end{flushleft}

\item \textbf{AgreementSwap (AS; syntax)}:
Singular and plural target forms are exchanged between contexts,
breaking a targeted agreement relation.
The task covers subject-verb, antecedent-anaphor, and determiner-noun agreement,
with short- and long-distance variants where applicable.
For example, a subject-verb agreement quadruplet is:
\begin{flushleft}
\(\begin{aligned}
    &\qquad S_1: \text{``The \textit{strategist} analyzes.''}
    && S_1^{*}: \text{``The \textit{strategists} analyzes.''} \\
    &\qquad S_2: \text{``The \textit{strategists} analyze.''}
    && S_2^{*}: \text{``The \textit{strategist} analyze.''}
\end{aligned}\)
\end{flushleft}

\end{itemize}

\paragraph{Benchmark construction.}
For each Libri-Light training-data scale,
we compute word frequencies from the corresponding subset of Libriheavy transcriptions.
Following the original benchmark, we retain nouns and verbs and group their surface forms into power-of-two frequency bins:
$\{[0], [1,2), [2,4), [4,8), \ldots, [256,512), [512,\infty)\}$.
The zero-frequency bin, which applies only to IS and AS, contains automatically derived inflected forms that do not occur in the corresponding training corpus.

Following the original LT-Swap pipeline,\footnote{\url{https://github.com/facebookresearch/lt-swap}}
we use the instruction-following Llama 3.3 70B model to generate candidate sentences and filter ambiguous
examples.
A quadruplet is retained only if the filtering model reliably distinguishes
both intended sentences from their corrupted counterparts.
For WS, the two targets are drawn from the same frequency bin.
For IS and AS,
each quadruplet is assigned to the bin of its less frequent target form.
For WS, IS, and AS, respectively, this yields
31,952, 33,218, and 40,928 sentence pairs at 600 hours;
41,418, 69,834, and 73,132 at 6K hours; and
39,166, 105,624, and 99,860 at 60K hours.

To create the spoken benchmark,
we synthesize all four sentences in each quadruplet using Piper TTS\footnote{\url{https://github.com/OHF-Voice/piper1-gpl}} with the Kathleen voice.
Each waveform is tokenized separately by the speech encoder associated with the evaluated SLM,
whereas text LMs are evaluated on the corresponding text.

\paragraph{Scoring.}
Each quadruplet yields two binary comparisons:
$S_1$ vs.\ $S_1^{*}$ and\ $S_2$ vs.\ $S_2^{*}$.
A comparison is counted as correct if the intended sentence receives a higher score than its corrupted counterpart.
We compute accuracy separately for each task and word-frequency bin;
chance performance is 50\%.
For each task,
Table~\ref{tab:main_results} reports the macro-average across its applicable frequency bins.
A breakdown by word-frequency bin is provided in Appendix~\ref{app:sssec:lt_swap}.

\subsubsection{ASR-PPL}
\label{app:sssec:asr_ppl}
We assess the linguistic plausibility of generated speech continuations using the perplexity of their ASR transcripts.
Perplexity alone can be misleading, however, because repetitive, low-temperature generations may achieve artificially favorable scores.
Following the motivation of auto-BLEU~\citep{lakhotia2021generative},
we therefore compare models using a common calibration target for lexical repetition.
For each model, we calibrate the decoding temperature on a development set to target the ground-truth baseline under the combined normalized unigram-bigram repetition score defined below,
and then evaluate the model on the test set using the calibrated temperature.
The complete evaluation protocol is described below.

\paragraph{Speech-continuation data.}
We sample 1,024 LibriSpeech utterances with durations between 6 and 15 seconds.
We report results on utterances from \texttt{test-clean} and calibrate temperatures on an identically constructed set of 1,024 utterances from \texttt{dev-clean}.
Each utterance is divided into a prompt and a continuation at approximately the 3-second mark.
Using forced alignments, we shift the split point to the nearest word boundary to avoid splitting a word.
Note that the transcripts in both \texttt{dev-clean} and \texttt{test-clean} are normalized (i.e., punctuation and casing removed).

\paragraph{Unit generation.}
For each utterance, we encode the prompt with a speech encoder to obtain ground-truth speech units.
Conditioned on these units,
each SLM autoregressively generates a continuation of up to 10 seconds.
We use nucleus sampling with $p=0.9$ and calibrate the temperature separately for each model.
For GSLM2, the generative head predicts the streams autoregressively within each time step.
To mitigate the accumulation of sampling noise across successive stream predictions,
we linearly anneal the temperature from its calibrated value for the first stream
to 50\% of that value for the final stream,
resetting the schedule at every time step.
Appendix~\ref{app:sssec:slm_ablation_studies} evaluates this annealing strategy.
Text LMs are conditioned on the reference transcript up to the same prompt boundary
and generate 48 tokens using the same nucleus-sampling parameter and model-specific calibrated temperatures.
This token budget yields approximately 30 words on average,
closely matching the average word count of the SLM continuations.

\paragraph{Unit synthesis.}
Generated units are converted to waveforms using the vocoder corresponding to their speech-unit space.
For deduplicated unit streams,
a separately trained duration predictor restores the temporal information removed by deduplication by predicting the number of frames associated with each generated unit.
The generated continuation is then expanded to the original frame rate,
while the prompt units preserve their original frame-level durations.
The complete prompt-continuation sequence is synthesized using the unit HiFi-GAN vocoder trained on the corresponding unit representation.

\paragraph{ASR and text normalization.}
We split each synthesized waveform at the known prompt boundary and transcribe only the generated continuation using Qwen3-Omni\footnote{\url{https://huggingface.co/Qwen/Qwen3-Omni-30B-A3B-Instruct}}.
For ASR-PPL, we condition on the ground-truth prompt transcript
and normalize both prompt and continuation by lowercasing, removing punctuation, expanding contractions, and spelling out numerals.
For LLMJ, we instead use a semantics-preserving normalization that lowercases text while retaining apostrophes and sentence-boundary punctuation.
For each evaluation,
we apply the corresponding task-specific normalization to both ASR transcripts and direct text-LM outputs,
making the textual representations of speech and text systems as comparable as possible.

\paragraph{Conditional perplexity.}
We compute ASR-PPL as the conditional perplexity of the normalized ASR transcript of each continuation,
conditioned on the normalized ground-truth prompt.
Let $W^{\mathsf{p}}$ and $W^{\mathsf{c}}$ denote the corresponding prompt and continuation token sequences, respectively.
Both sequences are tokenized using the tokenizer associated with Llama-3.2-1B~\citep{grattafiori2024llama}\footnote{\url{https://huggingface.co/meta-llama/Llama-3.2-1B}}.
Frozen Llama-3.2-1B scores each continuation conditioned on its prompt as
\begin{equation}
    \mathrm{ASR\text{-}PPL}
    \left(W^{\mathsf{c}} \mid W^{\mathsf{p}}\right)
    =
    \exp
    \left[
        -\frac{1}{\lvert W^{\mathsf{c}} \rvert}
        \sum_{u=1}^{\lvert W^{\mathsf{c}} \rvert}
        \log p_{\mathrm{LLM}}
        \left(
            w^{\mathsf{c}}_u \mid W^{\mathsf{p}}, w^{\mathsf{c}}_{<u}
        \right)
    \right].
\end{equation}
The prompt is provided only as left context and is excluded from the loss,
so ASR-PPL is computed solely over continuation tokens.

\paragraph{Temperature calibration.}
Decoding temperature controls the trade-off between generation quality and diversity.
Evaluating all models at a common temperature could therefore compare systems operating at substantially different levels of lexical repetition.
We instead calibrate each model's decoding temperature, for both SLMs and Text LMs,
to target the ground-truth repetition baseline.
Appendix~\ref{app:sssec:gen_eval_fixed_temp} reports additional results under a fixed decoding temperature.

Because auto-BLEU depends on sequence length,
e.g., longer transcripts offer more opportunities for repetition,
we compute it over the first 15 words of each normalized continuation transcript. 
Prior work commonly uses bigram auto-BLEU~\citep{lin2025align,maimon2025slamming},
but we find bigram repetition too sparse for reliable calibration in our setting:
only 3.2\% of ground-truth development continuations contain a repeated bigram.
We therefore combine unigram and bigram repetition,
with the unigram component providing a denser calibration signal.
Let $\mathrm{auto\text{-}BLEU}_{n}(\tau)$ denote the mean $n$-gram auto-BLEU of continuations generated at temperature $\tau$, and
let $\mathrm{auto\text{-}BLEU}_{n}^{\mathsf{gt}}$ denote the corresponding mean for ground-truth continuations.
We define the unigram-bigram normalized repetition score (NRS) as
\begin{equation}
    \mathrm{NRS}(\tau) = \frac{1}{2}
    \left(
        \frac{\mathrm{auto\text{-}BLEU}_{1}(\tau)}{\mathrm{auto\text{-}BLEU}_{1}^{\mathsf{gt}}} +
        \frac{\mathrm{auto\text{-}BLEU}_{2}(\tau)}{\mathrm{auto\text{-}BLEU}_{2}^{\mathsf{gt}}}
    \right).
\end{equation}
Thus, $\mathrm{NRS}=1$ indicates a match to the ground-truth baseline under this combined unigram-bigram repetition measure,
while values above and below one indicate greater and lower repetition according to this measure, respectively.

For each model, we evaluate NRS over a temperature grid with increments of 0.05.
We then linearly interpolate between the two adjacent grid points that bracket $\mathrm{NRS}=1$ to obtain the calibrated temperature $\hat{\tau}$ at a resolution of 0.01.
The resulting development-set calibration curves are shown in Appendix~\ref{app:sssec:calibration_curves}.
No test-set statistics are used for calibration.
Once selected, the model-specific temperatures are fixed and used unchanged for test-set generation.

For both development and test sets, we evaluate each model over ten runs.
Within each run, we summarize ASR-PPL by the median and NRS by the mean across examples.
We then report the mean and standard deviation of these per-run statistics over the ten runs,
as in Table~\ref{tab:gen_eval}.

\subsubsection{LLM-Judge}
\label{app:sssec:llm_judge}
We use Prometheus 2~\citep{kim2024prometheus} as an LLM-based judge to evaluate generated continuations.
The grading prompt below follows the official absolute-grading format\footnote{\url{https://huggingface.co/prometheus-eval/prometheus-8x7b-v2.0}},
with adaptations for open-ended continuation evaluation.
\begin{tcolorbox}
\scriptsize
\#\#\#Task Description:\\
An instruction (might include an Input inside it), a response to evaluate, a reference answer that gets a score of 5, and a score rubric representing a evaluation criteria are given.\\
\\
1.\hspace{0.5em}Write a detailed feedback that assess the quality of the response strictly based on the given score rubric, not evaluating in general.\\
2.\hspace{0.5em}After writing a feedback, write a score that is an integer between 1 and 5. You should refer to the score rubric.\\
3.\hspace{0.5em}The output format should look as follows: ``Feedback: (write feedback for the criteria) [RESULT] (an integer number between 1 and 5)''\\
4.\hspace{0.5em}Please do not generate any other opening, closing, and explanations.\\
\\
\#\#\#The instruction to evaluate:\\
Evaluate the passage below. $<$CONTINUATION\_START$>$ separates the given prompt from the generated continuation. Score only the continuation, but read both sides together. The boundary may be inside a sentence, name, quotation, or question, and the continuation may end at a generation limit.\\
The prompt and response use the same semantic normalization: lowercase text with apostrophes and sentence boundaries preserved. They may contain minor rendering or transcription artifacts. Judge the intended content when it remains understandable.\\
\\
\#\#\#Response to evaluate:\\
\{prompt\_transcript\}$<$CONTINUATION\_START$>$\{continuation\_transcript\}\\
\\
\#\#\#Reference Answer (Score 5):\\
Any natural, coherent, non-degenerate continuation that fits the marked boundary. Multiple answers
are valid; it need not match a particular event, add detail, or finish the passage.\\
\\
\#\#\#Score Rubrics:\\{}
[Semantic continuation quality: Judge boundary fit, internal coherence, and non-degeneration from
the actual words. Missing boundary punctuation and newly introduced entities are not errors when
the connection is clear. Ordinary word reuse is cohesion; claim copying only for an actual repeated
multiword phrase. Forgive isolated understandable transcription artifacts and ignore only the final
unfinished fragment. Do not invent missing links or let a coherent opening excuse later unrelated,
contradictory, repetitive, or uninterpretable text. Choose the highest fully satisfied score. Give
at most two feedback sentences and 60 words, cite the strongest evidence, and end with [RESULT] N.]\\
Score\hspace{0.25em}1:\hspace{0.5em}No connected meaning is recoverable, or most of the continuation is unrelated, contradictory, uninterpretable, copied, boilerplate, or looping.\\
Score\hspace{0.25em}2:\hspace{0.5em}A local connection exists, but major or repeated failures make the overall meaning difficult to recover.\\
Score\hspace{0.25em}3:\hspace{0.5em}The main meaning is recoverable and mostly connected, but a substantive defect or several local errors materially weaken it.\\
Score\hspace{0.25em}4:\hspace{0.5em}The continuation fits and remains coherent, with only localized minor errors and no substantive defect.\\
Score\hspace{0.25em}5:\hspace{0.5em}Every complete span fits, is coherent and natural, and has no identified semantic defect.\\
\\
\#\#\#Feedback:
\end{tcolorbox}

For SLMs, we reuse the ASR transcripts generated for the ASR-PPL evaluation;
for text LMs, we evaluate the generated text directly.
We apply the same semantic normalization to both modalities, preserving apostrophes and sentence boundaries.
The prompt and continuation are presented as a single passage separated by the explicit boundary marker ``$<$CONTINUATION\_START$>$'',
and only the continuation is scored.
Prometheus 2 assigns an integer score from 1 to 5 based on boundary compatibility, internal coherence, and non-degeneration.
The rubric allows the coherent introduction of new entities and ordinary lexical reuse,
while penalizing contradictions, unrelated content, multiword phrase copying, and repetitive loops.
It also tolerates isolated, intelligible transcription errors and disregards an unfinished final fragment caused by the generation limit.
We evaluate each model using the temperature calibrated during the ASR-PPL evaluation.
We conduct ten runs for each system and report the mean and standard deviation,
as in Table~\ref{tab:gen_eval}.

\subsubsection{ABX and MAP}
\label{app:sssec:abx_map}
We compute ABX~\citep{schatz2016abx} and MAP~\citep{carlin2011rapid} scores on both the \texttt{dev-clean} and \texttt{dev-other} subsets of the ZeroSpeech'21 development set~\citep{nguyen2020zero},
using the provided phoneme- and word-level alignments, respectively.

ABX is a discrimination task involving a triplet of representations $(a,b,x)$,
where $a$ and $b$ belong to known classes $A$ and $B$, respectively,
and the class of $x$ is unknown but belongs to $\{A,B\}$.
$x$ is assigned to class $A$ if $d(x,a) < d(x,b)$ and to class $B$ if $d(x,b) < d(x,a)$.
We report the error rate for discriminating between triphone classes,
where a triphone is a sequence of three consecutive phonemes.
We evaluate both within-speaker and across-speaker conditions.
In the within-speaker condition, all three tokens are produced by the same speaker. 
In the across-speaker condition, $a$ and $b$ are produced by the same speaker, 
while $x$ is produced by a different speaker.
For continuous ABX, we measure local distances between continuous representations using cosine distance.
For discrete ABX, we represent the argmax code assignments as one-hot vectors and use cosine distance.
In both cases, we aggregate local distances into sequence-level distances using dynamic time warping.
We use the fastabx library~\citep{poli2025fastabx} for scoring,
with chance performance corresponding to a 50\% error rate.

For MAP, we represent each word token by a single embedding obtained by mean-pooling frame representations over its aligned word segment.
Word classes correspond to orthographic word types from the forced alignments.
MAP evaluates a retrieval task: given any query word token $q$ of class $W$,
all other word tokens in the corpus are ranked by cosine similarity $\mathrm{cos}(q,\cdot)$.
Over this ranking,
we compute average precision as 
$\mathrm{AP}(q) = \frac{1}{R}\sum_{k=1}^N \mathrm{Precision}@k\,\mathrm{rel}_k$,
where $R$ is the number of other tokens of class $W$,
$N$ is the total number of retrieval candidates,
$\mathrm{Precision}@k$ is the proportion of relevant tokens among the top $k$ results,
and $\mathrm{rel}_k$ equals 1 if the token at rank $k$ belongs to class $W$ and 0 otherwise.
MAP is the mean of $\mathrm{AP}(q)$ over all query tokens. 
We use the speech-map library~\citep{algayres2020evaluating} for scoring,
with a chance MAP score of 0.56\%.

\section{Additional Experimental Results}
\label{app:additional_experiments}
This section presents additional experimental results that provide further insight into the design of GSLM2 and complement the main findings in Section~\ref{sec:experiments}.
The remainder of the section is organized as follows:
\begin{itemize}[leftmargin=1.5em]
    \item Appendix~\ref{app:ssec:supplementary_slm_evaluation}: Supplementary SLM evaluation
    \item Appendix~\ref{app:ssec:supplementary_encoder_evaluation}: Supplementary encoder evaluation
    \item Appendix~\ref{app:ssec:speech_resynthesis}: Speech resynthesis evaluation
    \item Appendix~\ref{app:ssec:statistical_study}: Statistical significance analysis
\end{itemize}

\subsection{Supplementary SLM Evaluation}
\label{app:ssec:supplementary_slm_evaluation}

This section supplements the main SLM results in Section~\ref{ssec:slm_eval} with additional experiments and analyses.
We first describe how we select encoder layers and the number of streams (Appendix~\ref{app:sssec:layer_selection}),
and estimate scaling efficiency relative to text LMs (Appendix~\ref{app:sssec:scaling_exponents}).
We then examine how decoding temperature and stream generation order affect continuation quality and repetition (Appendices~\ref{app:sssec:calibration_curves} and~\ref{app:sssec:effect_of_layer_order}).
We assess the contributions of self-distillation and single-stream supervision through ablations (Appendix~\ref{app:sssec:slm_ablation_studies}),
compare scoring heads and stream aggregation strategies (Appendix~\ref{app:sssec:scoring_choices}),
and test whether GSLM2 benefits from complementary information across encoder layers (Appendix~\ref{app:sssec:layer_diversity}).
Further evaluations compare GSLM2 with speech-capable LLMs (Appendix~\ref{app:sssec:slm_eval_speech_llm}),
analyze LT-Swap performance by word frequency (Appendix~\ref{app:sssec:lt_swap}),
and assess the robustness of generation rankings to decoding temperature (Appendix~\ref{app:sssec:gen_eval_fixed_temp}).
Finally, we report results for English, French, and German models trained on VoxPopuli (Appendix~\ref{app:sssec:slm_eval_voxpopuli})
and compare direct speech modeling with cascaded ASR-LM systems to examine
how transcription affects linguistic benchmark scores (Appendix~\ref{app:sssec:cascaded_asr_slm}).

\subsubsection{Layer Selection Strategy}
\label{app:sssec:layer_selection}
\begin{figure}[t]
\centering
\begin{minipage}[t]{0.43\linewidth}
  \centering
  \begin{tikzpicture}
\definecolor{spidr2blue}{RGB}{78,121,167}

\pgfdeclareplotmark{filled star}{%
  \pgfpathmoveto{\pgfqpoint{0pt}{\pgfplotmarksize}}%
  \pgfpathlineto{\pgfqpoint{.224514\pgfplotmarksize}{.309017\pgfplotmarksize}}%
  \pgfpathlineto{\pgfqpoint{.951057\pgfplotmarksize}{.309017\pgfplotmarksize}}%
  \pgfpathlineto{\pgfqpoint{.363271\pgfplotmarksize}{-.118034\pgfplotmarksize}}%
  \pgfpathlineto{\pgfqpoint{.587785\pgfplotmarksize}{-.809017\pgfplotmarksize}}%
  \pgfpathlineto{\pgfqpoint{0pt}{-.381966\pgfplotmarksize}}%
  \pgfpathlineto{\pgfqpoint{-.587785\pgfplotmarksize}{-.809017\pgfplotmarksize}}%
  \pgfpathlineto{\pgfqpoint{-.363271\pgfplotmarksize}{-.118034\pgfplotmarksize}}%
  \pgfpathlineto{\pgfqpoint{-.951057\pgfplotmarksize}{.309017\pgfplotmarksize}}%
  \pgfpathlineto{\pgfqpoint{-.224514\pgfplotmarksize}{.309017\pgfplotmarksize}}%
  \pgfpathclose
  \pgfusepath{fill,stroke}%
}

\pgfplotsset{
result axis/.style={
  width=5cm,
  height=3cm,
  scale only axis,
  xmin=4.7,
  xmax=12.3,
  ymin=60,
  ymax=76,
  xtick={5,6,7,8,9,10,11,12},
  ytick={60,65,70,75},
  minor ytick={62.5,67.5,72.5},
  xlabel={\small Layer Index},
  ylabel={\small Accuracy (\%)},
  xticklabel style={font=\scriptsize},
  yticklabel style={font=\scriptsize,xshift=2pt},
  xlabel style={font=\scriptsize,yshift=1pt},
  ylabel style={font=\scriptsize},
  ylabel style={
    font=\scriptsize,
    yshift=-2pt
  },
  tick align=outside,
  tick style={black!70},
  axis lines=left,
  axis line style={black!80,line width=0.9pt},
  xmajorgrids=true,
  ymajorgrids=true,
  yminorgrids=true,
  major grid style={
    draw=black!40,
    dash pattern=on 1pt off 1.4pt,
    line width=0.3pt
  },
  minor grid style={
    draw=black!30,
    dash pattern=on 1pt off 1.4pt,
    line width=0.3pt
  },
  legend columns=2,
  legend cell align=left,
  legend style={
    at={(0.5,1.05)},
    anchor=south,
    font=\scriptsize,
    draw=black!50,
    rounded corners=2pt,
    line width=0.8pt,
    fill=none,
    column sep=4pt
  },
  clip=false
},
major result/.style={
  color=spidr2blue,
  line width=1.2pt,
  mark=filled star,
  mark size=2.0pt,
  mark options={solid,sharp corners,fill=spidr2blue,draw=spidr2blue}
},
gray result 1/.style={
  color=black!55,
  line width=0.8pt,
  densely dashed,
  mark=triangle*,
  mark size=2.0pt,
  mark options={solid,fill=black!55,draw=black!55}
},
gray result 2/.style={
  color=black!42,
  line width=0.8pt,
  dashdotted,
  mark=square*,
  mark size=1.5pt,
  mark options={solid,fill=black!42,draw=black!42}
},
gray result 3/.style={
  color=black!30,
  line width=0.8pt,
  densely dotted,
  mark=diamond*,
  mark size=2.0pt,
  mark options={solid,fill=black!30,draw=black!30}
}
}

\begin{axis}[result axis]
\addlegendimage{major result}
\addlegendentry{Average}
\addlegendimage{gray result 1}
\addlegendentry{sWuggy}
\addlegendimage{gray result 2}
\addlegendentry{sBLIMP}
\addlegendimage{gray result 3}
\addlegendentry{tSC}

\addplot[gray result 1,forget plot] coordinates {
  (5,71.83) (6,72.00) (7,73.18) (8,73.64)
  (9,71.24) (10,70.45) (11,67.34) (12,64.75)
};

\addplot[gray result 2,forget plot] coordinates {
  (5,65.94) (6,65.32) (7,65.55) (8,63.60)
  (9,63.17) (10,62.29) (11,62.40) (12,60.61)
};

\addplot[gray result 3,forget plot] coordinates {
  (5,71.10) (6,72.44) (7,73.67) (8,73.34)
  (9,74.65) (10,73.62) (11,74.85) (12,74.57)
};

\addplot[major result,forget plot] coordinates {
  (5,69.62) (6,69.92) (7,70.80) (8,70.19)
  (9,69.69) (10,68.79) (11,68.20) (12,66.64)
};
\end{axis}
\end{tikzpicture}

\caption{Development-set accuracy of SpidR 2.0 SLM (Libri-Light 6K\,h; 90M parameters) using units extracted from different encoder layers. Average denotes the mean across sWuggy, sBLIMP, and tSC.}
\label{fig:spidr2_slm_dev_curve}
\end{minipage}
\hfill
\begin{minipage}[t]{0.53\linewidth}
  \centering
  \begin{tikzpicture}
\definecolor{majorred}{RGB}{225,87,89}

\pgfdeclareplotmark{filled star}{%
  \pgfpathmoveto{\pgfqpoint{0pt}{\pgfplotmarksize}}%
  \pgfpathlineto{\pgfqpoint{.224514\pgfplotmarksize}{.309017\pgfplotmarksize}}%
  \pgfpathlineto{\pgfqpoint{.951057\pgfplotmarksize}{.309017\pgfplotmarksize}}%
  \pgfpathlineto{\pgfqpoint{.363271\pgfplotmarksize}{-.118034\pgfplotmarksize}}%
  \pgfpathlineto{\pgfqpoint{.587785\pgfplotmarksize}{-.809017\pgfplotmarksize}}%
  \pgfpathlineto{\pgfqpoint{0pt}{-.381966\pgfplotmarksize}}%
  \pgfpathlineto{\pgfqpoint{-.587785\pgfplotmarksize}{-.809017\pgfplotmarksize}}%
  \pgfpathlineto{\pgfqpoint{-.363271\pgfplotmarksize}{-.118034\pgfplotmarksize}}%
  \pgfpathlineto{\pgfqpoint{-.951057\pgfplotmarksize}{.309017\pgfplotmarksize}}%
  \pgfpathlineto{\pgfqpoint{-.224514\pgfplotmarksize}{.309017\pgfplotmarksize}}%
  \pgfpathclose
  \pgfusepath{fill,stroke}%
}

\pgfplotsset{
dual result axis/.style={
  width=5cm,
  height=3cm,
  scale only axis,
  xmin=0.7,
  xmax=8.3,
  xtick={1,2,3,4,5,6,7,8},
  xlabel={\small Number of Streams},
  xticklabel style={font=\scriptsize},
  yticklabel style={font=\scriptsize},
  xlabel style={font=\scriptsize,yshift=1pt},
  ylabel style={font=\scriptsize},
  tick align=outside,
  tick style={black!70},
  axis line style={black!80,line width=0.9pt},
  clip=false
},
accuracy result/.style={
  color=majorred,
  line width=1.2pt,
  mark=filled star,
  mark size=2pt,
  mark options={solid,sharp corners,fill=majorred,draw=majorred}
},
compression result/.style={
  color=black!55,
  line width=0.8pt,
  densely dashed,
  mark=square*,
  mark size=1.5pt,
  mark options={solid,fill=black!55,draw=black!55}
}
}

\begin{axis}[
  dual result axis,
  axis x line*=bottom,
  axis y line*=left,
  ymin=70.5,
  ymax=73.9,
  ytick={71,72,73},
  minor ytick={71.5,72.5,73.5},
  ylabel={\small Accuracy (\%)},
  ylabel style={
    font=\scriptsize,
    yshift=-2pt
  },
  yticklabel style={font=\scriptsize,xshift=2pt},
  xmajorgrids=true,
  ymajorgrids=true,
  yminorgrids=true,
  major grid style={
    draw=black!40,
    dash pattern=on 1pt off 1.4pt,
    line width=0.3pt
  },
  minor grid style={
    draw=black!30,
    dash pattern=on 1pt off 1.4pt,
    line width=0.3pt
  },
  legend columns=2,
  legend cell align=left,
  legend style={
    at={(0.5,1.1)},
    anchor=south,
    font=\scriptsize,
    draw=black!50,
    rounded corners=2pt,
    line width=0.8pt,
    fill=none,
    column sep=4pt
  }
]
\addlegendimage{accuracy result}
\addlegendentry{Accuracy}
\addlegendimage{compression result}
\addlegendentry{Compression Rate}

\addplot[accuracy result,forget plot] coordinates {
  (1,70.80) (2,72.37) (3,72.61) (4,73.13) (5,73.52)
  (6,73.63) (7,73.32) (8,73.16)
};
\end{axis}

\begin{axis}[
  dual result axis,
  axis x line=none,
  axis y line*=right,
  xlabel={},
  ymin=0.5,
  ymax=3.9,
  ytick={1.0,2.0,3.0},
  minor ytick={1.5,2.5,3.5},
  yticklabels={1.0,2.0,3.0},
  ylabel={\small Compression Rate},
  ylabel style={
    font=\scriptsize,
    yshift=2pt
  },
  xtick=\empty,
  xmajorgrids=false,
  ymajorgrids=false,
  yminorgrids=false,
  axis line style={black!55,line width=0.9pt},
  tick style={black!55},
  yticklabel style={font=\scriptsize,text=black!55},
  ylabel style={font=\scriptsize,text=black!55}
]
\addplot[compression result,forget plot] coordinates {
  (1,3.287) (2,2.186) (3,1.790) (4,1.538) (5,1.443)
  (6,1.364) (7,1.304) (8,1.243)
};
\end{axis}
\end{tikzpicture}

\caption{Development-set mean accuracy and compression ratio of GSLM2 (Libri-Light 6K\,h; 90M parameters) for different numbers of streams. Streams are added using encoder layers in descending order of the mean accuracy reported in Figure~\ref{fig:spidr2_slm_dev_curve}.}
\label{fig:gslm2_dev_curve}
\end{minipage}
\end{figure}

We select the encoder layers for SpidR 2.0 SLM and GSLM2 based on aggregate development-set performance,
defined as the mean accuracy across the development splits of sWuggy, sBLIMP, and tSC.

Figure~\ref{fig:spidr2_slm_dev_curve} compares SpidR 2.0 SLM performance using speech units extracted from different encoder layers.
To enable faster iteration,
we conduct these experiments in the 6K-hour Libri-Light setting with the 90M-parameter configuration.
Although the three tasks favor different layers, mean accuracy is highest at layer~7.
We therefore use units extracted from layer~7 for SpidR 2.0 SLM.

Figure~\ref{fig:gslm2_dev_curve} shows the effect of varying the number of streams in GSLM2
under the same experimental setup.
Streams are added in descending order of their mean development-set accuracy in Figure~\ref{fig:spidr2_slm_dev_curve}.
Thus, the single-stream model uses layer~7, the two-stream model combines layers~7 and~8,
and each subsequent model adds the next-best layer.
Accuracy improves as streams are added, peaking at six streams before declining.
We define the compression ratio as the sequence length before (cross-stream) deduplication divided by the sequence length afterward;
thus, higher values indicate greater compression.
The ratio decreases steadily as more streams are added,
indicating that cross-stream deduplication becomes less effective and yields longer sequences.
We use six streams with layers
$\mathcal{J}=\{5,6,\ldots,10\}$,
which achieves the highest development-set accuracy with a compression ratio of 1.36.
However, this layer-selection heuristic is optional, and
GSLM2 can use all eight layers and still substantially outperform the single-stream model.

\subsubsection{Estimating Scaling Efficiency}
\label{app:sssec:scaling_exponents}

\begin{figure*}[t]
\centering
\input{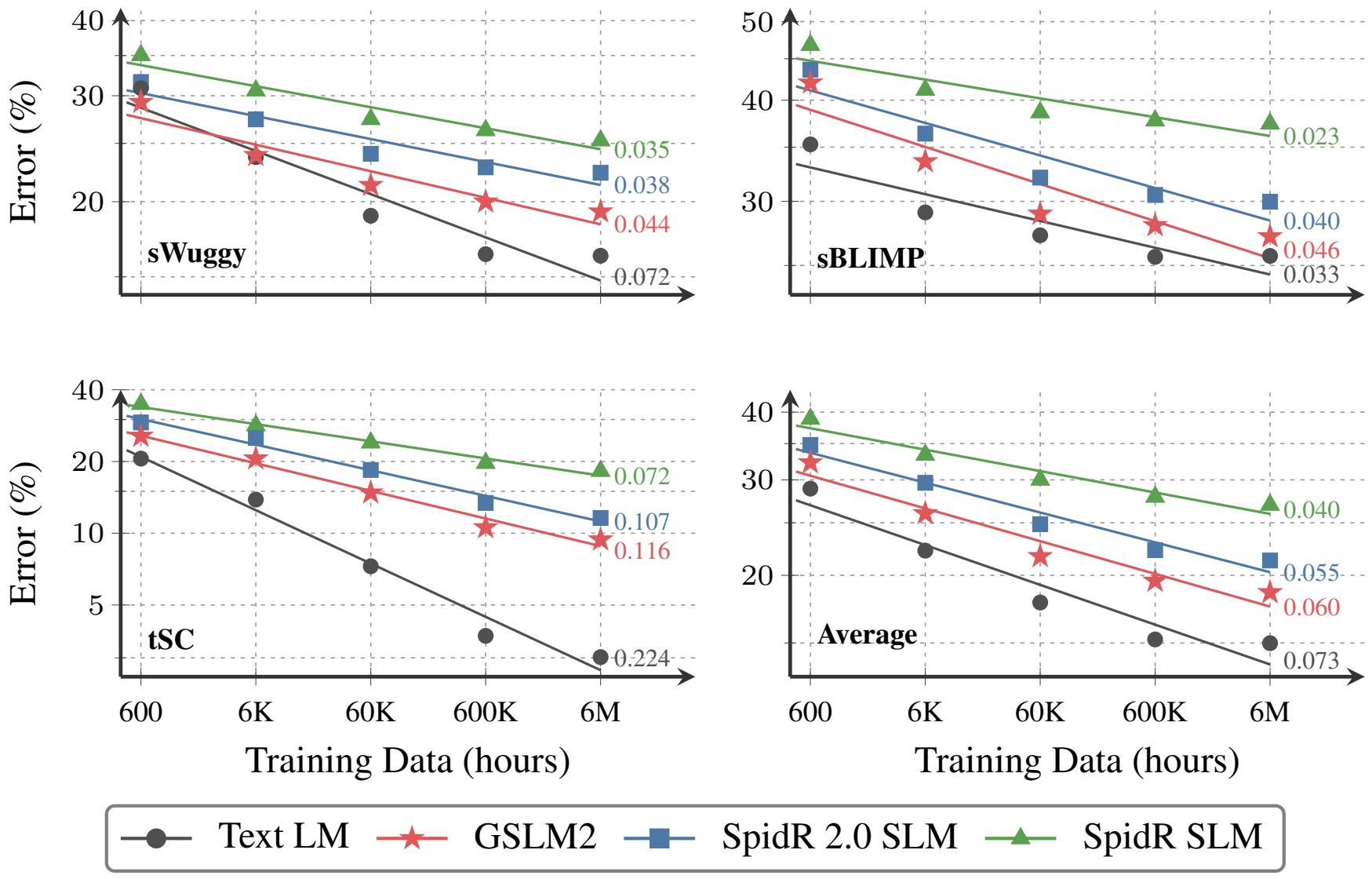}
\caption{Error scaling on conversational speech, both axes logarithmic. Each marker is a trained model; lines are the fitted power law
$\mathrm{error} = a\,N^{-\gamma}$, whose slope is $-\gamma$. Fitted exponents $\gamma$
are given in each panel; a larger $\gamma$ means error falls faster as training
data grows.}
\label{fig:log_error}
\end{figure*}

To estimate scaling efficiency in Figure~\ref{fig:scaling}, we fit the $\mathrm{error}$,
$100 - \mathrm{accuracy}$, as a power law in training hours,
$\mathrm{error} = a\,N^{-\gamma}$. Taking logarithms of both quantities makes
this a straight line of slope $-\gamma$, which we estimate by ordinary least
squares over the five measured scales. Here, $\gamma$ measures error reduction in relative terms---what fraction of the remaining error is removed per unit of data. It therefore does not shrink as the curves flatten near the ceiling (as can be seen in Figure~\ref{fig:scaling}), making it more appropriate than a slope fitted directly to raw accuracy.

Figure~\ref{fig:log_error} shows the resulting fits with both axes logarithmic;
the measured points fall close to straight lines (all $R^2$ values are above $0.92$ on the average accuracy). On the average
metric, we obtain $\gamma = 0.060$ for GSLM2, $0.055$ for SpidR~2.0
SLM and $0.040$ for SpidR SLM. The same ordering holds on each of sWuggy, sBLIMP and tSC
individually. The Text~LM remains ahead at $\gamma = 0.073$,
so GSLM2 narrows the efficiency gap. We also see that every model scales fastest
on tSC ($\gamma = 0.116$ for GSLM2, against $0.044$ on sWuggy), indicating that semantic
competence is where continued scaling pays most.

The two modeling changes can be separated by comparing consecutive pairs of models
in log-error space, where a vertical offset corresponds to a constant multiplicative
reduction in error. Replacing SpidR units with SpidR~2.0 widens the gap steadily
with scale, from $0.048$ to $0.103$ log units on the average metric.
Adding multi-stream modeling on top of SpidR~2.0 units instead
produces a gap that is mostly flat. The exponents show the same split: of
the total difference in $\gamma$ between SpidR and GSLM2 ($+0.021$), the unit upgrade
accounts for $+0.016$ and multi-stream modeling for $+0.005$. The unit encoder change therefore
alters the rate at which error falls with data, whereas multi-stream modeling
shifts the curve down by a constant factor.

\subsubsection{Calibration Curves for Generative Evaluation}
\label{app:sssec:calibration_curves}
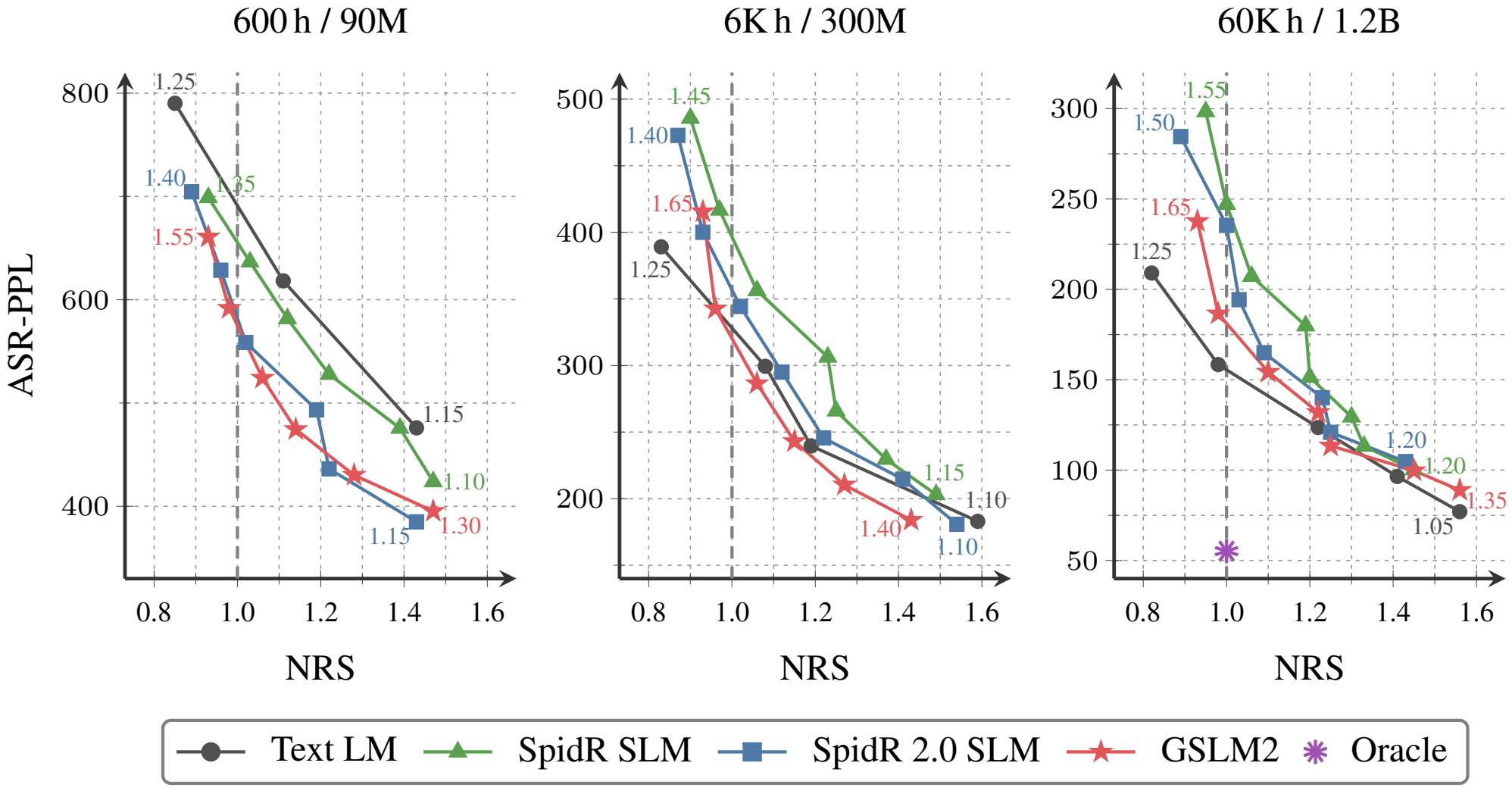
\begin{figure}[t]
\centering

\begin{tikzpicture}

\definecolor{spidrgreen}{RGB}{89,161,79}
\definecolor{spidr2blue}{RGB}{78,121,167}
\definecolor{gslm2red}{RGB}{225,87,89}
\definecolor{textgray}{RGB}{80,80,80}
\definecolor{oraclecolor}{RGB}{160,80,180}

\pgfdeclareplotmark{filled star}{%
    \pgfpathmoveto{
        \pgfqpoint{0pt}{\pgfplotmarksize}
    }%
    \pgfpathlineto{
        \pgfqpoint
            {.224514\pgfplotmarksize}
            {.309017\pgfplotmarksize}
    }%
    \pgfpathlineto{
        \pgfqpoint
            {.951057\pgfplotmarksize}
            {.309017\pgfplotmarksize}
    }%
    \pgfpathlineto{
        \pgfqpoint
            {.363271\pgfplotmarksize}
            {-.118034\pgfplotmarksize}
    }%
    \pgfpathlineto{
        \pgfqpoint
            {.587785\pgfplotmarksize}
            {-.809017\pgfplotmarksize}
    }%
    \pgfpathlineto{
        \pgfqpoint
            {0pt}
            {-.381966\pgfplotmarksize}
    }%
    \pgfpathlineto{
        \pgfqpoint
            {-.587785\pgfplotmarksize}
            {-.809017\pgfplotmarksize}
    }%
    \pgfpathlineto{
        \pgfqpoint
            {-.363271\pgfplotmarksize}
            {-.118034\pgfplotmarksize}
    }%
    \pgfpathlineto{
        \pgfqpoint
            {-.951057\pgfplotmarksize}
            {.309017\pgfplotmarksize}
    }%
    \pgfpathlineto{
        \pgfqpoint
            {-.224514\pgfplotmarksize}
            {.309017\pgfplotmarksize}
    }%
    \pgfpathclose
    \pgfusepath{fill,stroke}%
}

\pgfplotsset{
    comparison axis/.style={
        width=5cm,
        height=6cm,
        xmin=0.73,
        xmax=1.67,
        xtick={0.8,1.0,1.2,1.4,1.6},
        xticklabels={0.8,1.0,1.2,1.4,1.6},
        minor xtick={0.9,1.1,1.3,1.5},
        xlabel={\small NRS},
        ylabel={\small ASR-PPL},
        xticklabel style={
            font=\scriptsize
        },
        yticklabel style={
            font=\scriptsize,
            xshift=2pt
        },
        xlabel style={
            font=\scriptsize,
            yshift=1pt
        },
        ylabel style={
            font=\scriptsize,
            yshift=0pt
        },
        title style={
            font=\scriptsize
        },
        tick align=outside,
        tick style={
            black!70
        },
        axis lines=left,
        axis line style={
            black!80,
            line width=0.9pt
        },
        xmajorgrids=true,
        ymajorgrids=true,
        xminorgrids=true,
        yminorgrids=true,
        major grid style={
            draw=black!40,
            dash pattern=on 1pt off 1.4pt,
            line width=0.3pt
        },
        minor grid style={
            draw=black!40,
            dash pattern=on 1pt off 1.4pt,
            line width=0.3pt
        },
        line width=0.8pt,
        mark size=1.5pt,
        every axis plot/.append style={
            solid,
            mark options={solid}
        },
        extra x ticks={1.0},
        extra x tick labels={\empty},
        extra x tick style={
            grid=major,
            major grid style={
                draw=black!50,
                dash pattern=on 3pt off 2pt,
                line width=0.8pt
            },
            major tick length=0pt
        },
        clip=false
    },
    text lm/.style={
        color=textgray,
        mark=*,
        mark options={
            solid,
            fill=textgray,
            draw=textgray
        }
    },
    spidr slm/.style={
        color=spidrgreen,
        mark=triangle*,
        mark size=2pt,
        mark options={
            solid,
            fill=spidrgreen,
            draw=spidrgreen
        }
    },
    spidr2 slm/.style={
        color=spidr2blue,
        mark=square*,
        mark options={
            solid,
            fill=spidr2blue,
            draw=spidr2blue
        }
    },
    gslm2/.style={
        color=gslm2red,
        mark=filled star,
        mark size=2pt,
        mark options={
            solid,
            sharp corners,
            fill=gslm2red,
            draw=gslm2red
        }
    },
    oracle/.style={
        color=oraclecolor,
        mark=10-pointed star,
        mark size=3pt,
        mark options={
            solid,
            fill=oraclecolor,
            draw=oraclecolor
        }
    },
}

\tikzset{
    temp start/.style={
        pos=0,
        font=\tiny,
        anchor=north,
        xshift=0pt,
        yshift=0pt,
        inner sep=1.0pt,
        text opacity=1
    },
    temp end/.style={
        pos=1,
        font=\tiny,
        anchor=north,
        xshift=0pt,
        yshift=0pt,
        inner sep=1.0pt,
        text opacity=1
    }
}

\begin{groupplot}[
    comparison axis,
    group style={
        group size=3 by 1,
        horizontal sep=0.9cm
    }
]

\nextgroupplot[
    title={\small 600\,h / 90M},
    ymin=330,
    ymax=820,
    ytick={200,400,600,800},
    minor ytick={300,500,700},
    legend to name=sharedlegend,
    legend columns=5,
    legend cell align=left,
    legend style={
        font=\small,
        draw=black!50,
        rounded corners=2pt,
        line width=0.8pt,
        fill=none,
        column sep=4pt
    }
]

\addplot[
    text lm
] coordinates {
(1.43,475.95) %
(1.11,618.26) %
(0.85,790.18) %
}
node[temp start, anchor=south west, xshift=0pt, yshift=0pt] {1.15}
node[temp end, anchor=south, xshift=0pt, yshift=2pt] {1.25};
\addlegendentry{Text LM}

\addplot[
    spidr slm
] coordinates {
(1.47,423.84) %
(1.39,475.65) %
(1.22,527.84) %
(1.12,581.72) %
(1.03,636.66) %
(0.93,698.93) %
}
node[temp start, anchor=west, xshift=1pt, yshift=0pt] {1.10}
node[temp end, anchor=south west, xshift=0pt, yshift=0pt] {1.35};
\addlegendentry{SpidR SLM}

\addplot[
    spidr2 slm
] coordinates {
(1.43,384.92) %
(1.22,436.28) %
(1.19,493.35) %
(1.02,558.62) %
(0.96,628.56) %
(0.89,704.51) %
}
node[temp start, anchor=north east, xshift=0pt, yshift=0pt] {1.15}
node[temp end, anchor=south east, xshift=0pt, yshift=0pt] {1.40};
\addlegendentry{SpidR 2.0 SLM}

\addplot[
    gslm2
] coordinates {
(1.47,395.26) %
(1.28,430.19) %
(1.14,474.46) %
(1.06,524.21) %
(0.98,591.85) %
(0.93,660.89) %
}
node[temp start, anchor=north west, xshift=0pt, yshift=0pt] {1.30}
node[temp end, anchor=east, xshift=-2pt, yshift=0pt] {1.55};
\addlegendentry{GSLM2}

\addlegendimage{only marks, oracle}
\addlegendentry{Oracle}

\nextgroupplot[
    title={\small 6K\,h / 300M},
    ylabel={},
    ymin=140,
    ymax=520,
    ytick={100,200,300,400,500},
    minor ytick={150,250,350,450}
]

\addplot[
    text lm
] coordinates {
(1.59,182.98) %
(1.19,239.65) %
(1.08,299.39) %
(0.83,389.08) %
}
node[temp start, anchor=south, xshift=2pt, yshift=2pt] {1.10}
node[temp end, anchor=north east, xshift=4pt, yshift=-2pt] {1.25};

\addplot[
    spidr slm
] coordinates {
(1.49,203.08) %
(1.37,229.63) %
(1.25,265.90) %
(1.23,306.35) %
(1.06,356.26) %
(0.97,416.70) %
(0.90,485.71) %
}
node[temp start, anchor=south, xshift=2pt, yshift=2pt] {1.15}
node[temp end, anchor=south, xshift=0pt, yshift=2pt] {1.45};

\addplot[
    spidr2 slm
] coordinates {
(1.54,180.78) %
(1.41,214.67) %
(1.22,245.62) %
(1.12,295.05) %
(1.02,344.34) %
(0.93,400.02) %
(0.87,472.75) %
}
node[temp start, anchor=north, xshift=0pt, yshift=-2pt] {1.10}
node[temp end, anchor=east, xshift=-1pt, yshift=0pt] {1.40};

\addplot[
    gslm2
] coordinates {
(1.43,183.93) %
(1.27,210.41) %
(1.15,243.01) %
(1.06,286.48) %
(0.96,342.66) %
(0.93,415.57) %
}
node[temp start, anchor=east, xshift=-1pt, yshift=-2pt] {1.40}
node[temp end, anchor=east, xshift=-1pt, yshift=2pt] {1.65};

\nextgroupplot[
    title={\small 60K\,h / 1.2B},
    ylabel={},
    ymin=40,
    ymax=320,
    ytick={50,100,150,200,250,300},
    minor ytick={75,125,175,225,275}
]

\addplot[
    text lm
] coordinates {
(1.56,77.02) %
(1.41,96.50) %
(1.22,123.59) %
(0.98,158.40) %
(0.82,209.04) %
}
node[temp start, anchor=north east, xshift=0pt, yshift=0pt] {1.05}
node[temp end, anchor=south, xshift=0pt, yshift=2pt] {1.25};

\addplot[
    spidr slm
] coordinates {
(1.45,100.27) %
(1.33,113.24) %
(1.30,129.47) %
(1.20,151.41) %
(1.19,179.86) %
(1.06,207.29) %
(1.00,247.12) %
(0.95,298.36) %
}
node[temp start, anchor=west, xshift=1pt, yshift=1pt] {1.20}
node[temp end, anchor=south, xshift=0pt, yshift=2pt] {1.55};

\addplot[
    spidr2 slm
] coordinates {
(1.43,104.75) %
(1.25,120.92) %
(1.23,140.04) %
(1.09,164.97) %
(1.03,194.31) %
(1.00,235.41) %
(0.89,284.55) %
}
node[temp start, anchor=south, xshift=0pt, yshift=2pt] {1.20}
node[temp end, anchor=south east, xshift=0pt, yshift=0pt] {1.50};

\addplot[
    gslm2
] coordinates {
(1.56,88.90) %
(1.45,99.75) %
(1.25,113.45) %
(1.22,131.92) %
(1.10,154.29) %
(0.98,186.56) %
(0.93,237.67) %
}
node[temp start, anchor=north west, xshift=0pt, yshift=1pt] {1.35}
node[temp end, anchor=south east, xshift=0pt, yshift=0pt] {1.65};

\addplot[
    only marks,
    oracle
] coordinates {
(1.0,55.18)
};

\end{groupplot}

\node[
    anchor=north,
    yshift=-1.1cm
] at (group c2r1.south) {
    \pgfplotslegendfromname{sharedlegend}
};

\end{tikzpicture}

\vspace{-0.2cm}
\caption{
Quality-diversity sweeps for temperature calibration.
Each point represents an evaluated temperature, and endpoint labels indicate the temperature range of each sweep.
The dashed line at $\mathrm{NRS}=1$ indicates the calibration target,
where generated continuations match the repetition level of ground-truth continuations under combined unigram-bigram auto-BLEU.
Values above and below one indicate greater and lower repetition, respectively.
Results are averaged over ten random runs.
The oracle is computed directly from the ground-truth continuations.
}
\label{fig:calibration_curve}

\end{figure}

Figure~\ref{fig:calibration_curve} shows the development-set quality-diversity sweeps used to calibrate the decoding temperature (see Appendix~\ref{app:sssec:asr_ppl}).
Across models, increasing the temperature generally raises both NRS and ASR-PPL over the evaluated range.

Each sweep brackets the target $\mathrm{NRS}=1$.
For each model, we estimate the calibrated temperature $\hat{\tau}$ by linearly interpolating between the two adjacent points that cross this target and rounding the result to the nearest 0.01.
The resulting temperatures, reported in Table~\ref{tab:gen_eval},
enable comparisons using a common calibration target rather than at a fixed decoding temperature.
ASR-PPL is reported to characterize generation quality and does not influence temperature selection.
All temperatures are calibrated on the development set and then held fixed for test-set generation.

\subsubsection{Effect of Layer Order on Generation Performance}
\label{app:sssec:effect_of_layer_order}

\begin{figure*}[t]
\centering

\begin{minipage}[t]{0.48\textwidth}
    \vspace{0pt}
    \begin{tikzpicture}

\definecolor{gslm2red}{RGB}{225,87,89}

\pgfdeclareplotmark{filled star}{%
    \pgfpathmoveto{
        \pgfqpoint{0pt}{\pgfplotmarksize}
    }%
    \pgfpathlineto{
        \pgfqpoint
            {.224514\pgfplotmarksize}
            {.309017\pgfplotmarksize}
    }%
    \pgfpathlineto{
        \pgfqpoint
            {.951057\pgfplotmarksize}
            {.309017\pgfplotmarksize}
    }%
    \pgfpathlineto{
        \pgfqpoint
            {.363271\pgfplotmarksize}
            {-.118034\pgfplotmarksize}
    }%
    \pgfpathlineto{
        \pgfqpoint
            {.587785\pgfplotmarksize}
            {-.809017\pgfplotmarksize}
    }%
    \pgfpathlineto{
        \pgfqpoint
            {0pt}
            {-.381966\pgfplotmarksize}
    }%
    \pgfpathlineto{
        \pgfqpoint
            {-.587785\pgfplotmarksize}
            {-.809017\pgfplotmarksize}
    }%
    \pgfpathlineto{
        \pgfqpoint
            {-.363271\pgfplotmarksize}
            {-.118034\pgfplotmarksize}
    }%
    \pgfpathlineto{
        \pgfqpoint
            {-.951057\pgfplotmarksize}
            {.309017\pgfplotmarksize}
    }%
    \pgfpathlineto{
        \pgfqpoint
            {-.224514\pgfplotmarksize}
            {.309017\pgfplotmarksize}
    }%
    \pgfpathclose
    \pgfusepath{fill,stroke}%
}

\pgfplotsset{
    comparison axis/.style={
        width=6cm,
        height=6cm,
        xmin=0.85,
        xmax=1.55,
        xtick={0.9,1.0,1.1,1.2,1.3,1.4,1.5},
        xticklabels={0.9,1.0,1.1,1.2,1.3,1.4,1.5},
        xlabel={\small NRS},
        ylabel={\small ASR-PPL},
        xticklabel style={
            font=\scriptsize
        },
        yticklabel style={
            font=\scriptsize,
            xshift=2pt
        },
        xlabel style={
            font=\scriptsize,
            yshift=1pt
        },
        ylabel style={
            font=\scriptsize,
            yshift=0pt
        },
        title style={
            font=\scriptsize
        },
        tick align=outside,
        tick style={
            black!70
        },
        axis lines=left,
        axis line style={
            black!80,
            line width=0.9pt
        },
        xmajorgrids=true,
        ymajorgrids=true,
        xminorgrids=true,
        yminorgrids=true,
        major grid style={
            draw=black!40,
            dash pattern=on 1pt off 1.4pt,
            line width=0.3pt
        },
        minor grid style={
            draw=black!40,
            dash pattern=on 1pt off 1.4pt,
            line width=0.3pt
        },
        line width=0.8pt,
        mark size=1.5pt,
        every axis plot/.append style={
            solid,
            mark options={solid}
        },
        extra x ticks={1.0},
        extra x tick labels={\empty},
        extra x tick style={
            grid=major,
            major grid style={
                draw=black!50,
                dash pattern=on 3pt off 2pt,
                line width=0.8pt
            },
            major tick length=0pt
        },
        clip=false
    },
    gslm2/.style={
        color=gslm2red,
        line width=1.2pt,
        mark=filled star,
        mark size=2pt,
        mark options={
            solid,
            sharp corners,
            fill=gslm2red,
            draw=gslm2red
        }
    },
    gslm2_ta1/.style={
        color=black!55,
        line width=0.8pt,
        densely dashed,
        mark=triangle*,
        mark size=2pt,
        mark options={
            solid,
            fill=black!55,
            draw=black!55
        }
    },
    gslm2_asc/.style={
        color=black!42,
        line width=0.8pt,
        dashdotted,
        mark=square*,
        mark size=1.5pt,
        mark options={
            solid,
            fill=black!42,
            draw=black!42
        }
    },
    gslm2_desc/.style={
        color=black!30,
        line width=0.8pt,
        densely dotted,
        mark=diamond*,
        mark size=1.5pt,
        mark options={
            solid,
            fill=black!30,
            draw=black!30
        }
    },
}

\tikzset{
    temp start/.style={
        pos=0,
        font=\tiny,
        anchor=north,
        xshift=0pt,
        yshift=0pt,
        inner sep=1.0pt,
        text opacity=1
    },
    temp end/.style={
        pos=1,
        font=\tiny,
        anchor=north,
        xshift=0pt,
        yshift=0pt,
        inner sep=1.0pt,
        text opacity=1
    }
}

\begin{groupplot}[
    comparison axis,
    group style={
        group size=1 by 1
    }
]

\nextgroupplot[
    title={\small 6K\,h / 300M},
    ymin=140,
    ymax=510,
    ytick={100,200,300,400,500},
    minor ytick={150,250,350,450},
    legend to name=sharedlegend_g,
    legend columns=1,
    legend cell align=left,
    legend style={
        font=\small,
        draw=black!50,
        rounded corners=2pt,
        line width=0.8pt,
        fill=none,
        column sep=4pt
    }
]

\addplot[
    gslm2
] coordinates {
(1.43,183.93) %
(1.27,210.41) %
(1.15,243.01) %
(1.06,286.48) %
(0.96,342.66) %
(0.93,415.57) %
};
\addlegendentry{Bitrate: low $\rightarrow$ high (default)}

\addplot[
    gslm2_ta1
] coordinates {
(1.49,161.86) %
(1.52,176.47) %
(1.32,198.85) %
(1.12,223.50) %
(1.14,269.05) %
(1.03,310.29) %
(0.96,373.22) %
(0.95,416.144)
};
\addlegendentry{Bitrate: high $\rightarrow$ low}

\addplot[
    gslm2_asc
] coordinates {
(1.52,174.80) %
(1.39,198.00) %
(1.28,229.06) %
(1.19,268.34) %
(1.12,319.96) %
(0.96,363.22) %
(1.00,485.85) %
};
\addlegendentry{Layer index: low $\rightarrow$ high}

\addplot[
    gslm2_desc
] coordinates {
(1.47,182.27) %
(1.27,205.24) %
(1.22,233.45) %
(1.14,268.76) %
(0.98,312.18) %
(0.95,359.08) %
(0.88,421.71) %
};
\addlegendentry{Layer index: high $\rightarrow$ low}

\end{groupplot}

\node[
    anchor=north,
    yshift=-1.0cm
] at (group c1r1.south) {
    \pgfplotslegendfromname{sharedlegend_g}
};

\end{tikzpicture}

\vspace{-0.2cm}
\caption{
Effect of layer generation order on the quality-diversity curve for GSLM2 (Libri-Light 6K\,h; 300M parameters).
The adopted low-to-high bitrate order is highlighted in red, while the alternative orders are shown in gray.
}
\label{fig:calibration_curve_gen_order}
\end{minipage}
\hfill
\begin{minipage}[t]{0.48\textwidth}
    \vspace{0pt}
    \begin{tikzpicture}

\definecolor{gslm2red}{RGB}{225,87,89}

\pgfdeclareplotmark{filled star}{%
    \pgfpathmoveto{
        \pgfqpoint{0pt}{\pgfplotmarksize}
    }%
    \pgfpathlineto{
        \pgfqpoint
            {.224514\pgfplotmarksize}
            {.309017\pgfplotmarksize}
    }%
    \pgfpathlineto{
        \pgfqpoint
            {.951057\pgfplotmarksize}
            {.309017\pgfplotmarksize}
    }%
    \pgfpathlineto{
        \pgfqpoint
            {.363271\pgfplotmarksize}
            {-.118034\pgfplotmarksize}
    }%
    \pgfpathlineto{
        \pgfqpoint
            {.587785\pgfplotmarksize}
            {-.809017\pgfplotmarksize}
    }%
    \pgfpathlineto{
        \pgfqpoint
            {0pt}
            {-.381966\pgfplotmarksize}
    }%
    \pgfpathlineto{
        \pgfqpoint
            {-.587785\pgfplotmarksize}
            {-.809017\pgfplotmarksize}
    }%
    \pgfpathlineto{
        \pgfqpoint
            {-.363271\pgfplotmarksize}
            {-.118034\pgfplotmarksize}
    }%
    \pgfpathlineto{
        \pgfqpoint
            {-.951057\pgfplotmarksize}
            {.309017\pgfplotmarksize}
    }%
    \pgfpathlineto{
        \pgfqpoint
            {-.224514\pgfplotmarksize}
            {.309017\pgfplotmarksize}
    }%
    \pgfpathclose
    \pgfusepath{fill,stroke}%
}

\pgfplotsset{
    comparison axis/.style={
        width=6cm,
        height=6cm,
        xmin=0.85,
        xmax=1.55,
        xtick={0.9,1.0,1.1,1.2,1.3,1.4,1.5},
        xticklabels={0.9,1.0,1.1,1.2,1.3,1.4,1.5},
        xlabel={\small NRS},
        ylabel={\small ASR-PPL},
        xticklabel style={
            font=\scriptsize
        },
        yticklabel style={
            font=\scriptsize,
            xshift=2pt
        },
        xlabel style={
            font=\scriptsize,
            yshift=1pt
        },
        ylabel style={
            font=\scriptsize,
            yshift=0pt
        },
        title style={
            font=\scriptsize
        },
        tick align=outside,
        tick style={
            black!70
        },
        axis lines=left,
        axis line style={
            black!80,
            line width=0.9pt
        },
        xmajorgrids=true,
        ymajorgrids=true,
        xminorgrids=true,
        yminorgrids=true,
        major grid style={
            draw=black!40,
            dash pattern=on 1pt off 1.4pt,
            line width=0.3pt
        },
        minor grid style={
            draw=black!40,
            dash pattern=on 1pt off 1.4pt,
            line width=0.3pt
        },
        line width=0.8pt,
        mark size=1.5pt,
        every axis plot/.append style={
            solid,
            mark options={solid}
        },
        extra x ticks={1.0},
        extra x tick labels={\empty},
        extra x tick style={
            grid=major,
            major grid style={
                draw=black!50,
                dash pattern=on 3pt off 2pt,
                line width=0.8pt
            },
            major tick length=0pt
        },
        clip=false
    },
    gslm2/.style={
        color=gslm2red,
        line width=1.2pt,
        mark=filled star,
        mark size=2pt,
        mark options={
            solid,
            sharp corners,
            fill=gslm2red,
            draw=gslm2red
        }
    },
    gslm2_no_sd/.style={
        color=black!55,
        line width=0.8pt,
        densely dashed,
        mark=triangle*,
        mark size=2pt,
        mark options={
            solid,
            fill=black!55,
            draw=black!55
        }
    },
    gslm2_gen_only/.style={
        color=black!42,
        line width=0.8pt,
        dashdotted,
        mark=square*,
        mark size=1.5pt,
        mark options={
            solid,
            fill=black!42,
            draw=black!42
        }
    },
    gslm2_ta1/.style={
        color=black!30,
        line width=0.8pt,
        densely dotted,
        mark=diamond*,
        mark size=1.5pt,
        mark options={
            solid,
            fill=black!30,
            draw=black!30
        }
    },
}

\tikzset{
    temp start/.style={
        pos=0,
        font=\tiny,
        anchor=north,
        xshift=0pt,
        yshift=0pt,
        inner sep=1.0pt,
        text opacity=1
    },
    temp end/.style={
        pos=1,
        font=\tiny,
        anchor=north,
        xshift=0pt,
        yshift=0pt,
        inner sep=1.0pt,
        text opacity=1
    }
}

\begin{groupplot}[
    comparison axis,
    group style={
        group size=1 by 1
    }
]

\nextgroupplot[
    title={\small 6K\,h / 300M},
    ymin=140,
    ymax=510,
    ytick={100,200,300,400,500},
    minor ytick={150,250,350,450},
    legend to name=sharedlegend_a,
    legend columns=1,
    legend cell align=left,
    legend style={
        font=\small,
        draw=black!50,
        rounded corners=2pt,
        line width=0.8pt,
        fill=none,
        column sep=4pt
    }
]

\addplot[
    gslm2
] coordinates {
(1.43,183.93) %
(1.27,210.41) %
(1.15,243.01) %
(1.06,286.48) %
(0.96,342.66) %
(0.93,415.57) %
};
\addlegendentry{GSLM2}

\addplot[
    gslm2_no_sd
] coordinates {
(1.47,289.20) %
(1.30,320.17) %
(1.19,355.91) %
(1.15,384.51) %
(1.08,439.80) %
(0.97,490.19) %
};
\addlegendentry{w/o self-distillation ($\mathcal{L}_{\mathsf{distill}} \rightarrow \mathcal{L}_{\mathsf{ce}}$)}

\addplot[
    gslm2_gen_only
] coordinates {
(1.45,231.19) %
(1.32,252.23) %
(1.19,282.91) %
(1.11,320.88) %
(1.03,363.38) %
(0.97,410.00) %
(0.90,482.93) %
};
\addlegendentry{w/o stream-specific heads ($\alpha=1.0$)}

\addplot[
    gslm2_ta1
] coordinates {
(1.35,202.87) %
(1.19,232.49) %
(1.12,277.75) %
(1.10,324.05) %
(0.98,394.00) %
(1.06,460.68) %
};
\addlegendentry{w/o temperature annealing}

\end{groupplot}

\node[
    anchor=north,
    yshift=-1.0cm
] at (group c1r1.south) {
    \pgfplotslegendfromname{sharedlegend_a}
};

\end{tikzpicture}

\vspace{-0.2cm}
\caption{
Ablation study of GSLM2 on generative performance.
The default model (Libri-Light 6K\,h; 300M parameters) is highlighted in red,
while the ablated variants are shown in gray.
}
\label{fig:calibration_curve_ablation}
\end{minipage}

\end{figure*}

Figure~\ref{fig:calibration_curve_gen_order} compares the development-set quality-diversity curves obtained with different generation orders for 300M-parameter GSLM2 trained on 6K-hour Libri-Light.
We consider two bitrate-based orders for $\pi$ (see Figure~\ref{fig:codebook-layerwise} for the layer-wise bitrates):
ascending bitrate with $(10,5,6,9,8,7)$, and descending bitrate with $(7,8,9,6,5,10)$.
We also evaluate ascending and descending layer-index orders,
given by $(5,6,7,8,9,10)$ and $(10,9,8,7,6,5)$, respectively.

Among the bitrate-based orders,
descending bitrate shows higher ASR-PPL near the calibration target at $\mathrm{NRS}=1$ and exhibits a less smooth trajectory in the low-temperature range.
In contrast, ascending bitrate produces a smoother and more predictable trade-off,
making temperature calibration more reliable.
The layer-index orders exhibit a larger difference.
Ascending layer order causes ASR-PPL to increase sharply near the calibration target and produces an irregular curve in this region.
Descending layer order performs substantially better and remains close to the ascending-bitrate configuration.
However, layer index depends on the set of modeled layers,
whereas bitrate directly reflects the information rate of each stream.

We therefore adopt ascending bitrate as the generation order for GSLM2.
This choice combines competitive generation performance with stable calibration behavior
and remains applicable when the set of layers used for multi-stream modeling changes.
By generating lower-bitrate streams before higher-bitrate ones,
it also induces a coarse-to-fine schedule,
which is consistent with prior work demonstrating the effectiveness of coarse-to-fine generation~\citep{copet2023simple,borsos2023audiolm,defossez2024moshi}.
Across the generation orders considered,
performance on the other benchmarks (i.e., ZS21, SC, and LT-Swap) remains essentially unchanged;
the observed differences are confined to generative evaluation.

\subsubsection{Ablation Studies}
\label{app:sssec:slm_ablation_studies}

\begin{table}
    \centering
    \caption{Ablations of GSLM2 (Libri-Light 6K\,h; 300M parameters) on the primary evaluation metrics, including sWuggy, sBLIMP, and tSC. \texttt{A1} and \texttt{A3} modify the full model (\texttt{A0}), while \texttt{A2} builds on \texttt{A1}. Values are accuracies (\%; chance level: 50\%).}
    \label{tab:slm_eval_ablation}

    \small
    \renewcommand{\arraystretch}{1.12}

    \begin{tabular}{l@{\hspace{0.5em}}lccc}
        \toprule
        & & \multicolumn{2}{c}{\textbf{ZS21} ($\uparrow$)} & \textbf{SC} ($\uparrow$) \\
        \cmidrule(l{0.3em}r{0.3em}){3-4}
        \cmidrule(l{0.3em}r{0.3em}){5-5}
        \texttt{ID} & \textbf{Model} & sWuggy & sBLIMP & tSC \\
        \midrule
        \texttt{A0} & GSLM2 & \textbf{76.3} & \textbf{67.2} & \textbf{78.9} \\
        \texttt{A1} & $\hookrightarrow$ w/o self-distillation ($\mathcal{L}_{\mathsf{distill}} \rightarrow \mathcal{L}_{\mathsf{ce}}$) & 68.5 & 57.3 & 70.1 \\
        \texttt{A2} & \phantom{$\hookrightarrow$} $\hookrightarrow$ w/o single-stream loss ($\beta = 0.0$) & 67.9 & 55.8 & 65.4 \\
        \texttt{A3} & $\hookrightarrow$ Halve context length \& batch size & 75.7 & 66.8 & 78.3 \\
        \midrule
        \texttt{A4} & SpidR 2.0 SLM & 73.9 & 65.1 & 74.7 \\
        \texttt{A5} & $\hookrightarrow$ matched GSLM2 training FLOPs & 73.3 & 64.5 & 75.1 \\
        \bottomrule
    \end{tabular}%
\end{table}

We ablate training objectives and configuration of GSLM2 to assess their contributions to the primary metrics, i.e., sWuggy, sBLIMP, and tSC (Table~\ref{tab:slm_eval_ablation}).
We also evaluate the effects of stream-specific supervision and temperature annealing on generative performance (Figure~\ref{fig:calibration_curve_ablation}).

\textbf{Self-distillation is critical to performance gains.}
In \texttt{A1} (Table~\ref{tab:slm_eval_ablation}),
we replace the self-distillation loss with a cross-entropy loss ($\mathcal{L}_{\mathsf{distill}}\rightarrow\mathcal{L}_{\mathsf{ce}}$ in $\mathcal{L}_{\mathsf{slm}}$),
training the stream-specific heads on hard speech-unit targets at the same transition positions.
Relative to the full model (\texttt{A0}),
this reduces sWuggy, sBLIMP, and tSC by 7.8, 9.9, and 8.8 percentage points, respectively.
Building on \texttt{A1}, \texttt{A2} sets $\beta=0.0$,
removing the single-stream loss $\mathcal{L}_{\mathsf{single}}$ and reallocating its weight to the cross-entropy loss.
Training then uses only the cross-entropy and generative losses,
and scores decline by a further 0.6, 1.5, and 4.7 points, respectively.
The substantial degradation in \texttt{A1} highlights the importance of transition-level self-distillation.
The additional decline in \texttt{A2} suggests that retaining the single-stream objective remains beneficial even without self-distillation, particularly on tSC.

\textbf{Performance is largely retained with shorter contexts and smaller batches.}
GSLM2 uses longer contexts and larger batches to maintain coverage of non-repeated speech units (see Appendix~\ref{app:ssec:training}).
Halving both settings (\texttt{A3}) reduces sWuggy, sBLIMP, and tSC by only 0.6, 0.4, and 0.6 points, respectively,
relative to the full model (\texttt{A0}).
The resulting model still outperforms the single-stream baseline in \texttt{A4}.
It also requires 10\% fewer training FLOPs than the full configuration (Table~\ref{tab:flops}), 
offering lower computational cost with only modest performance degradation.
In addition, increasing the batch size of SpidR 2.0 SLM to match GSLM2's training FLOPs (\texttt{A5}) does not improve its overall performance.

\textbf{Self-distillation also improves generative performance.}
For generation, we evaluate the same replacement of self-distillation with cross-entropy used in \texttt{A1}.
We also evaluate a separate variant that removes the stream-specific heads entirely and sets $\alpha=1.0$ in $\mathcal{L}_{\mathsf{slm}}$,
retaining only the multi-stream generative loss $\mathcal{L}_{\mathsf{gen}}$.
The full model achieves lower ASR-PPL than both variants at comparable NRS values across the evaluated range (Figure~\ref{fig:calibration_curve_ablation}),
supporting the benefit of self-distillation over cross-entropy supervision and of stream-specific supervision more broadly.
These gains are consistent with the auxiliary objectives regularizing the shared temporal representations used by the generative head
through per-stream transition modeling and consistency between single-stream and multi-stream predictions.

\textbf{Temperature annealing improves and stabilizes generation.}
During generation, GSLM2 anneals the sampling temperature across successive stream predictions within each tuple (see Appendix~\ref{app:sssec:asr_ppl}).
Without annealing, ASR-PPL is higher at comparable NRS values,
including the calibration target $\mathrm{NRS}=1$.
At higher decoding temperatures, NRS decreases after exceeding this target, while ASR-PPL continues to rise.
With annealing, NRS increases monotonically over the evaluated temperature range,
yielding a more predictable quality-diversity curve.
These observations are consistent with annealing helping to limit the accumulation of sampling noise across stream predictions within each tuple.

\subsubsection{Scoring Choices of GSLM2}
\label{app:sssec:scoring_choices}
We examine how the choice of output head affects paired-stimulus evaluation of GSLM2,
using the same model trained at 6K\,hours in Table~\ref{tab:main_results}.
Our default scoring protocol uses the stream-specific heads:
for each stream, we average log probabilities over its unit-change positions,
including the first unit, and then average the resulting scores across streams.
This is consistent with the event selection used by independently deduplicated single-stream baselines, while retaining the full multi-stream history as context.
For comparison, we evaluate the generative head by averaging log probabilities over all retained positions, including per-stream repetitions.
Averaging these scores across streams gives the joint sequence log-likelihood normalized by the number of predicted units.

\begin{table}
    \centering
    \caption{Performance of GSLM2 (Libri-Light 6K\,h; 300M parameters) using different heads for scoring. Stream-specific heads score unit-change positions, whereas the generative head scores all retained positions. Individual rows select one stream's score while retaining all input streams; ``All streams'' averages per-stream scores before making paired comparisons. Scores are accuracies (\%; chance level: 50\%).}

    \label{tab:scoring_choices}

    \small
    \renewcommand{\arraystretch}{1.12}

    \begin{tabular}{cccccccc}
        \toprule
        & & \multicolumn{3}{c}{\textbf{Stream-specific head}} & \multicolumn{3}{c}{\textbf{Generative head}} \\
        \cmidrule(l{0.3em}r{0.3em}){3-5}
        \cmidrule(l{0.3em}r{0.3em}){6-8}
        Order $j$ & Layer $\pi(j)$ & sWuggy & sBLIMP & tSC & sWuggy & sBLIMP & tSC \\
        \midrule
        1 & 10 & 70.1 & 62.8 & 77.2 & 57.2 & 53.6 & 59.3 \\
        2 &  5 & 71.7 & 64.8 & 73.7 & 61.8 & 53.5 & 55.9 \\
        3 &  6 & 72.1 & 64.7 & 77.6 & 57.5 & 52.0 & 53.2 \\
        4 &  9 & 71.5 & 64.0 & 79.3 & 60.9 & 52.2 & 61.8 \\
        5 &  8 & 72.3 & 63.0 & 77.7 & 55.6 & 50.8 & 51.8 \\
        6 &  7 & 72.8 & 65.2 & 77.2 & 56.2 & 51.5 & 54.6 \\
        \cdashlinelr{1-8}
        \multicolumn{2}{c}{All streams} & 76.3 & 67.2 & 78.9 & 66.7 & 56.3 & 61.7 \\
        \bottomrule
    \end{tabular}%
\end{table}

Table~\ref{tab:scoring_choices} reports accuracies obtained by scoring individual streams while retaining the full multi-stream input.
Stream-specific predictions condition only on the preceding multi-stream tuples.
For the generative head, each row instead scores the corresponding conditional factor,
which additionally conditions on the observed units of earlier streams within the current tuple.
Stream-specific scoring outperforms generative scoring for every stream on all three tasks, 
supporting our default transition score for these linguistic comparisons.
The gap persists at $j=1$, where both heads condition only on temporal context.
Thus, within-tuple conditioning at evaluation cannot alone explain the difference.

These findings are consistent with prior work reporting benefits from removing consecutive unit repetitions in spoken language modeling~\citep{lakhotia2021generative}.
Related approaches separate content and duration modeling:
\citet{kharitonov2022text} model deduplicated units and their durations separately,
while \citet{nguyen2023generative} show that supervision at unit-change positions improves unit prediction even when repeated inputs are retained for cross-stream alignment.
The stream-specific heads similarly emphasize transitions in both training and scoring.
The generative head additionally predicts and scores the persistence events retained by cross-stream deduplication.
Including these repetition terms in the averaged score is likely to reduce its sensitivity to the linguistic contrasts being evaluated.

Aggregating stream scores improves sWuggy and sBLIMP over the best individual stream-specific results.
On tSC, aggregation remains close to the strongest individual result from layer~9.
We examine this ensembling effect further in Appendix~\ref{app:sssec:layer_diversity}.
Our main evaluations use the same unweighted all-stream transition score across benchmarks, avoiding task-specific output-stream selection.
These results assess paired-stimulus discrimination;
generation uses the generative head and is evaluated separately through prompt-conditioned continuations.

\subsubsection{Effect of Layer Diversity}
\label{app:sssec:layer_diversity}
\begin{table}[t]
    \centering
    \caption{Effect of layer diversity. All GSLM2 variants use six streams and cross-stream deduplication. The repeated-layer control copies the layer-7 argmax sequence across streams; the sampled control independently samples units from the same layer's categorical distributions for each stream. Scores are accuracies (\%; chance level: 50\%).}
    \label{tab:layer_diversity}

    \small
    \renewcommand{\arraystretch}{1.12}

    \begin{tabular}{lcccc}
        \toprule
        & & \multicolumn{2}{c}{\textbf{ZS21} ($\uparrow$)}
          & \textbf{SC} ($\uparrow$) \\
        \cmidrule(lr){3-4}
        \cmidrule(l){5-5}
        \textbf{Model} & \textbf{Unit layers}
        & sWuggy & sBLIMP & tSC \\
        \midrule
        SpidR 2.0 SLM
        & 7 & 73.9 & 65.1 & 74.7 \\
        \midrule
        \multirow{3}{*}{GSLM2}
        & 5--10
        & \textbf{76.3} & \textbf{67.2} & \textbf{78.9} \\
        & 7 (repeated)
        & 73.7 & 65.3 & 74.4 \\
        & 7 (sampled)
        & 69.9 & 61.0 & 68.4 \\
        \bottomrule
    \end{tabular}
\end{table}

We investigate whether GSLM2's gains arise from complementary information across encoder layers or from aggregating multiple noisy realizations of the same representation.
Here, we compare the standard six-layer configuration (from the 6K\,h / 300M setup in Table~\ref{tab:main_results}) with two six-stream controls derived exclusively from layer 7,
which is used by the single-stream SpidR 2.0 SLM.
All three configurations use independently initialized stream-specific heads,
trained jointly with a shared temporal backbone.
The first control repeats the layer-7 argmax sequence across streams, testing score aggregation across independently initialized heads over identical units.
The second independently samples units from the layer's categorical distributions for each stream,
introducing variation in unit identities and stream boundaries.
This sampling control is motivated by the benefits of stochastic tokenization in neural machine translation~\citep{kudo2018subword,provilkov2020bpe}
and consistency training across alternative tokenizations in cross-lingual transfer~\citep{wang2021multi}.
We adjust the sampling temperature to match GSLM2's cross-stream deduplication rate (i.e., 1.36 from Appendix~\ref{app:sssec:layer_selection}).

As shown in Table~\ref{tab:layer_diversity},
the standard multi-layer configuration outperforms both controls on all three benchmarks.
Repeating layer 7 yields performance close to the single-stream baseline,
while categorical sampling reduces accuracy.
Thus, neither score aggregation across independently initialized heads over identical units nor stochastic tokenization at a matched deduplication rate reproduces GSLM2's gains in these experiments.
These findings support the importance of layer diversity
and are consistent with our hypothesis that different encoder layers provide complementary information.

\subsubsection{Comparison with Speech-Capable LLMs}
\label{app:sssec:slm_eval_speech_llm}

To compare GSLM2 with recent speech-capable LLMs,
we evaluate instruction-tuned models that accept speech directly on identical paired-stimulus benchmarks.
These models do not expose likelihood scores for their speech inputs,
and the likelihood-based evaluation used for conventional SLMs is not applicable.
Therefore, we reformulate each benchmark item as a forced-choice question between two recordings:
the benchmark's correct candidate and its incorrect counterpart.

\paragraph{\textbf{Prompting.}}
All models receive the following system prompt:
\begin{quote}
\small
You are participating in a speech understanding benchmark.

You will hear spoken audio options labeled A and B. Your task is to choose the better option according to the instruction.

Answer with exactly one character: A or B. Do not explain your answer. Do not transcribe the audio.
\end{quote}

The two recordings are provided as options A and B jointly with the task-specific instruction:
\begin{itemize}[leftmargin=1.5em,topsep=0pt]
\item sWuggy:
\begin{quote}
\small
You are given two spoken word candidates, Option A and Option B.

Choose the option that sounds more like a real English word.
\end{quote}

\item sBLIMP:
\begin{quote}
\small
You are given two spoken English sentence candidates, Option A and Option B.

Choose the option that sounds more grammatical, natural, and well-formed in English.
\end{quote}

\item tSC and sSC:
\begin{quote}
\small
You are given two spoken stories, Option A and Option B.

Each story contains the same context but a different ending.

Choose the story whose ending is more coherent, plausible, and natural given the preceding context.
\end{quote}
\end{itemize}

\paragraph{\textbf{Scoring.}}
We use greedy decoding with a 256-token output limit configured to each model's native multimodal interface.
A strictly valid response yields an unambiguous A or B choice.
Invalid responses are simply counted as incorrect.
To measure positional bias,
we doubly evaluate each pair, reversing the order of the correct and incorrect recordings.
The reported score pools decisions across both orientations.
Consequently, a model that always selects the same position regardless of configuration receives an accuracy of 50\%.

For GSLM2, we retain the conventional scoring procedure.
Therefore, the speech LLMs are scored differently, and
these results should not be interpreted as a controlled comparison of their underlying speech modeling capabilities.
Instead, the speech-LLM scores measure end-to-end task performance,
jointly reflecting speech perception, linguistic reasoning, instruction following, and response generation.
We report them as a practical reference for how frontier speech LLMs perform and to contextualize the extent to which GSLM2 remains competitive.

\paragraph{\textbf{Speech LLMs.}}
We evaluate the following open-weight checkpoints:
Gemma 4 (4B\footnote{\url{https://huggingface.co/google/gemma-4-E4B-it}} and 12B\footnote{\url{https://huggingface.co/google/gemma-4-12B-it}}),
Qwen3-Omni\footnote{\url{https://huggingface.co/Qwen/Qwen3-Omni-30B-A3B-Instruct}},
Phi-4\footnote{\url{https://huggingface.co/microsoft/Phi-4-multimodal-instruct}},
Voxtral (3B\footnote{\url{https://huggingface.co/mistralai/Voxtral-Mini-3B-2507}} and 24B\footnote{\url{https://huggingface.co/mistralai/Voxtral-Small-24B-2507}}),
MiniCPM-o 4.5\footnote{\url{https://huggingface.co/openbmb/MiniCPM-o-4_5}}, and
Nemotron 3 Nano Omni\footnote{\url{https://huggingface.co/nvidia/Nemotron-3-Nano-Omni-30B-A3B-Reasoning-BF16}}.
We also evaluate the proprietary models via their respective APIs:
Gemini 2.5 Flash\footnote{\url{https://ai.google.dev/gemini-api/docs/models/gemini-2.5-flash}},
Gemini 3.5 Flash\footnote{\url{https://ai.google.dev/gemini-api/docs/models/gemini-3.5-flash}}, and
GPT-Realtime-2\footnote{\url{https://developers.openai.com/api/docs/models/gpt-realtime-2}}.

\begin{table}
    \centering
    \caption{Results for open-weight and proprietary speech-capable LLMs on ZeroSpeech'21 (ZS21) and StoryCloze (SC). Speech LLMs use prompted forced-choice evaluation. GSLM2 is included as a speech-only reference evaluated using likelihood-based scoring. Scores are accuracies (\%; chance level: 50\%).}
    \label{tab:slm_eval_speech_llm}

    \small
    \renewcommand{\arraystretch}{1.12}

    \begin{tabular}{lccccc}
        \toprule
        & & \multicolumn{2}{c}{\textbf{ZS21} ($\uparrow$)} & \multicolumn{2}{c}{\textbf{SC} ($\uparrow$)} \\
        \cmidrule(l{0.3em}r{0.3em}){3-4}
        \cmidrule(l{0.3em}r{0.3em}){5-6}
        \textbf{Model}
        & \textbf{\#Params}
        & sWuggy
        & sBLIMP
        & tSC
        & sSC \\
        \midrule

        \addlinespace[-2.5pt]
        \rowcolor{gray!15}
        \multicolumn{6}{l}{\textbf{Ours}} \\

        GSLM2 (Libri-Light 60K\,h) & 1.2B & 79.3 & 71.8 & 84.5 & 57.6 \\
        GSLM2 (Conversational Speech 6M\,h) & 6.5B & \textbf{80.7} & \textbf{72.9} & \textbf{90.5} & \textbf{63.2} \\

        \addlinespace[2pt]
        \rowcolor{gray!15}
        \multicolumn{6}{l}{\textbf{Open-weight speech LLMs}} \\
        \multirow{2}{*}{Gemma 4~\citep{team2026gemma}}
        & 4B & 72.6 & 67.7 & 88.6 & 80.4 \\
        & 12B & 43.8 & 71.2 & 97.4 & 92.4 \\
        Qwen3-Omni~\citep{xu2025qwen3} & 30B & 84.0 & \textbf{81.5} & 97.8 & 92.1 \\
        Phi-4~\citep{abouelenin2025phi} & 5.6B & 66.3 & 66.9 & 79.9 & 70.6 \\
        \multirow{2}{*}{Voxtral~\citep{liu2025voxtral}} 
        & 3B & 79.1 & 73.8 & 84.8 & 79.7 \\
        & 24B & 81.1 & 79.9 & \textbf{98.6} & \textbf{97.3} \\
        MiniCPM-o 4.5~\citep{cui2026minicpm} & 9B & \textbf{84.7} & 78.4 & 97.3 & 93.1 \\
        Nemotron 3 Nano Omni~\citep{deshmukh2026nemotron} & 30B & 75.8 & 64.9 & 55.5 & 54.8 \\

        \addlinespace[2pt]
        \rowcolor{gray!15}
        \multicolumn{6}{l}{\textbf{Proprietary speech LLMs}} \\
        Gemini 2.5 Flash & -- & 68.7 & 80.8 & 99.4 & 98.3 \\
        Gemini 3.5 Flash & -- & \textbf{86.2} & 88.1 & \textbf{99.8} & 99.4 \\
        GPT-Realtime-2 & -- & 83.5 & \textbf{91.0} & 98.8 & \textbf{99.6} \\
        \bottomrule
    \end{tabular}%
\end{table}

\paragraph{Results.}
Table~\ref{tab:slm_eval_speech_llm} presents speech LLM results on the ZS21 and SC benchmarks,
alongside GSLM2 results from the 60K-hour Libri-Light (Table~\ref{tab:main_results}) and 6M-hour conversational speech experiments (Figure~\ref{fig:scaling}).
Among open-weight speech LLMs, performance varies substantially across models and benchmarks:
MiniCPM-o 4.5 leads on sWuggy, Qwen3-Omni on sBLIMP, and Voxtral 24B on both SC tasks.
Proprietary models achieve the highest scores across all four benchmarks,
with all three exceeding 98\% on both tSC and sSC.
Given the scoring differences discussed above, GSLM2 provides a practical reference:
its ZS21 scores are comparable to those of several speech LLMs of similar size (e.g., Gemma 4, Phi-4, and Voxtral),
while the strongest speech LLMs obtain higher SC scores, particularly on sSC.
Near-saturation on SC should also be interpreted in light of frontier speech LLMs' substantially larger training corpora,
including the text used for pretraining.
These benchmarks remain informative at the speech-only scales studied here,
but extending speech-text data-scaling studies to larger models and datasets will require more challenging evaluations.

\subsubsection{LongTail-Swap Results by Word-Frequency Bin}
\label{app:sssec:lt_swap}

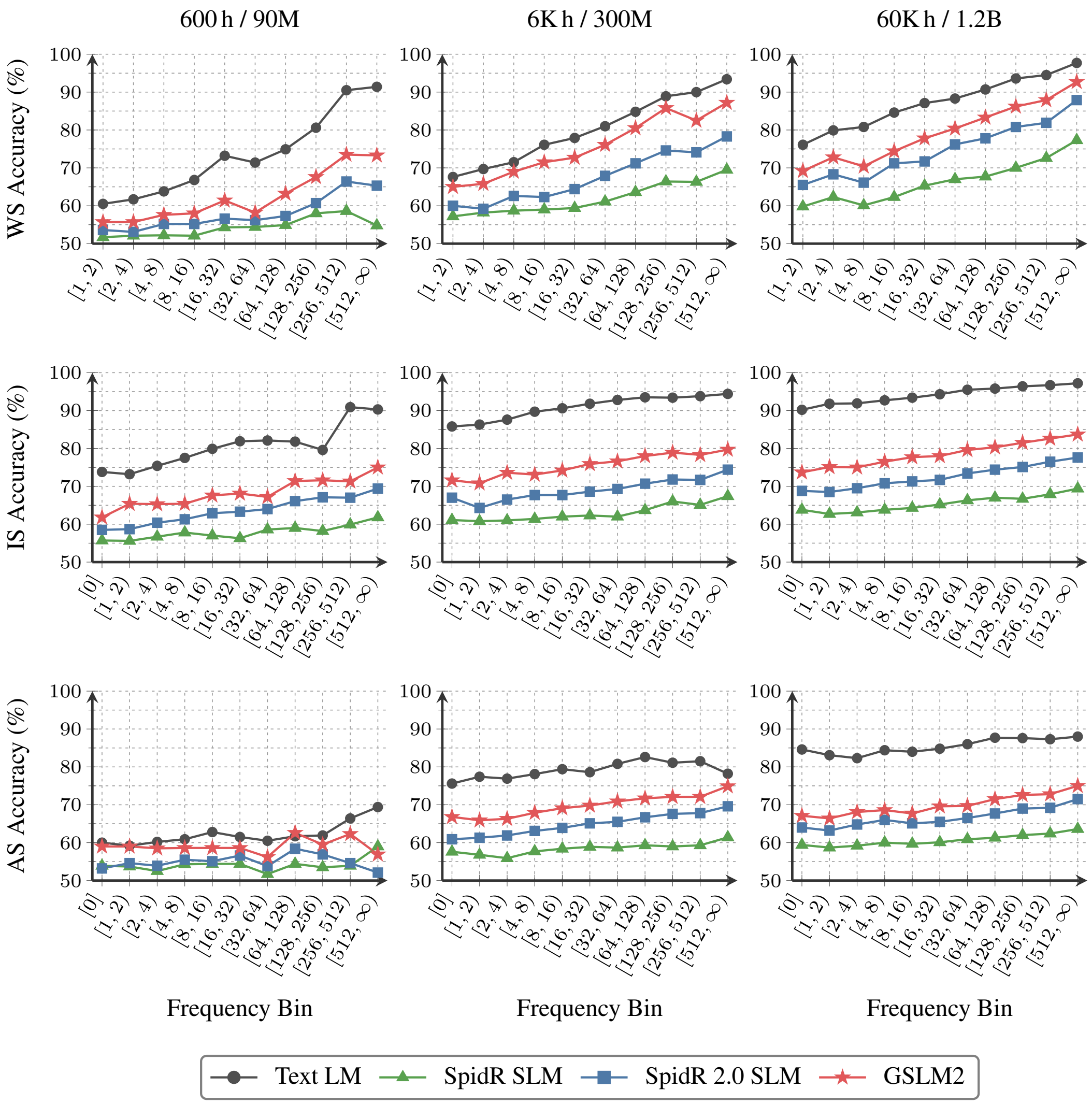
\begin{figure}[t]
\centering
\begin{tikzpicture}

\definecolor{textgray}{RGB}{80,80,80}
\definecolor{spidrgreen}{RGB}{89,161,79}
\definecolor{spidr2blue}{RGB}{78,121,167}
\definecolor{gslm2red}{RGB}{225,87,89}

\pgfdeclareplotmark{filled star}{%
  \pgfpathmoveto{\pgfqpoint{0pt}{\pgfplotmarksize}}%
  \pgfpathlineto{\pgfqpoint{.224514\pgfplotmarksize}{.309017\pgfplotmarksize}}%
  \pgfpathlineto{\pgfqpoint{.951057\pgfplotmarksize}{.309017\pgfplotmarksize}}%
  \pgfpathlineto{\pgfqpoint{.363271\pgfplotmarksize}{-.118034\pgfplotmarksize}}%
  \pgfpathlineto{\pgfqpoint{.587785\pgfplotmarksize}{-.809017\pgfplotmarksize}}%
  \pgfpathlineto{\pgfqpoint{0pt}{-.381966\pgfplotmarksize}}%
  \pgfpathlineto{\pgfqpoint{-.587785\pgfplotmarksize}{-.809017\pgfplotmarksize}}%
  \pgfpathlineto{\pgfqpoint{-.363271\pgfplotmarksize}{-.118034\pgfplotmarksize}}%
  \pgfpathlineto{\pgfqpoint{-.951057\pgfplotmarksize}{.309017\pgfplotmarksize}}%
  \pgfpathlineto{\pgfqpoint{-.224514\pgfplotmarksize}{.309017\pgfplotmarksize}}%
  \pgfpathclose
  \pgfusepath{fill,stroke}%
}

\pgfplotstableread{
bin text spidr spidr2 gslm
0 60.5 51.7 53.6 55.7
1 61.7 52.1 53.1 55.7
2 63.8 52.2 55.2 57.6
3 66.8 52.1 55.2 58.0
4 73.2 54.3 56.6 61.4
5 71.4 54.4 56.2 58.2
6 74.9 54.9 57.3 63.2
7 80.6 58.0 60.7 67.6
8 90.5 58.6 66.4 73.5
9 91.4 54.8 65.3 73.3
}\dataSmall

\pgfplotstableread{
bin text spidr spidr2 gslm
0 67.6 57.2 60.0 65.0
1 69.7 58.2 59.2 65.8
2 71.5 58.7 62.6 69.0
3 76.1 59.0 62.3 71.5
4 77.9 59.4 64.4 72.7
5 81.0 61.1 67.9 76.1
6 84.8 63.6 71.2 80.5
7 88.9 66.4 74.6 85.8
8 90.0 66.3 74.1 82.5
9 93.4 69.5 78.3 87.2
}\dataMedium

\pgfplotstableread{
bin text spidr spidr2 gslm
0  76.1  59.8  65.5  69.2
1  79.9  62.3  68.3  72.8
2  80.8  60.1  66.1  70.4
3  84.6  62.3  71.2  74.4
4  87.1  65.3  71.7  77.8
5  88.3  67.0  76.2  80.4
6  90.7  67.7  77.8  83.3
7  93.6  70.0  80.8  86.2
8  94.5  72.6  81.9  87.9
9  97.7  77.3  87.9  92.7
}\dataLarge

\pgfplotsset{
  comparison axis/.style={
    width=5.35cm, height=4cm,
    xmin=-0.35, xmax=9.35,
    ymin=50, ymax=100,
    xtick={0,1,2,3,4,5,6,7,8,9},
    xticklabels={
      {$[1,2)$}, {$[2,4)$}, {$[4,8)$}, {$[8,16)$},
      {$[16,32)$}, {$[32,64)$}, {$[64,128)$},
      {$[128,256)$}, {$[256,512)$}, {$[512,\infty)$}
    },
    ytick={50,60,70,80,90,100},
    minor ytick={55,65,75,85,95},
    ylabel={\small WS Accuracy (\%)},
    xticklabel style={font=\scriptsize,rotate=60,anchor=east},
    yticklabel style={font=\scriptsize,xshift=2pt},
    xlabel style={font=\scriptsize,yshift=1pt},
    ylabel style={font=\scriptsize},
    title style={font=\small},
    tick align=outside, tick style={black!70},
    axis lines=left,
    axis line style={black!80,line width=0.9pt},
    xmajorgrids=true, ymajorgrids=true, yminorgrids=true,
    major grid style={
      draw=black!40,dash pattern=on 1pt off 1.4pt,line width=0.3pt
    },
    minor grid style={
      draw=black!40,dash pattern=on 1pt off 1.4pt,line width=0.3pt
    },
    line width=0.8pt, mark size=1.5pt,
    every axis plot/.append style={solid,mark options={solid}},
    clip=false
  },
  text lm/.style={
    color=textgray,mark=*,
    mark options={solid,fill=textgray,draw=textgray}
  },
  spidr slm/.style={
    color=spidrgreen,mark=triangle*,mark size=2pt,
    mark options={solid,fill=spidrgreen,draw=spidrgreen}
  },
  spidr2 slm/.style={
    color=spidr2blue,mark=square*,
    mark options={solid,fill=spidr2blue,draw=spidr2blue}
  },
  gslm2/.style={
    color=gslm2red,mark=filled star,mark size=2pt,
    mark options={solid,sharp corners,fill=gslm2red,draw=gslm2red}
  }
}

\begin{groupplot}[
  comparison axis,
  group style={group size=3 by 1,horizontal sep=0.7cm}
]

\nextgroupplot[
  title={600\,h / 90M},
]
\addplot[text lm] table[x=bin,y=text] {\dataSmall};
\addplot[spidr slm] table[x=bin,y=spidr] {\dataSmall};
\addplot[spidr2 slm] table[x=bin,y=spidr2] {\dataSmall};
\addplot[gslm2] table[x=bin,y=gslm] {\dataSmall};

\nextgroupplot[title={6K\,h / 300M},ylabel={}]
\addplot[text lm] table[x=bin,y=text] {\dataMedium};
\addplot[spidr slm] table[x=bin,y=spidr] {\dataMedium};
\addplot[spidr2 slm] table[x=bin,y=spidr2] {\dataMedium};
\addplot[gslm2] table[x=bin,y=gslm] {\dataMedium};

\nextgroupplot[title={60K\,h / 1.2B},ylabel={}]
\addplot[text lm] table[x=bin,y=text] {\dataLarge};
\addplot[spidr slm] table[x=bin,y=spidr] {\dataLarge};
\addplot[spidr2 slm] table[x=bin,y=spidr2] {\dataLarge};
\addplot[gslm2] table[x=bin,y=gslm] {\dataLarge};

\end{groupplot}

\end{tikzpicture}
\begin{tikzpicture}

\definecolor{textgray}{RGB}{80,80,80}
\definecolor{spidrgreen}{RGB}{89,161,79}
\definecolor{spidr2blue}{RGB}{78,121,167}
\definecolor{gslm2red}{RGB}{225,87,89}

\pgfdeclareplotmark{filled star}{%
  \pgfpathmoveto{\pgfqpoint{0pt}{\pgfplotmarksize}}%
  \pgfpathlineto{\pgfqpoint{.224514\pgfplotmarksize}{.309017\pgfplotmarksize}}%
  \pgfpathlineto{\pgfqpoint{.951057\pgfplotmarksize}{.309017\pgfplotmarksize}}%
  \pgfpathlineto{\pgfqpoint{.363271\pgfplotmarksize}{-.118034\pgfplotmarksize}}%
  \pgfpathlineto{\pgfqpoint{.587785\pgfplotmarksize}{-.809017\pgfplotmarksize}}%
  \pgfpathlineto{\pgfqpoint{0pt}{-.381966\pgfplotmarksize}}%
  \pgfpathlineto{\pgfqpoint{-.587785\pgfplotmarksize}{-.809017\pgfplotmarksize}}%
  \pgfpathlineto{\pgfqpoint{-.363271\pgfplotmarksize}{-.118034\pgfplotmarksize}}%
  \pgfpathlineto{\pgfqpoint{-.951057\pgfplotmarksize}{.309017\pgfplotmarksize}}%
  \pgfpathlineto{\pgfqpoint{-.224514\pgfplotmarksize}{.309017\pgfplotmarksize}}%
  \pgfpathclose
  \pgfusepath{fill,stroke}%
}

\pgfplotstableread{
bin text spidr spidr2 gslm
0   73.8  55.7  58.5  61.8
1   73.2  55.6  58.7  65.4
2   75.4  56.7  60.4  65.3
3   77.5  57.8  61.3  65.4
4   79.9  57.0  62.9  67.6
5   81.9  56.3  63.3  68.1
6   82.1  58.6  64.0  67.2
7   81.8  59.0  66.1  71.4
8   79.6  58.2  67.1  71.6
9   90.9  59.9  67.0  71.3
10  90.3  61.8  69.4  75.0
}\dataSmall

\pgfplotstableread{
bin text spidr spidr2 gslm
0   85.8  61.1  67.0  71.6
1   86.3  60.8  64.3  70.8
2   87.6  61.0  66.5  73.6
3   89.7  61.4  67.7  73.1
4   90.6  62.0  67.7  74.2
5   91.8  62.3  68.6  75.9
6   92.8  62.0  69.3  76.6
7   93.5  63.7  70.7  78.0
8   93.4  66.0  71.8  78.9
9   93.8  65.1  71.7  78.3
10  94.4  67.4  74.4  79.7
}\dataMedium

\pgfplotstableread{
bin text spidr spidr2 gslm
0   90.2  63.8  68.8  73.7
1   91.8  62.7  68.5  75.1
2   91.9  63.1  69.5  75.0
3   92.7  63.8  70.8  76.5
4   93.4  64.3  71.3  77.7
5   94.3  65.2  71.7  78.0
6   95.5  66.3  73.4  79.6
7   95.8  67.0  74.4  80.3
8   96.4  66.7  75.1  81.5
9   96.7  67.9  76.5  82.6
10  97.2  69.4  77.6  83.7
}\dataLarge

\pgfplotsset{
  comparison axis/.style={
    width=5.35cm, height=4cm,
    xmin=-0.35, xmax=10.35,
    ymin=50, ymax=100,
    xtick={0,1,2,3,4,5,6,7,8,9,10},
    xticklabels={
      {$[0]$}, {$[1,2)$}, {$[2,4)$}, {$[4,8)$}, {$[8,16)$},
      {$[16,32)$}, {$[32,64)$}, {$[64,128)$},
      {$[128,256)$}, {$[256,512)$}, {$[512,\infty)$}
    },
    ytick={50,60,70,80,90,100},
    minor ytick={55,65,75,85,95},
    ylabel={\small IS Accuracy (\%)},
    xticklabel style={font=\scriptsize,rotate=60,anchor=east},
    yticklabel style={font=\scriptsize,xshift=2pt},
    xlabel style={font=\scriptsize,yshift=1pt},
    ylabel style={font=\scriptsize},
    title style={font=\small},
    tick align=outside, tick style={black!70},
    axis lines=left,
    axis line style={black!80,line width=0.9pt},
    xmajorgrids=true, ymajorgrids=true, yminorgrids=true,
    major grid style={
      draw=black!40,dash pattern=on 1pt off 1.4pt,line width=0.3pt
    },
    minor grid style={
      draw=black!40,dash pattern=on 1pt off 1.4pt,line width=0.3pt
    },
    line width=0.8pt, mark size=1.5pt,
    every axis plot/.append style={solid,mark options={solid}},
    clip=false
  },
  text lm/.style={
    color=textgray,mark=*,
    mark options={solid,fill=textgray,draw=textgray}
  },
  spidr slm/.style={
    color=spidrgreen,mark=triangle*,mark size=2pt,
    mark options={solid,fill=spidrgreen,draw=spidrgreen}
  },
  spidr2 slm/.style={
    color=spidr2blue,mark=square*,
    mark options={solid,fill=spidr2blue,draw=spidr2blue}
  },
  gslm2/.style={
    color=gslm2red,mark=filled star,mark size=2pt,
    mark options={solid,sharp corners,fill=gslm2red,draw=gslm2red}
  }
}

\begin{groupplot}[
  comparison axis,
  group style={group size=3 by 1,horizontal sep=0.7cm}
]

\nextgroupplot[
]
\addplot[text lm] table[x=bin,y=text] {\dataSmall};
\addplot[spidr slm] table[x=bin,y=spidr] {\dataSmall};
\addplot[spidr2 slm] table[x=bin,y=spidr2] {\dataSmall};
\addplot[gslm2] table[x=bin,y=gslm] {\dataSmall};

\nextgroupplot[
    ylabel={},
]
\addplot[text lm] table[x=bin,y=text] {\dataMedium};
\addplot[spidr slm] table[x=bin,y=spidr] {\dataMedium};
\addplot[spidr2 slm] table[x=bin,y=spidr2] {\dataMedium};
\addplot[gslm2] table[x=bin,y=gslm] {\dataMedium};

\nextgroupplot[
    ylabel={},
]
\addplot[text lm] table[x=bin,y=text] {\dataLarge};
\addplot[spidr slm] table[x=bin,y=spidr] {\dataLarge};
\addplot[spidr2 slm] table[x=bin,y=spidr2] {\dataLarge};
\addplot[gslm2] table[x=bin,y=gslm] {\dataLarge};

\end{groupplot}

\end{tikzpicture}
\begin{tikzpicture}

\definecolor{textgray}{RGB}{80,80,80}
\definecolor{spidrgreen}{RGB}{89,161,79}
\definecolor{spidr2blue}{RGB}{78,121,167}
\definecolor{gslm2red}{RGB}{225,87,89}

\pgfdeclareplotmark{filled star}{%
  \pgfpathmoveto{\pgfqpoint{0pt}{\pgfplotmarksize}}%
  \pgfpathlineto{\pgfqpoint{.224514\pgfplotmarksize}{.309017\pgfplotmarksize}}%
  \pgfpathlineto{\pgfqpoint{.951057\pgfplotmarksize}{.309017\pgfplotmarksize}}%
  \pgfpathlineto{\pgfqpoint{.363271\pgfplotmarksize}{-.118034\pgfplotmarksize}}%
  \pgfpathlineto{\pgfqpoint{.587785\pgfplotmarksize}{-.809017\pgfplotmarksize}}%
  \pgfpathlineto{\pgfqpoint{0pt}{-.381966\pgfplotmarksize}}%
  \pgfpathlineto{\pgfqpoint{-.587785\pgfplotmarksize}{-.809017\pgfplotmarksize}}%
  \pgfpathlineto{\pgfqpoint{-.363271\pgfplotmarksize}{-.118034\pgfplotmarksize}}%
  \pgfpathlineto{\pgfqpoint{-.951057\pgfplotmarksize}{.309017\pgfplotmarksize}}%
  \pgfpathlineto{\pgfqpoint{-.224514\pgfplotmarksize}{.309017\pgfplotmarksize}}%
  \pgfpathclose
  \pgfusepath{fill,stroke}%
}

\pgfplotstableread{
bin text spidr spidr2 gslm
0   60.0  53.9  53.2  59.0
1   59.2  53.7  54.6  59.0
2   60.2  52.5  53.9  58.5
3   60.9  54.3  55.5  58.6
4   62.8  54.4  55.1  58.6
5   61.5  54.4  56.6  58.6
6   60.5  51.7  53.8  56.2
7   61.7  54.4  58.4  62.6
8   61.9  53.5  56.9  59.5
9   66.4  53.9  54.6  62.3
10  69.4  59.0  52.1  56.9
}\dataSmall

\pgfplotstableread{
bin text spidr spidr2 gslm
0   75.6  57.6  60.9  66.8
1   77.4  56.8  61.3  65.9
2   76.9  55.9  61.9  66.3
3   78.1  57.7  63.1  67.9
4   79.4  58.4  63.9  69.1
5   78.6  58.9  65.1  69.8
6   80.8  58.7  65.5  70.9
7   82.6  59.3  66.7  71.7
8   81.1  59.0  67.6  72.1
9   81.5  59.3  67.8  72.1
10  78.2  61.4  69.6  74.9
}\dataMedium

\pgfplotstableread{
bin text spidr spidr2 gslm
0   84.6  59.4  64.0  67.1
1   83.1  58.7  63.2  66.4
2   82.3  59.2  64.8  68.1
3   84.4  60.0  66.0  68.6
4   84.0  59.7  65.1  67.7
5   84.8  60.1  65.5  69.6
6   86.0  60.9  66.5  69.7
7   87.7  61.3  67.7  71.5
8   87.6  62.0  69.0  72.6
9   87.3  62.4  69.2  72.8
10  88.0  63.6  71.5  75.0
}\dataLarge

\pgfplotsset{
  comparison axis/.style={
    width=5.35cm, height=4cm,
    xmin=-0.35, xmax=10.35,
    ymin=50, ymax=100,
    xtick={0,1,2,3,4,5,6,7,8,9,10},
    xticklabels={
      {$[0]$}, {$[1,2)$}, {$[2,4)$}, {$[4,8)$}, {$[8,16)$},
      {$[16,32)$}, {$[32,64)$}, {$[64,128)$},
      {$[128,256)$}, {$[256,512)$}, {$[512,\infty)$}
    },
    ytick={50,60,70,80,90,100},
    minor ytick={55,65,75,85,95},
    xlabel={\small Frequency Bin},
    ylabel={\small AS Accuracy (\%)},
    xticklabel style={font=\scriptsize,rotate=60,anchor=east},
    yticklabel style={font=\scriptsize,xshift=2pt},
    xlabel style={font=\scriptsize,yshift=1pt},
    ylabel style={font=\scriptsize},
    title style={font=\small},
    tick align=outside, tick style={black!70},
    axis lines=left,
    axis line style={black!80,line width=0.9pt},
    xmajorgrids=true, ymajorgrids=true, yminorgrids=true,
    major grid style={
      draw=black!40,dash pattern=on 1pt off 1.4pt,line width=0.3pt
    },
    minor grid style={
      draw=black!40,dash pattern=on 1pt off 1.4pt,line width=0.3pt
    },
    line width=0.8pt, mark size=1.5pt,
    every axis plot/.append style={solid,mark options={solid}},
    clip=false
  },
  text lm/.style={
    color=textgray,mark=*,
    mark options={solid,fill=textgray,draw=textgray}
  },
  spidr slm/.style={
    color=spidrgreen,mark=triangle*,mark size=2pt,
    mark options={solid,fill=spidrgreen,draw=spidrgreen}
  },
  spidr2 slm/.style={
    color=spidr2blue,mark=square*,
    mark options={solid,fill=spidr2blue,draw=spidr2blue}
  },
  gslm2/.style={
    color=gslm2red,mark=filled star,mark size=2pt,
    mark options={solid,sharp corners,fill=gslm2red,draw=gslm2red}
  }
}

\begin{groupplot}[
  comparison axis,
  group style={group size=3 by 1,horizontal sep=0.7cm}
]

\nextgroupplot[
  legend to name=binnedsharedlegend,
  legend columns=4,
  legend cell align=left,
  legend style={
    font=\small,draw=black!50,rounded corners=2pt,
    line width=0.8pt,fill=none,column sep=4pt
  }
]
\addplot[text lm] table[x=bin,y=text] {\dataSmall};
\addlegendentry{Text LM}
\addplot[spidr slm] table[x=bin,y=spidr] {\dataSmall};
\addlegendentry{SpidR SLM}
\addplot[spidr2 slm] table[x=bin,y=spidr2] {\dataSmall};
\addlegendentry{SpidR 2.0 SLM}
\addplot[gslm2] table[x=bin,y=gslm] {\dataSmall};
\addlegendentry{GSLM2}

\nextgroupplot[
    ylabel={}
]
\addplot[text lm] table[x=bin,y=text] {\dataMedium};
\addplot[spidr slm] table[x=bin,y=spidr] {\dataMedium};
\addplot[spidr2 slm] table[x=bin,y=spidr2] {\dataMedium};
\addplot[gslm2] table[x=bin,y=gslm] {\dataMedium};

\nextgroupplot[
    ylabel={},
]
\addplot[text lm] table[x=bin,y=text] {\dataLarge};
\addplot[spidr slm] table[x=bin,y=spidr] {\dataLarge};
\addplot[spidr2 slm] table[x=bin,y=spidr2] {\dataLarge};
\addplot[gslm2] table[x=bin,y=gslm] {\dataLarge};

\end{groupplot}

\node[anchor=north,yshift=-2.1cm] at (group c2r1.south) {
  \pgfplotslegendfromname{binnedsharedlegend}
};
\end{tikzpicture}
\vspace{-0.4cm}
\caption{LT-Swap accuracy on WordSwap (WS), InflectionSwap (IS), and AgreementSwap (AS) by word-frequency bin at the three Libri-Light training scales. Chance level: 50\%.}
\label{fig:ltswap_per_bin}
\end{figure}

To better understand the remaining LT-Swap gaps to Text LM in Table~\ref{tab:main_results},
we examine whether these gaps are concentrated on low-frequency words.
Figure~\ref{fig:ltswap_per_bin} presents WS, IS, and AS performance by word frequency across all three Libri-Light training-data scales.
The aggregate SLM ranking largely holds within bins:
SpidR 2.0 SLM generally outperforms SpidR SLM,
while GSLM2 surpasses SpidR 2.0 SLM in every bin across all tasks and scales.
Thus, the gains from multi-stream modeling extend across the frequency distribution.

On WS, accuracy generally increases with word frequency at all three scales.
This association is weaker for SpidR SLM than for the other SLMs, particularly at 600 hours.
At this scale, the gap to Text LM generally widens with frequency for all three SLMs.
At 6K and 60K hours, the gap varies less across frequency bins.

IS also shows a positive association between frequency and accuracy,
but substantial gaps to Text LM persist throughout the distribution.
The IS deficit highlighted in the main results therefore extends beyond low-frequency items.

AS exhibits a less consistent frequency trend at 600 hours,
with SLM scores fluctuating across bins.
At 6K and 60K hours, accuracy generally increases with frequency,
but Text LM remains ahead in every bin.
Its advantage over SLMs is consistently larger at 60K than at 6K hours,
showing that the wider aggregate gap extends throughout the frequency distribution.

These results suggest that the limitations exposed by IS and AS are not confined to the lexical long tail.
Although sBLIMP also evaluates syntax and includes similar inflection and agreement contrasts,
its aggregate score combines a broader range of grammatical phenomena.
Moreover, the cascaded ASR-LM evaluation in Appendix~\ref{app:sssec:cascaded_asr_slm} reveals TTS artifacts that can bias performance on several sBLIMP subtasks.
Our spoken LT-Swap benchmark is less affected by these specific artifacts and contains substantially more sentence pairs on the specific categories.
Its results therefore provide complementary evidence that inflectional and agreement judgments remain challenging for SLMs, even for frequent word items.

\subsubsection{Generative Evaluation at Fixed Temperatures}
\label{app:sssec:gen_eval_fixed_temp}
\begin{table}[t]
    \centering
    \caption{Speech-continuation performance at fixed decoding temperatures $\tau\in\{1.0,0.8\}$, complementing the results with calibrated temperatures in Table~\ref{tab:gen_eval}. Oracle rows evaluate ground-truth continuations.}
    \label{tab:gen_eval_fixed_temp}

    \small
    \renewcommand{\arraystretch}{1.12}

    \begin{tabular}{clcccc}
    \toprule
    & & \multicolumn{2}{c}{\textbf{ASR-PPL} ($\downarrow$)} & \multicolumn{2}{c}{\textbf{LLMJ} ($\uparrow$)} \\
    \cmidrule(l{0.3em}r{0.3em}){3-4}\cmidrule(l{0.3em}r{0.3em}){5-6}
    \multicolumn{2}{l}{\textbf{Model}} & $\tau=1.0$ & $\tau=0.8$ & $\tau=1.0$ & $\tau=0.8$ \\
    \midrule

    \multirow{4}{*}[-3pt]{\rotatebox{90}{Oracle}}
    & Text LM & \multicolumn{2}{c}{55.2} & \multicolumn{2}{c}{4.55} \\
    \cmidrule(l{0.2em}r{0.2em}){2-6}
    & SpidR SLM & \multicolumn{2}{c}{60.1} & \multicolumn{2}{c}{4.54} \\
    & SpidR 2.0 SLM & \multicolumn{2}{c}{70.1} & \multicolumn{2}{c}{4.41} \\
    & GSLM2 & \multicolumn{2}{c}{59.5} & \multicolumn{2}{c}{4.53} \\
    \cmidrule(l{0.2em}r{0.2em}){1-6}

    \multirow{4}{*}[-3pt]{\rotatebox{90}{600\,h / 90M}}
    & Text LM & 229.1\,$\pm$\,5.8 & \phantom{0}76.8\,$\pm$\,2.4 & 2.08\,$\pm$\,0.01 & 2.22\,$\pm$\,0.02  \\
    \cmidrule(l{0.2em}r{0.2em}){2-6}
    & SpidR SLM & 355.8\,$\pm$\,9.3 & 232.6\,$\pm$\,4.0 & 1.91\,$\pm$\,0.01 & 2.01\,$\pm$\,0.01 \\
    & SpidR 2.0 SLM & 264.2\,$\pm$\,6.0 & 163.4\,$\pm$\,3.1 & 1.99\,$\pm$\,0.02 & 2.10\,$\pm$\,0.02 \\
    & GSLM2 & 244.9\,$\pm$\,7.9 & 188.0\,$\pm$\,4.2 & 2.02\,$\pm$\,0.01 & 2.08\,$\pm$\,0.03 \\
    \cmidrule(l{0.2em}r{0.2em}){1-6}

    \multirow{4}{*}[-2pt]{\rotatebox{90}{6K\,h / 300M}}
    & Text LM & 114.7\,$\pm$\,3.1 & \phantom{0}45.3\,$\pm$\,1.0 & 2.66\,$\pm$\,0.05 & 2.87\,$\pm$\,0.02 \\
    \cmidrule(l{0.2em}r{0.2em}){2-6}
    & SpidR SLM & 143.2\,$\pm$\,3.8 & \phantom{0}87.0\,$\pm$\,2.4 & 2.81\,$\pm$\,0.03 & 3.00\,$\pm$\,0.03 \\
    & SpidR 2.0 SLM & 141.9\,$\pm$\,1.4 & \phantom{0}79.9\,$\pm$\,2.7 & 2.74\,$\pm$\,0.04 & 2.94\,$\pm$\,0.03 \\
    & GSLM2 & \phantom{0}89.5\,$\pm$\,2.4 & \phantom{0}69.0\,$\pm$\,1.6 & 3.01\,$\pm$\,0.04 & 3.01\,$\pm$\,0.02 \\
    \cmidrule(l{0.2em}r{0.2em}){1-6}

    \multirow{4}{*}[-2pt]{\rotatebox{90}{60K\,h / 1.2B}}
    & Text LM & \phantom{0}66.2\,$\pm$\,1.5 & \phantom{0}30.3\,$\pm$\,0.5 & 3.27\,$\pm$\,0.04 & 3.51\,$\pm$\,0.02 \\
    \cmidrule(l{0.2em}r{0.2em}){2-6}
    & SpidR SLM & \phantom{0}67.7\,$\pm$\,1.4 & \phantom{0}45.8\,$\pm$\,0.8 & 3.53\,$\pm$\,0.02 & 3.65\,$\pm$\,0.04 \\
    & SpidR 2.0 SLM & \phantom{0}63.6\,$\pm$\,1.3 & \phantom{0}40.7\,$\pm$\,1.2 & 3.55\,$\pm$\,0.03 & 3.66\,$\pm$\,0.03 \\
    & GSLM2 & \phantom{0}51.1\,$\pm$\,1.6 & \phantom{0}40.6\,$\pm$\,0.8 & 3.60\,$\pm$\,0.03 & 3.59\,$\pm$\,0.02 \\
    \bottomrule
    \end{tabular}
\end{table}
The main results in Table~\ref{tab:gen_eval} use model-specific temperatures calibrated to target $\mathrm{NRS}=1$ on the development set (see Appendix~\ref{app:sssec:asr_ppl}).
Although this uses a common repetition target,
temperature calibration is not standard practice in typical LM inference.
Moreover, the resulting temperatures are relatively high, ranging from 1.20 to 1.59.
We therefore complement the calibrated evaluation with results at $\tau \in \{1.0,0.8\}$,
two settings closer to conventional inference, in Table~\ref{tab:gen_eval_fixed_temp}.

At both fixed temperatures,
every model and data scale obtains lower ASR-PPL and higher LLMJ than when using calibrated temperatures.
The differences can be substantial:
at 60K hours, GSLM2 improves from an ASR-PPL of 189.1 and an LLMJ score of 3.05 at its calibrated temperature $\hat{\tau}=1.59$ to 51.1 and 3.60, respectively, at $\tau=1.0$.
At this temperature, Text LM performs best on both metrics in the 600-hour setting,
while GSLM2 performs best at 6K and 60K hours,
showing that its advantage at larger scales persists under a conventional decoding setup.
At $\tau=0.8$, Text LM achieves the lowest ASR-PPL at every scale,
whereas the highest LLMJ score is obtained by the Text LM at 600 hours, GSLM2 at 6K hours, and SpidR 2.0 at 60K hours.
Thus, the fixed-temperature results provide a practically relevant complement to the comparison using calibrated temperatures while also showing that model rankings depend on the decoding temperature.

We also report oracle results obtained by evaluating the ground-truth continuations.
For Text LM, the ground-truth transcript is scored directly.
For each SLM, in contrast, the ground-truth speech is converted into the corresponding unit representation, resynthesized with its unit vocoder, and transcribed by ASR before scoring.
Consequently, the SLM oracles perform consistently worse than the Text LM oracle,
with higher ASR-PPL and lower LLMJ scores.
This gap captures degradation introduced by the multi-stage SLM evaluation pipeline,
meaning that direct comparisons between SLMs and Text LM are not strictly equivalent.
Oracle performance also varies across SLMs because their unit vocoders differ in resynthesis quality,
although SpidR and GSLM2 achieve comparable oracle scores.
Resynthesis artifacts increase ASR errors (see Appendix~\ref{app:ssec:speech_resynthesis}),
which propagate to both evaluation metrics.

A notable limitation of ASR-PPL appears at the lower fixed temperatures.
At 60K hours and $\tau=0.8$, every model obtains a lower ASR-PPL than its corresponding ground-truth oracle.
Because the oracle contains the actual ground-truth continuations,
these results cannot meaningfully indicate that the generated continuations are better than the ground truth,
particularly as their LLMJ scores remain substantially below the oracle scores.
Instead, they show that ASR-PPL can strongly favor the more predictable outputs produced by low-temperature decoding.
ASR-PPL should therefore be interpreted together with semantic-quality and diversity measurements,
even when evaluating models under conventional inference settings.

\subsubsection{Multilingual evaluation on VoxPopuli}
\label{app:sssec:slm_eval_voxpopuli}

To examine how speech-data quality and language affect SLM performance,
we train English, French, and German models on VoxPopuli.
For each language, all SLM variants are trained on 6K hours of speech,
using the same procedure as the 6K-hour Libri-Light models.
The corresponding Text LM is trained on ASR transcripts of the same data
(see Appendix~\ref{app:ssec:datasets_and_preprocessing} for details).
The English models are evaluated on the same benchmarks used in the Libri-Light experiments,
whereas the French and German models are evaluated on our language-specific adaptations of sBLIMP and tSC (see Appendices~\ref{app:sssec:zerospeech21} and~\ref{app:sssec:storycloze}).

\begin{table}[t]
    \centering
    \caption{Results for SLMs trained on 6K-hour English, French, and German data from VoxPopuli. Text LMs serve as toplines, which are trained using ASR transcriptions. Scores are accuracies (\%; chance level: 50\%).}
    \label{tab:slm_eval_voxpopuli}

    \small
    \renewcommand{\arraystretch}{1.12}

    \begin{tabular}{llccc}
    \toprule
    \textbf{Language} & \textbf{Model} & \textbf{sWuggy} & \textbf{sBLIMP} & \textbf{tSC} \\
    \midrule

    \multirow{4}{*}[-2pt]{English}
    & Text LM & 77.0 & 63.9 & 76.8 \\
    \cmidrule(l{0.2em}r{0.2em}){2-5}
    & SpidR SLM & 60.3 & 53.3 & 69.3 \\
    & SpidR 2.0 SLM & 62.9 & 55.3 & 69.6 \\
    & GSLM2 & \textbf{65.6} & \textbf{57.2} & \textbf{72.7} \\
    \cmidrule(l{0.3em}r{0.3em}){1-5}

    \multirow{4}{*}[-2pt]{French}
    & Text LM & -- & 95.0 & 75.7 \\
    \cmidrule(l{0.2em}r{0.2em}){2-5}
    & SpidR SLM & -- & 58.8 & 63.9 \\
    & SpidR 2.0 SLM & -- & 66.0 & 70.2 \\
    & GSLM2 & -- & \textbf{69.4} & \textbf{72.0} \\
    \cmidrule(l{0.3em}r{0.3em}){1-5}

    \multirow{4}{*}[-2pt]{German}
    & Text LM & -- & 88.9 & 74.8 \\
    \cmidrule(l{0.2em}r{0.2em}){2-5}
    & SpidR SLM & -- & 56.2 & 63.4 \\
    & SpidR 2.0 SLM & -- & 67.1 & 68.9 \\
    & GSLM2 & -- & \textbf{72.3} & \textbf{71.4} \\
    \bottomrule
    \end{tabular}
\end{table}

Table~\ref{tab:slm_eval_voxpopuli} reports the results.
Across all three languages, the SLMs exhibit the same ranking observed on Libri-Light:
SpidR 2.0 consistently outperforms SpidR, while GSLM2 yields further gains.
This consistency supports the generalizability of the SpidR 2.0 encoder and the multi-stream modeling approach across languages.
Appendix~\ref{app:sssec:discophon} further investigates the multilingual capabilities of the SpidR 2.0 encoder.

Compared with the corresponding 6K-hour Libri-Light results in Table~\ref{tab:main_results},
the English VoxPopuli setting exhibits a wider performance gap between GSLM2 and Text LM,
particularly on sWuggy and sBLIMP.
The gap increases from 3.2 to 11.4 percentage points on sWuggy and from -0.4 to 6.7 percentage points on sBLIMP,
whereas the tSC gap changes only slightly, from 5.9 to 4.1 percentage points.
Because sWuggy and sBLIMP may be particularly sensitive to the preservation of fine-grained phonetic information,
the selective widening of these gaps suggests that noisier training speech can reduce data efficiency in spoken language modeling.
We attribute this pattern to the more challenging recording conditions and greater acoustic heterogeneity of VoxPopuli relative to the comparatively clean Libri-Light corpus.

\subsubsection{Cascaded ASR-LM Evaluation}
\label{app:sssec:cascaded_asr_slm}
\begin{table}[t]
    \centering
    \caption{Performance of Text LMs and GSLM2 (Libri-Light 6K\,h; 300M parameters) under different training and evaluation conditions. Text LMs are trained on either ground-truth or ASR transcripts. In the cascaded setting, spoken stimuli are transcribed and the resulting transcripts are scored by the Text LM. GSLM2 is trained and evaluated directly on speech. Scores are accuracies (\%; chance level: 50\%).}
    \label{tab:cascade}

    \small
    \renewcommand{\arraystretch}{1.12}

    \begin{tabular}{lllccc}
    \toprule
    & & & \multicolumn{2}{c}{\textbf{ZS21} ($\uparrow$)} & \textbf{SC} ($\uparrow$) \\
    \cmidrule(l{0.3em}r{0.3em}){4-5}\cmidrule(l{0.3em}r{0.3em}){6-6}
    \textbf{Model} & \textbf{Training Input} & \textbf{Evaluation Input}
    & (s)Wuggy & (s)BLIMP & tSC \\
    \midrule

    \multirow{4}{*}[-2pt]{Text LM}
    & \multirow{2}{*}{Ground-truth text}
    & \texttt{B1}: Ground-truth text             & 79.5 & 66.8 & 84.8 \\
    & & \texttt{B2}: ASR transcript of speech    & 70.6 & 71.0 & 75.4 \\
    \cmidrule(lr){2-6}
    & \multirow{2}{*}{ASR transcripts}
    & \texttt{B3}: Ground-truth text             & 81.2 & 66.9 & 85.1 \\
    & & \texttt{B4}: ASR transcript of speech    & 71.5 & 69.1 & 77.4 \\
    \midrule

    GSLM2 & Speech & Speech          & 76.3 & 67.2 & 78.9 \\

    \bottomrule
    \end{tabular}
\end{table}
We evaluate a cascaded ASR-LM approach to spoken language modeling,
in which spoken evaluation stimuli are first transcribed by an ASR system and the resulting transcripts are scored by Text LM.
We consider Text LM trained on either ground-truth or ASR-generated transcripts and compare them with GSLM2,
which is trained and evaluated directly on speech.
We use Whisper large-v3 as the ASR frontend.
Table~\ref{tab:cascade} reports results for the 6K-hour Libri-Light setting with 300M parameters.
The results for Text LM trained and evaluated on ground-truth text (\texttt{B1}),
as well as those for GSLM2, are identical to the corresponding results in Table~\ref{tab:main_results}.
Evaluation on ASR transcripts (\texttt{B2} and \texttt{B4}) measures
performance on spoken stimuli from sWuggy, sBLIMP, and tSC.
Evaluation on ground-truth text (\texttt{B1} and \texttt{B3}) measures
performance on the corresponding textual inputs:
Wuggy and BLiMP items and ground-truth tSC transcripts built from StoryCloze.

For the Text LM trained on ground-truth text,
replacing ground-truth evaluation text with ASR transcripts substantially degrades performance on sWuggy and tSC (\texttt{B1} vs.\ \texttt{B2}).
On sWuggy, this is partly due to the difficulty of transcribing nonwords: 
Whisper may map an unfamiliar nonword to a plausible word or phrase,
thereby changing the lexical contrast presented to the LM.
On tSC, Whisper can produce different transcriptions of the first four sentences shared by the positive and negative stories when each full story is transcribed independently.
These inconsistencies introduce score variation unrelated to the final continuation,
making it harder for the LM to identify the more plausible story.
In contrast, performance improves on sBLIMP when the same model is evaluated on ASR transcripts. 
We find that this improvement is partly driven by several artifacts introduced by the TTS system used to construct sBLIMP.
For example, certain words in the ungrammatical distractors, such as ``WHO,''
are retained in uppercase in the TTS input and pronounced letter by letter.
The resulting unnatural pronunciations provide unintended cues associated with negative examples.
When reflected in ASR transcripts, these cues can make the discrimination task easier, 
independently of the grammatical contrast being tested.
Such artifacts may also explain why GSLM2 outperforms Text LM on sBLIMP in Table~\ref{tab:main_results}.

\begin{figure}[t]
\centering
\begin{tikzpicture}

\definecolor{textgray}{RGB}{80,80,80}
\definecolor{cascadepurple}{RGB}{148,103,189}
\definecolor{gslm2red}{RGB}{225,87,89}

\pgfdeclareplotmark{filled star}{%
  \pgfpathmoveto{\pgfqpoint{0pt}{\pgfplotmarksize}}%
  \pgfpathlineto{\pgfqpoint{.224514\pgfplotmarksize}{.309017\pgfplotmarksize}}%
  \pgfpathlineto{\pgfqpoint{.951057\pgfplotmarksize}{.309017\pgfplotmarksize}}%
  \pgfpathlineto{\pgfqpoint{.363271\pgfplotmarksize}{-.118034\pgfplotmarksize}}%
  \pgfpathlineto{\pgfqpoint{.587785\pgfplotmarksize}{-.809017\pgfplotmarksize}}%
  \pgfpathlineto{\pgfqpoint{0pt}{-.381966\pgfplotmarksize}}%
  \pgfpathlineto{\pgfqpoint{-.587785\pgfplotmarksize}{-.809017\pgfplotmarksize}}%
  \pgfpathlineto{\pgfqpoint{-.363271\pgfplotmarksize}{-.118034\pgfplotmarksize}}%
  \pgfpathlineto{\pgfqpoint{-.951057\pgfplotmarksize}{.309017\pgfplotmarksize}}%
  \pgfpathlineto{\pgfqpoint{-.224514\pgfplotmarksize}{.309017\pgfplotmarksize}}%
  \pgfpathclose
  \pgfusepath{fill,stroke}%
}

\pgfplotstableread{
bin text cascade gslm
0  67.6  63.8  65.0
1  69.7  66.7  65.8
2  71.5  71.6  69.0
3  76.1  74.8  71.5
4  77.9  79.3  72.7
5  81.0  84.1  76.1
6  84.8  87.6  80.5
7  88.9  91.0  85.8
8  90.0  91.6  82.5
9  93.4  94.4  87.2
}\dataSmall

\pgfplotstableread{
bin text cascade gslm
0   85.8  64.0  71.6
1   86.3  65.0  70.8
2   87.6  66.7  73.6
3   89.7  67.5  73.1
4   90.6  67.8  74.2
5   91.8  68.9  75.9
6   92.8  68.0  76.6
7   93.5  70.2  78.0
8   93.4  69.6  78.9
9   93.8  69.4  78.3
10  94.4  66.4  79.7
}\dataMedium

\pgfplotstableread{
bin text cascade gslm
0   75.6  61.6  66.8
1   77.4  63.2  65.9
2   76.9  64.7  66.3
3   78.1  65.2  67.9
4   79.4  67.1  69.1
5   78.6  69.0  69.8
6   80.8  68.5  70.9
7   82.6  70.4  71.7
8   81.1  71.4  72.1
9   81.5  71.7  72.1
10  78.2  73.1  74.9
}\dataLarge

\pgfplotsset{
  comparison axis/.style={
    width=5.35cm, height=4cm,
    xmin=-0.35,
    ymin=50, ymax=100,
    ytick={50,60,70,80,90,100},
    minor ytick={55,65,75,85,95},
    ylabel={\small Accuracy (\%)},
    xticklabel style={font=\scriptsize,rotate=60,anchor=east},
    yticklabel style={font=\scriptsize,xshift=2pt},
    xlabel style={font=\scriptsize,yshift=1pt},
    ylabel style={font=\scriptsize},
    title style={font=\small},
    tick align=outside, tick style={black!70},
    axis lines=left,
    axis line style={black!80,line width=0.9pt},
    xmajorgrids=true, ymajorgrids=true, yminorgrids=true,
    major grid style={
      draw=black!40,dash pattern=on 1pt off 1.4pt,line width=0.3pt
    },
    minor grid style={
      draw=black!40,dash pattern=on 1pt off 1.4pt,line width=0.3pt
    },
    line width=0.8pt, mark size=1.5pt,
    every axis plot/.append style={solid,mark options={solid}},
    clip=false
  },
  text lm/.style={
    color=textgray,mark=*,
    mark options={solid,fill=textgray,draw=textgray}
  },
  cascade slm/.style={
    color=cascadepurple,mark=pentagon*,mark size=2pt,
    mark options={solid,fill=cascadepurple,draw=cascadepurple}
  },
  gslm2/.style={
    color=gslm2red,mark=filled star,mark size=2pt,
    mark options={solid,sharp corners,fill=gslm2red,draw=gslm2red}
  }
}

\begin{groupplot}[
  comparison axis,
  group style={group size=3 by 1,horizontal sep=0.7cm}
]

\nextgroupplot[
  title={WordSwap (WS)},
    xtick={0,1,2,3,4,5,6,7,8,9},
    xticklabels={
      {$[1,2)$}, {$[2,4)$}, {$[4,8)$}, {$[8,16)$},
      {$[16,32)$}, {$[32,64)$}, {$[64,128)$},
      {$[128,256)$}, {$[256,512)$}, {$[512,\infty)$}
    },
    xmax=9.35,
  legend to name=binnedsharedlegend2,
  legend columns=3,
  legend cell align=left,
  legend style={
    font=\small,draw=black!50,rounded corners=2pt,
    line width=0.8pt,fill=none,column sep=4pt
  }
]
\addplot[text lm] table[x=bin,y=text] {\dataSmall};
\addlegendentry{Text LM (\texttt{B1})}
\addplot[cascade slm] table[x=bin,y=cascade] {\dataSmall};
\addlegendentry{Cascaded ASR-LM (\texttt{B4})}
\addplot[gslm2] table[x=bin,y=gslm] {\dataSmall};
\addlegendentry{GSLM2}

\nextgroupplot[
    title={InflectionSwap (IS)},
    ylabel={},
    xtick={0,1,2,3,4,5,6,7,8,9,10},
    xticklabels={
      {$[0]$}, {$[1,2)$}, {$[2,4)$}, {$[4,8)$}, {$[8,16)$},
      {$[16,32)$}, {$[32,64)$}, {$[64,128)$},
      {$[128,256)$}, {$[256,512)$}, {$[512,\infty)$}
    },
    xmax=10.35,
]
\addplot[text lm] table[x=bin,y=text] {\dataMedium};
\addplot[cascade slm] table[x=bin,y=cascade] {\dataMedium};
\addplot[gslm2] table[x=bin,y=gslm] {\dataMedium};

\nextgroupplot[
    title={AgreementSwap (AS)},
    ylabel={},
    xtick={0,1,2,3,4,5,6,7,8,9,10},
    xticklabels={
      {$[0]$}, {$[1,2)$}, {$[2,4)$}, {$[4,8)$}, {$[8,16)$},
      {$[16,32)$}, {$[32,64)$}, {$[64,128)$},
      {$[128,256)$}, {$[256,512)$}, {$[512,\infty)$}
    },
    xmax=10.35,
]
\addplot[text lm] table[x=bin,y=text] {\dataLarge};
\addplot[cascade slm] table[x=bin,y=cascade] {\dataLarge};
\addplot[gslm2] table[x=bin,y=gslm] {\dataLarge};

\end{groupplot}

\node[anchor=north,yshift=-1.4cm] at (group c2r1.south) {
  \pgfplotslegendfromname{binnedsharedlegend2}
};
\end{tikzpicture}
\vspace{-0.2cm}
\caption{LT-Swap accuracy by word-frequency bin, comparing cascaded ASR-LM system (\texttt{B4} in Table~\ref{tab:cascade}) with Text LM and GSLM2. Chance level: 50\%.}
\label{fig:ltswap_cascade}
\end{figure}
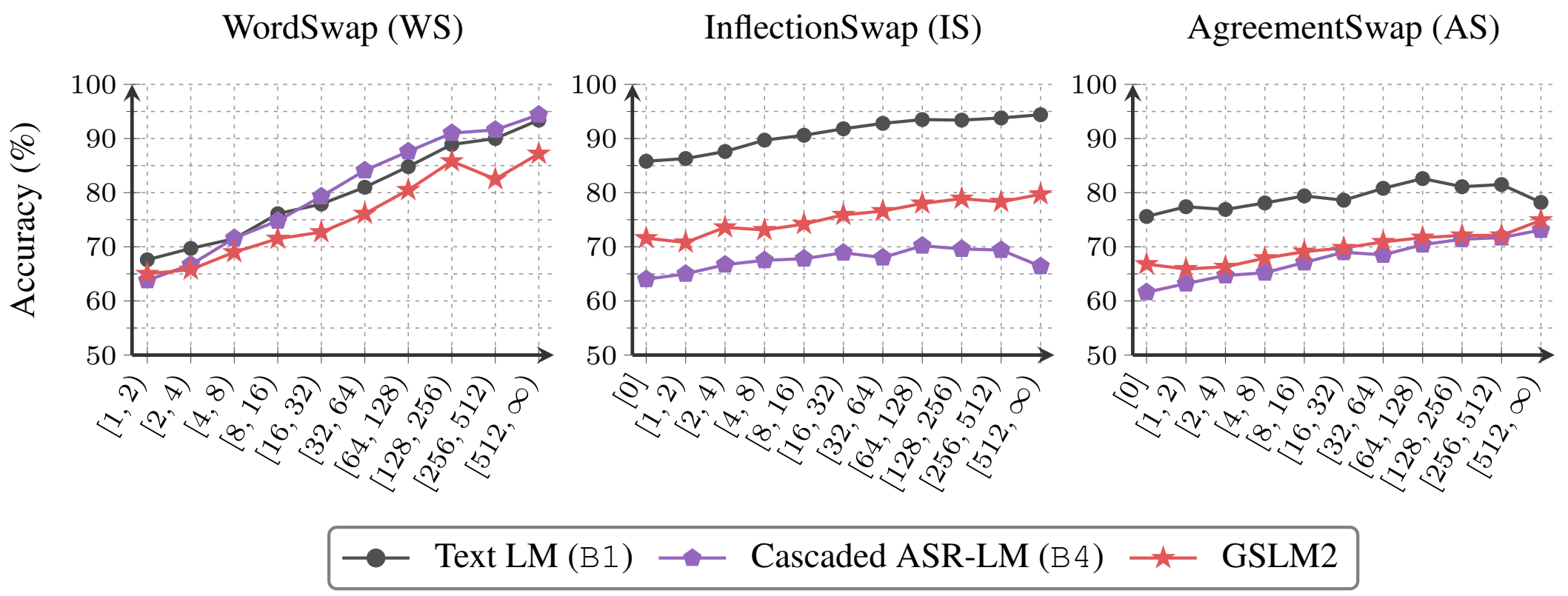

Training Text LM on ASR transcripts yields a modest improvement on sWuggy,
while performance on sBLIMP and tSC remains similar (\texttt{B1} vs.\ \texttt{B3}).
When the LM is trained on ASR transcripts and evaluated on ASR-transcribed speech (\texttt{B4}),
the resulting system forms a fully cascaded ASR-LM pipeline,
with ASR serving as the frontend to the LM.
Compared with this cascaded system,
GSLM2 performs substantially better on sWuggy and tSC.
Figure~\ref{fig:ltswap_cascade} extends this comparison to LT-Swap,
showing results for the cascaded system (\texttt{B4}), Text LM (\texttt{B1}), and GSLM2
across word-frequency bins.
Average WS accuracy is similar for Text LM and the cascade (80.1\% and 80.5\%).
On IS and AS, however, the cascade performs substantially worse than Text LM
and is also outperformed by GSLM2.
This degradation is partly attributable to Whisper altering the intended minimal contrasts,
particularly in ungrammatical examples involving inflection or agreement.
Together, these results suggest an advantage to modeling speech directly,
as a direct SLM avoids errors and biases introduced by an intermediate ASR representation.

\subsection{Supplementary Encoder Evaluation}
\label{app:ssec:supplementary_encoder_evaluation}

This section presents supplementary experiments and analyses of SpidR 2.0's continuous representations and learned discrete units.
We characterize codebook usage, unit duration, bitrate, and phone-code alignment (Appendices~\ref{app:sssec:codebook-metrics} and~\ref{app:sssec:phone-code-assignments}),
and examine how regularization and training dynamics shape the compact code vocabularies of SpidR 2.0 (Appendices~\ref{app:sssec:regularization} and~\ref{app:sssec:training-curves}).
We perform progressive ablations to assess the contributions of individual components (Appendix~\ref{app:sssec:spidr2_ablation})
and compare learned codes with post-hoc $k$-means quantization (Appendix~\ref{app:sssec:codes-vs-kmeans}).
We also analyze similarities between streams from different encoder layers (Appendix~\ref{app:sssec:stream-similarities}),
compare a broad range of speech encoders on ABX and MAP (Appendix~\ref{app:ssec:literature-encoder-evals}),
and examine how proxy metrics correlate with SLM performance (Appendix~\ref{app:sssec:proxy-target-correlations}).
Finally, we extend the evaluation to a multilingual setup on DiscoPhon (Appendix~\ref{app:sssec:discophon})
and supervised probing tasks on SUPERB (Appendix~\ref{app:sssec:superb}).

\subsubsection{Codebook Metrics}
\label{app:sssec:codebook-metrics}

\begin{figure}[t]
\centering
\includegraphics[width=0.95\textwidth]{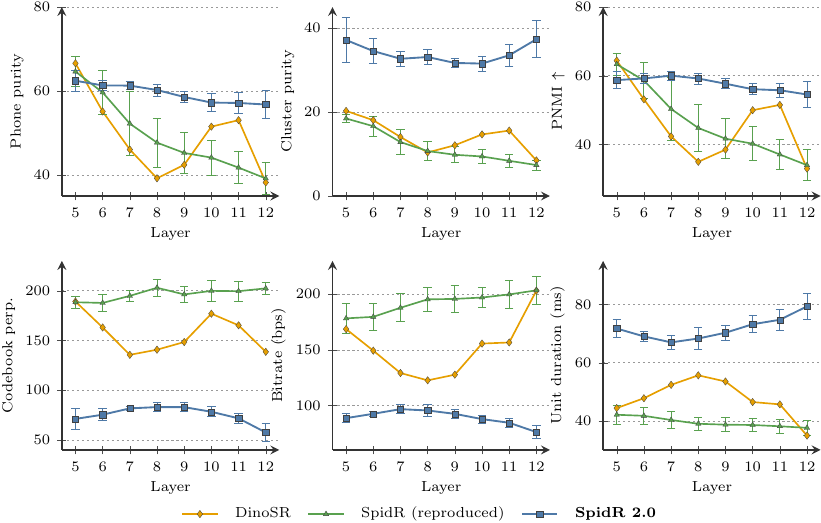}
\caption{Layerwise codebook quality on ZeroSpeech'21 \texttt{dev-clean}. DinoSR uses the official released checkpoint; SpidR and SpidR 2.0 show mean $\pm$ standard deviation over eight training seeds. Purity and PNMI are in \%.}
\label{fig:codebook-layerwise}
\end{figure}

For each encoder layer, we align the discrete codes with forced phone alignments on the ZeroSpeech'21 \texttt{dev-clean} set.
Let $\mathsf{P}$ and $\mathsf{C}$ denote the corresponding frame-level phone and code random variables, respectively.
We compute their empirical joint distribution $\mathbf{P} = [P(p,c)]$,
where $P(p,c)=\Pr(\mathsf{P}=p,\mathsf{C}=c)$.
The resulting unit assignments are characterized using six metrics in three groups:
\begin{itemize}[leftmargin=1.5em,topsep=0pt]
    \item \textbf{Diversity.} 
    Codebook perplexity is $\mathrm{PPL} = 2^{H(\mathsf{C})}$, where $H(\mathsf{C}) = -\sum_c P(c)\log_2 P(c)$ is the marginal entropy of code usage on this evaluation split,
    measured in bits, and $P(c)=\sum_p P(p,c)$.
    It measures the effective number of codes used and is upper-bounded by the codebook size (256 here).
    \item \textbf{Phone-code alignment (purity, PNMI).} 
    Phone purity, $\sum_c \max_p P(p,c)$, is the fraction of frames whose phone label matches the most frequent phone associated with their assigned code (how consistently a code is assigned to a single phone).
    Cluster purity, $\sum_p \max_c P(p,c)$, is the fraction of frames assigned to the most frequent code for their phone (how strongly a phone is concentrated in a single code). 
    Phone-normalized mutual information (PNMI), $I(\mathsf{P};\mathsf{C})/H(\mathsf{P})$, is the mutual information between phones and codes, normalized by phone entropy.
    It is important to note that all three metrics are heavily dependent on codebook usage: using more codes can increase phone purity and PNMI while reducing cluster purity.
    They should therefore be interpreted alongside perplexity.
    \item \textbf{Compression (bitrate, unit duration).} 
    Both compression metrics are computed on the run-length deduplicated code sequence (consecutive identical codes collapsed to a single unit segment). 
    Using a definition similar to that in~\cite{chang2025dc}, we compute bitrate as $(N/T)\,H_{\mathsf{dedup}}$ bits per second (bps), where $N$ is the number of segments, $T$ is the total duration in seconds, and $H_{\mathsf{dedup}}$ is the entropy, in bits, of the deduplicated code distribution.
    Unit duration (in ms) is the average span length (i.e., feature stride $\times$ total frames $/$ total segments).
\end{itemize}

\paragraph{Observations.}
Figure~\ref{fig:codebook-layerwise} reveals three complementary trends.
\textbf{(i) Diversity}:
SpidR 2.0 uses less than half as many effective codes per layer as SpidR,
matching the drop in active codes observed during training (see Figure~\ref{fig:training-curves}). 
\textbf{(ii) Purity / PNMI}:
Although SpidR keeps roughly constant codebook usage over layers, both  purity metrics and PNMI degrade with depth. This resonates with the poorer performance on downstream metrics observed in Figure~\ref{fig:layerwise}.
SpidR 2.0 achieves substantially higher cluster purity at every layer, 
indicating that a larger fraction of frames map to the dominant code for their phone.
Together with its lower codebook perplexity, this suggests a more compact representation.
Although SpidR 2.0 has lower phone purity and PNMI than SpidR at their respective best layers,
it outperforms SpidR on all other downstream metrics.
\textbf{(iii) Compression}:
SpidR 2.0 produces longer units (${\sim}$70 to 90\,ms vs.\ ${\sim}$40\,ms) at roughly half the bitrate.
This is closer to mean phone duration in this dataset:
93\,ms including silence and 82\,ms excluding silence.
Together, these trends indicate that SpidR 2.0 learns coarser units without sacrificing phone-level identifiability.

\subsubsection{Phone-Code Assignments}
\label{app:sssec:phone-code-assignments}

\begin{figure}[t]
\centering
\includegraphics[width=0.9\textwidth]{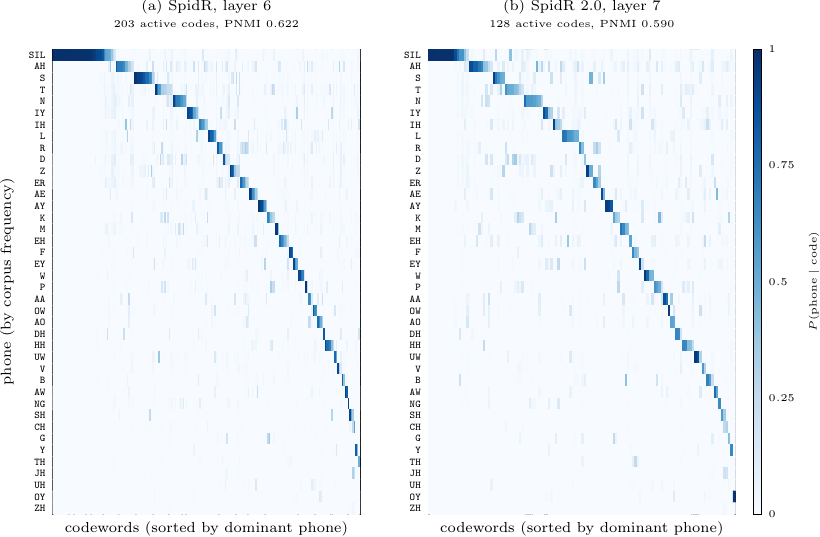}
\caption{Phone content of the discrete units. Conditional distribution $P(\mathrm{phone}\mid\mathrm{code})$ on LibriSpeech \texttt{dev-clean},
computed using forced alignments at 10\,ms and discrete codes at 20\,ms.
Rows are the 40 phone labels ordered by corpus frequency (\texttt{SIL} first);
columns are the active codewords only (203 in (a) SpidR and 128 in (b) SpidR 2.0), grouped by their dominant phone in row order and sorted within each group by the conditional probability of that phone. Concentration on a single phone within each column indicates high phone purity; the column ordering arranges these peaks into a staircase pattern.}
\label{fig:proba-phone-code}
\end{figure}

Figure~\ref{fig:proba-phone-code} shows the conditional distribution $P(p \mid c)$ over phone labels $p$ for each code $c$.
The concentration of these distributions around dominant phones is consistent with the high phone purity of SpidR 2.0 reported in Figure~\ref{fig:codebook-layerwise}.
If we assign each code to its dominant phone,
nine phones account for 50\% of the total codebook usage in SpidR 2.0,
compared with ten in SpidR.
Codes assigned to silence constitute over 20\% of SpidR's codebook, compared with 13\% for SpidR 2.0,
indicating that SpidR 2.0 allocates a smaller share of its codebook to silence.

\subsubsection{Ablation Studies}
\label{app:sssec:spidr2_ablation}

\begin{table}[t]
\centering
\caption{ABX ablation results (\%; lower is better), averaged over \texttt{dev-clean} and \texttt{dev-other} and over within- and across-speaker conditions. Modifications are cumulative. Values are mean $\pm$ standard deviation across eight seeds of encoder training. For each variant, $l^*$ denotes the layer with the lowest task-averaged ABX error. $l=5$ and $l=10$ provide fixed-layer comparisons.}
\label{tab:ablation-abx}
\small
\renewcommand{\arraystretch}{1.12}
\begin{tabular}{l c c c c}
\toprule
\textbf{Variant} & \textbf{Best layer $l^*$} & $l = l^*$ & $l=5$ & $l=10$ \\
\midrule
SpidR (reproduced) & 6 & 5.16 $\pm$ 1.37 & 5.25 $\pm$ 0.76 & 10.90 $\pm$ 2.97 \\
+ shared heads & 6 & 4.82 $\pm$ 0.57 & 5.75 $\pm$ 0.74 & 6.78 $\pm$ 1.02 \\
+ regularization & 9 & 4.12 $\pm$ 0.34 & 8.75 $\pm$ 0.75 & 4.29 $\pm$ 0.42 \\
+ normalizations & 10 & 3.58 $\pm$ 0.31 & 5.88 $\pm$ 0.27 & 3.58 $\pm$ 0.31 \\
+ $\mathrm{LR}=2.0\times10^{-4}$ (\textbf{SpidR 2.0}) & 10 & \textbf{2.88 $\pm$ 0.12} & \textbf{4.73 $\pm$ 0.22} & \textbf{2.88 $\pm$ 0.12} \\
\bottomrule
\end{tabular}
\end{table}

\begin{table}[t]
\centering
\caption{MAP ablation results (\%; higher is better). Modifications are cumulative. Values are mean $\pm$ standard deviation across eight seeds of encoder training. For each variant, all scores are reported at the layer with the highest average MAP score.}
\label{tab:ablation-map}
\small
\renewcommand{\arraystretch}{1.12}
\begin{tabular}{l c c c c}
\toprule
\textbf{Variant} & \textbf{Best layer} & \textbf{Average} & \textbf{dev-clean} & \textbf{dev-other} \\
\midrule
SpidR (reproduced) & 4 & 58.3 $\pm$ 2.5 & 65.2 $\pm$ 3.0 & 51.4 $\pm$ 3.9 \\
+ shared heads & 5 & 61.3 $\pm$ 3.4 & 66.2 $\pm$ 5.4 & 56.4 $\pm$ 4.1 \\
+ regularization & 7 & 68.0 $\pm$ 1.6 & 73.3 $\pm$ 2.2 & 62.7 $\pm$ 2.4 \\
+ normalizations & 8 & \textbf{72.3 $\pm$ 0.3} & \textbf{77.0 $\pm$ 0.3} & \textbf{67.5 $\pm$ 0.4} \\
+ $\mathrm{LR}=2.0\times10^{-4}$ (\textbf{SpidR 2.0}) & 7 & 69.2 $\pm$ 0.3 & 75.3 $\pm$ 0.4 & 63.1 $\pm$ 0.5 \\
\bottomrule
\end{tabular}
\end{table}

\begin{table}[t]
\centering
\caption{Ablation results for downstream SLM metrics: sWuggy, sBLIMP, and tSC (\%; higher is better). Modifications are cumulative. The SLM has 90M parameters and is trained on 6K hours of Libri-Light. Values are mean $\pm$ standard deviation across eight seeds of encoder training. For each variant, all scores are reported at the layer that maximizes average accuracy across the three metrics.}
\label{tab:ablation-slm}
\small
\renewcommand{\arraystretch}{1.12}
\begin{tabular}{l c c c c c}
\toprule
\textbf{Variant} & \textbf{Best layer} & \textbf{Average} & \textbf{sWuggy} & \textbf{sBLIMP} & \textbf{tSC} \\
\midrule
SpidR (reproduced) & 5 & 63.9 $\pm$ 0.6 & 66.4 $\pm$ 1.4 & 56.4 $\pm$ 1.0 & 68.9 $\pm$ 0.8 \\
+ shared heads & 5 & 58.3 $\pm$ 0.7 & 59.9 $\pm$ 1.4 & 51.6 $\pm$ 0.8 & 63.3 $\pm$ 1.3 \\
+ regularization & 7 & 66.7 $\pm$ 0.9 & 67.5 $\pm$ 2.3 & 60.4 $\pm$ 1.0 & 72.3 $\pm$ 0.9 \\
+ normalizations & 8 & 69.1 $\pm$ 0.7 & 70.1 $\pm$ 1.6 & 63.0 $\pm$ 1.2 & 74.1 $\pm$ 0.7 \\
+ $\mathrm{LR}=2.0\times10^{-4}$ (\textbf{SpidR 2.0}) & 7 & \textbf{70.1 $\pm$ 0.7} & \textbf{72.1 $\pm$ 1.2} & \textbf{64.0 $\pm$ 1.2} & \textbf{74.2 $\pm$ 1.3} \\
\bottomrule
\end{tabular}
\end{table}

\begin{figure}[t]
\centering
\includegraphics[width=0.95\textwidth]{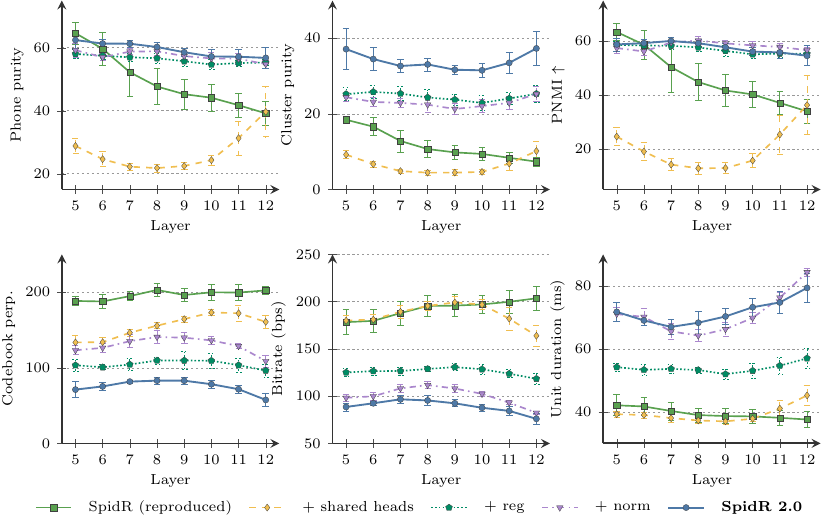}
\caption{Layerwise codebook metrics for the cumulative ablations from SpidR to SpidR 2.0, evaluated on ZeroSpeech'21 \texttt{dev-clean}. Panels show the same metrics as Figure~\ref{fig:codebook-layerwise}. Results are mean $\pm$ standard deviation across eight seeds.}
\label{fig:codebook-ablation-layerwise}
\end{figure}

The transition from SpidR to SpidR 2.0 consists of three successive modifications, followed by hyperparameter tuning.
Throughout this process, we guide encoder improvements using ABX as our primary proxy for downstream SLM performance.

\paragraph{Shared heads.}
As the main change resides in the tokenizer, we first replace the online codebooks on the teacher branch,
updated through iterative clustering~\citep{liu2023dinosr,poli2025spidr},
with the student's heads.
Without regularization, this change admits stable collapsed solutions.
This can be seen in Figure~\ref{fig:codebook-ablation-layerwise}, where all phone-code alignment metrics (i.e., phone/code purity and PNMI) suddenly drop.
However, continuous representations retain their performance on ABX and MAP (Tables~\ref{tab:ablation-abx} and~\ref{tab:ablation-map}, respectively),
whereas discrete units perform worse on downstream SLM tasks (Table~\ref{tab:ablation-slm}).

\paragraph{Regularization.}
We next introduce both regularization terms (i.e., Eq.~(\ref{eq:L_reg})) using the baseline weights $\lambda$ and $\mu$ and their associated schedules.
Adding both of them jointly is necessary, since they are both heavily co-dependent: changing $\lambda$ requires adjusting $\mu$ to keep similar codebook usage and prediction confidence.
Not only does this improve performance, already surpassing the SpidR baseline on most metrics, but it also reduces the standard deviation of ABX and MAP across training runs.
In addition, the resulting units tend to be longer and less diverse, with a codebook perplexity of 100 units and a mean duration of 55\,ms, yielding a lower bitrate.

\paragraph{Normalizations.}
We then add layer normalization before each linear projection layer inside the shared heads and normalize the projection weights to unit norm during the forward pass.
The projection weights are also initialized on the unit sphere.
These changes further increase unit duration while slightly increasing codebook diversity.
This yields better results in all metrics, with the largest improvements in MAP.
Interestingly, the standard deviation of MAP also decreases by a factor of five, while ABX variance seems unaffected.

\paragraph{Hyperparameter tuning.}
Finally, we perform a small sweep over optimization hyperparameters.
In our experiments, none of those yielded significantly beneficial changes but lowering the learning rate to $2.0 \times 10^{-4}$.
With this setting,
the prediction heads result in a more compact codebook
and improve performance on ABX proxy and the target SLM metrics,
at the cost of lower MAP.

\subsubsection{Regularization}
\label{app:sssec:regularization}

The objective of the entropy regularization terms (in Eq.~(\ref{eq:L_reg})) is to prevent the shared codebooks from collapsing towards two stable but uninformative equilibria.
The most classic case is codebook collapse: the student and teacher branches collapse onto a single arbitrary code regardless of the input.
To avoid this, we hypothesize that, over a batch of sequences, any informative sequence of discrete units uses a significant number of codes,
though the appropriate level of diversity is task-dependent.
Hence, we use the entropy of the batch-averaged probability distributions $H_c(\mathbb{E}_t[\mathbf{p}_{l,t}])$ (\emph{marginal entropy} or ME) as a proxy to be maximized.
Then, the heads in both student and teacher branches may converge to a near-uniform distribution over codes, regardless of the input.
As a differentiable surrogate, we minimize the mean predictive entropy $\mathbb{E}_t[H_c(\mathbf{p}_{l,t})]$ (\emph{conditional entropy} or CoE) to keep individual predictions confident.

\paragraph{Setup.}
In the default configuration, the marginal-entropy weight $\lambda$ is set to $0.1$ for the first 20k steps and then reduced to $0.001$,
while the conditional-entropy weight $\mu$ remains fixed at $0.1$.
We apply both terms to the teacher distribution only.
We initially choose this configuration using ABX, MAP, prediction confidence, and codebook usage as proxies.
We then conduct a sweep targeting development-set SLM performance and bitrate (defined in Appendix~\ref{app:sssec:codebook-metrics}).
We compare the default (reported in Table~\ref{tab:encoder-eval}) with four alternative configurations:
\begin{itemize}[leftmargin=1.5em,topsep=0pt]
    \item \textbf{Teacher-only regularization} (\texttt{A} and \texttt{B}).
    We give $\lambda$ and $\mu$ the same schedule: both remain fixed at $0.1$ in setup \texttt{A}, or both decrease from $0.1$ to $0.001$ after 20k steps in setup \texttt{B}. We also tested constant ME regularization with stepped CoE regularization, but omit this configuration because it collapsed.
    \item \textbf{Teacher and student regularization} (\texttt{C} and \texttt{D}). Setup \texttt{C} additionally applies regularization to the student, using the baseline weights and schedules for both branches. Because the SSL objective tends to produce teacher and student entropies of similar magnitude and with similar dynamics, this approximately doubles the regularization contribution to the total loss. Setup \texttt{D} therefore applies regularization to both branches with halved baseline weights.
\end{itemize}

\begin{table}[t]
\centering
\small
\renewcommand{\arraystretch}{1.12}
\setlength{\tabcolsep}{5pt}
\caption{Effect of entropy regularization on ABX and MAP (\%) for continuous representations from SpidR 2.0. All variants share the same base recipe as SpidR 2.0 and differ only in the regularization schedule/target. For each variant, ABX and MAP are reported at their respective best layers $l^*$, selected by task-averaged performance.}
\label{tab:reg-schedule}
\begin{tabular}{l c cccc c c cc c}
\toprule
& & \multicolumn{5}{c}{\textbf{ABX} ($\downarrow$)} & & \multicolumn{3}{c}{\textbf{MAP} ($\uparrow$)} \\
\cmidrule(l{0.2em}r{0.2em}){3-7}\cmidrule(l{0.2em}r{0.2em}){9-11}
& & \multicolumn{2}{c}{Within} & \multicolumn{2}{c}{Across} & & & & & \\
\cmidrule(l{0.2em}r{0.2em}){3-4}\cmidrule(l{0.2em}r{0.2em}){5-6}
\textbf{Variant} & $l^*$ & cln & oth & cln & oth & Avg. & $l^*$ & cln & oth & Avg. \\
\midrule
SpidR 2.0 (ME\,step, CoE\,constant) & 10 & 2.33 & 2.74 & 2.56 & 3.70 & \textbf{2.83} & 7 & 75.6 & 63.6 & \textbf{69.6} \\
\midrule
\texttt{A}: ME\,constant, CoE\,constant & 8 & 2.61 & 3.00 & 2.87 & 3.93 & 3.10 & 7 & 75.4 & 63.3 & 69.4 \\
\texttt{B}: ME\,step, CoE\,step & 10 & 2.57 & 3.18 & 2.79 & 4.10 & 3.16 & 7 & 72.1 & 60.0 & 66.0 \\
\midrule
\texttt{C}: Baseline $+$ student reg. ($1\times$) & 9 & 2.36 & 2.77 & 2.58 & 3.73 & 2.86 & 6 & 75.2 & 62.2 & 68.7 \\
\texttt{D}: Baseline $+$ student reg. ($0.5\times$) & 10 & 2.37 & 2.82 & 2.61 & 3.91 & 2.93 & 7 & 74.7 & 62.3 & 68.5 \\
\bottomrule
\end{tabular}
\end{table}

\begin{figure}[t]
\centering
\includegraphics[width=0.95\textwidth]{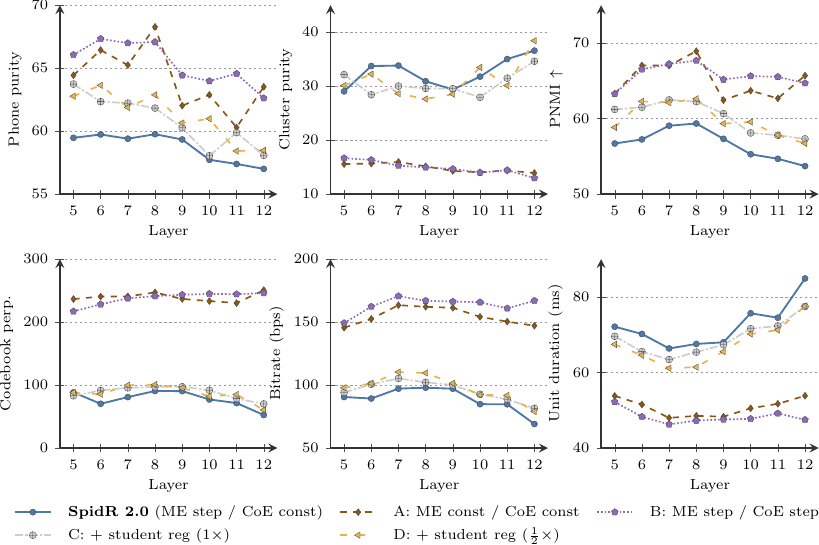}
\caption{Layerwise codebook metrics on ZeroSpeech'21 \texttt{dev-clean} for the regularization setups in Appendix~\ref{app:sssec:regularization}. Panels show the same metrics as Figure~\ref{fig:codebook-layerwise}. All runs use the SpidR 2.0 recipe and seed~0, differing only in regularization. The baseline reduces the ME weight $\lambda$ from $0.1$ to $0.001$ after 20k steps and keeps the CoE weight $\mu=0.1$. \texttt{A} keeps both weights at $0.1$; \texttt{B} reduces both to $0.001$. \texttt{C} and \texttt{D} regularize both teacher and student using full and half baseline weights, respectively. Purity and PNMI are in \%.}
\label{fig:codebook-reg-layerwise}
\end{figure}

\paragraph{Results.}
Table~\ref{tab:reg-schedule} compares the above regularization configurations using ABX and MAP on continuous representations.
We first notice that setup \texttt{B} suffers from a strong decrease in performance compared to the baseline.
Though it reaches values of ABX and MAP in the same orders of magnitude as the baseline,
setup \texttt{A} yielded larger codebook usage, as depicted in Figure \ref{fig:codebook-reg-layerwise}. 
At layer~7, it uses all 256 active codes (vs.\ 128 for SpidR 2.0).
Its units are also shorter on average (48\,ms vs.\ 66\,ms),
resulting in a higher bitrate (163\,bps vs.\ 97\,bps).

Setups \texttt{C} and \texttt{D} show small differences to baseline, around the standard deviation variations.
Note that, because these experiments are limited to a single seed due to time constraints,
we cannot establish whether these differences are robust across training runs.

Importantly, these results also inform the broader design choices in GSLM2.
Our regularization settings prioritize MAP, bitrate, and unit duration as proxies for downstream SLM performance (see Appendix~\ref{app:sssec:proxy-target-correlations}).
Other applications may favor shorter units and greater codebook usage, which can be encouraged by adjusting the regularization weights.
Setup \texttt{A} illustrates this trade-off: it achieves nearly maximal codebook usage and unit durations of 45-55\,ms,
with only a small reduction in MAP.

\subsubsection{Training Curves}
\label{app:sssec:training-curves}

\begin{figure}[t]
\centering
\includegraphics[width=0.9\textwidth]{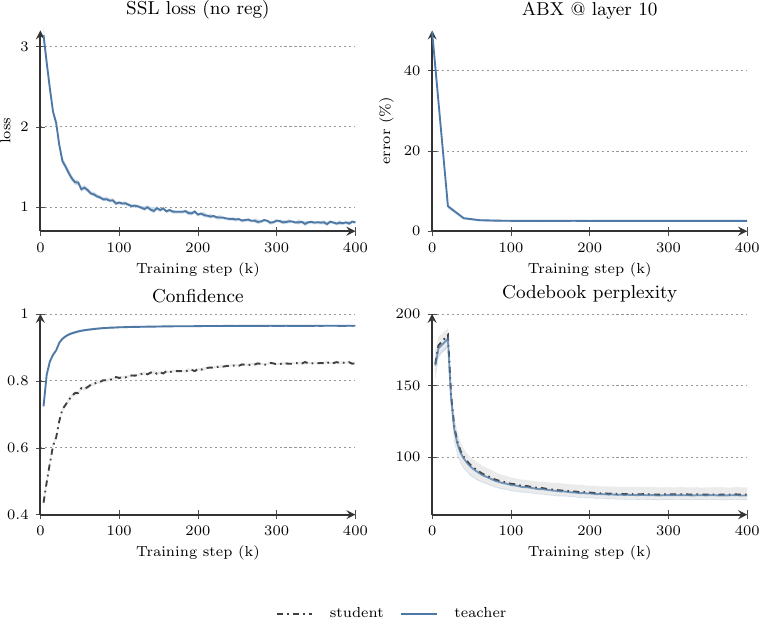}
\caption{Training dynamics of SpidR 2.0. We report the mean (line), as well as the minimum and maximum for each training step (shaded area) across eight seeds. \emph{Top row:} SSL loss and task-averaged ABX error at layer~10. \emph{Bottom row:} predictive confidence, $\mathbb{E}_{l,t}[\max_c \mathbf{p}_{l,t}[c]]$, and codebook perplexity, $\mathbb{E}_l[2^{H_c(\mathbb{E}_t[\mathbf{p}_{l,t}])}]$, for the teacher (solid) and student (dash-dotted).}
\label{fig:training-curves}
\end{figure}

To facilitate reproducibility, Figure~\ref{fig:training-curves} shows the SpidR 2.0 training curves aggregated over eight seeds.
The SSL loss (top left) decreases smoothly throughout training.
ABX at layer~10 (the best-performing ABX layer at the final step) decreases exponentially through the first half of training, and plateaus at around 200k steps.
In our experiments, while ABX and other proxy metrics plateau at this point,
downstream SLM performance continues to improve up to 400k steps.

The confidence trajectories (bottom left) show an interesting phenomenon:
the student and teacher branches sharpen their assignment distributions at different rates. 
While the teacher confidence rises quickly to ${\sim}$0.96, an almost certain confidence, the student confidence climbs gradually to ${\sim}$0.85.
The persistent gap of about 0.11 reflects the gap in available context between the teacher and the student,
where, under the masking policy, the student sees only 50\% of the frames in each training sequence.

Codebook perplexity (bottom right) makes the ME regularization schedule visible:
it grows for 20k steps when $\lambda = 0.1$, then drops sharply when $\lambda$ is reduced.
Perplexity subsequently declines more gradually, reaching approximately 73 by the end of training,
which gives a total of 129 active codes on the development sets (cf.\ Figure~\ref{fig:codebook-layerwise}).
The continued decline in perplexity during the second half of training indicates that the model progressively commits to a compact unit set after the diversity-enforcing pressure is relaxed, while downstream SLM performance continues to improve.

\subsubsection{DiscoPhon}
\label{app:sssec:discophon}

We report zero-shot results for SpidR 2.0 on DiscoPhon~\citep{poli2026discophon},
a multilingual benchmark evaluating unsupervised phoneme discovery across 12 languages.
Its tasks assess complementary properties of speech representation:
\begin{itemize}[leftmargin=1.5em,topsep=0pt]
    \item \emph{Unit quality}, measured by PNMI, captures how much information discrete units retain about reference phone identities.
    \item \emph{Phone recognition}, measured by phone error rate (PER), evaluates recovery of the phone sequence through substitution, deletion, and insertion errors, independently of precise boundary timing.
    \item \emph{Phone segmentation} evaluates boundary placement: F1 summarizes boundary precision and recall, while R-value penalizes over-segmentation more strongly.
    \item \emph{ABX} measures how well continuous representations distinguish phonemes. We average within- and across-speaker ABX errors.
\end{itemize}

For recognition and segmentation, we use a many-to-one mapping that assigns each codeword to the reference phone with which it most frequently co-occurs,
allowing multiple codewords to represent the same phone.
We evaluate the same encoders reported in Table~\ref{tab:encoder-eval},
all trained on LibriSpeech.
We also train SpidR and SpidR 2.0 on a VoxPopuli subset comprising 300 hours of speech from each of 19 languages (VP19),
extracted using the VoxPopuli procedure described in Appendix~\ref{app:ssec:datasets_and_preprocessing}.

\begin{table}[t]
    \centering
    \caption{DiscoPhon cross-lingual evaluation. All metrics are reported at each model's best-PNMI layer, selected by averaging PNMI across dev and test languages. Scores are averaged over languages within each split and expressed in \%. ABX is computed on continuous triphone representations and averaged over within-/across-speaker conditions. Dev languages: German, Swahili, Tamil, Thai, Turkish, Ukrainian. Test languages: Basque, English, French, Japanese, Mandarin, Wolof.}
    \label{tab:discophon}
    \small
    \renewcommand{\arraystretch}{1.12}
    \begin{tabular}{l c ccccc}
        \toprule
        \textbf{Model} & \textbf{Layer} & \textbf{F1} ($\uparrow$) & \textbf{R-val} ($\uparrow$) & \textbf{PER} ($\downarrow$) & \textbf{ABX} ($\downarrow$) & \textbf{PNMI} ($\uparrow$) \\
        \midrule
        \multicolumn{7}{l}{\textit{Dev languages}} \\
        \midrule
        DinoSR (LibriSpeech) & 5 & 61.1 & 48.9 & 82.4 & 10.8 & 49.0 \\
        SpidR (LibriSpeech) & 5 & 62.5 & 50.4 & 79.1 & 11.8 & 52.8 \\
        SpidR (VP19) & 5 & 63.0 & 44.8 & 82.8 & \textbf{8.7} & \textbf{55.5} \\
        \textbf{SpidR 2.0} (LibriSpeech) & 6 & 58.8 & 61.6 & 65.2 & 9.1 & 48.6 \\
        \textbf{SpidR 2.0} (VP19) & 6 & \textbf{64.3} & \textbf{65.3} & \textbf{61.1} & 8.9 & 50.0 \\
        \midrule
        \multicolumn{7}{l}{\textit{Test languages}} \\
        \midrule
        DinoSR (LibriSpeech) & 5 & 61.4 & 45.6 & 83.4 & 8.4 & 51.3 \\
        SpidR (LibriSpeech) & 5 & \textbf{64.2} & 45.2 & 80.8 & 9.5 & 56.2 \\
        SpidR (VP19) & 5 & 63.4 & 36.9 & 88.6 & \textbf{6.4} & \textbf{58.1} \\
        \textbf{SpidR 2.0} (LibriSpeech) & 6 & 59.6 & 58.6 & 66.7 & 7.1 & 51.3 \\
        \textbf{SpidR 2.0} (VP19) & 6 & 63.7 & \textbf{60.5} & \textbf{65.4} & 6.8 & 51.6 \\
        \bottomrule
    \end{tabular}%
\end{table}

Table~\ref{tab:discophon} reports the DiscoPhon results.
Consistent with the results in Figure \ref{fig:codebook-layerwise},
SpidR 2.0 yields lower PNMI than SpidR across the dev and test language splits for both pretraining datasets.
In addition, it is important to note that we are not reporting the results coming from the best ABX layer, but rather best PNMI, as it is our closest discrete phoneme mapping similarity evaluation metric.
This choice has little effect on SpidR's ABX scores, as its phonetic information is concentrated in layers~5-6.
For SpidR 2.0 trained on VP19, however, ABX errors decrease to 7.6\% and 5.8\% on dev and test languages at layer~10,
compared with 8.9\% and 6.8\% at the reported layer~6.

We then notice that SpidR 2.0 substantially improves phone recognition over SpidR,
even when pretrained exclusively on English speech from LibriSpeech.
Multilingual pretraining on VP19 further reduces its PER.
The segmentation results depend on the metric.
Relative to SpidR, SpidR 2.0 obtains lower F1 scores with LibriSpeech pretraining and slightly higher scores with VP19 pretraining.
R-value, however, improves consistently, by 11.2-13.4 absolute points with LS pretraining and 20.5-23.6 points with VP19 pretraining.
Since R-value penalizes over-segmentation more strongly than F1,
these gains suggest reduced over-segmentation.
It also agrees with the longer units produced by SpidR 2.0,
whose durations more closely match typical phone durations (see Appendix~\ref{app:sssec:codebook-metrics}).

\subsubsection{SUPERB}
\label{app:sssec:superb}

\begin{table}[t]
\centering
\caption{Downstream evaluation on SUPERB. For DinoSR, SpidR, and SpidR 2.0, we evaluate five learning rates for each trained downstream probe and report the test score of the run with the best development-set performance. For WavLM, we report the scores from the official SUPERB leaderboard. ASR reports word error rate without LM decoding; PR reports phone error rate. KS, IC, and ER report accuracy; SF reports slot-type F1 and slot-value character error rate (CER). QbE reports maximum term weighted value, scaled by 100.}
\label{tab:superb}
\small
\renewcommand{\arraystretch}{1.12}
\setlength{\tabcolsep}{4pt}

\begin{tabular}{l cc cc ccc c}
\toprule
& \multicolumn{2}{c}{\textbf{Recognition}} & \multicolumn{2}{c}{\textbf{Detection}} & \multicolumn{3}{c}{\textbf{Semantics}} & \textbf{Paraling.} \\
\cmidrule(l{0.2em}r{0.2em}){2-3}
\cmidrule(l{0.2em}r{0.2em}){4-5}
\cmidrule(l{0.2em}r{0.2em}){6-8}
\cmidrule(l{0.2em}r{0.2em}){9-9}
\textbf{Model} & ASR ($\downarrow$) & PR ($\downarrow$) & KS ($\uparrow$) & QbE ($\uparrow$) & IC ($\uparrow$) & SF-F1 ($\uparrow$) & SF-CER ($\downarrow$) & ER ($\uparrow$) \\
\midrule
DinoSR & \textbf{4.78} & \textbf{3.66} & 96.17 & 4.99 & \textbf{98.63} & 88.85 & 23.66 & \textbf{66.15} \\
SpidR  & 7.23 & 5.17 & 95.39 & 8.02 & 95.28 & 86.92 & 27.54 & 65.52 \\
SpidR 2.0 & 6.61 & 4.34 & 96.27 & 7.68 & 97.94 & 86.29 & 27.29 & 64.03 \\
WavLM (\textsc{Base}) & 6.21 & 4.84 & \textbf{96.79} & \textbf{8.70} & \textbf{98.63} & \textbf{89.38} & \textbf{22.86} & 65.94 \\
\midrule
WavLM \textsc{Large} & 3.44 & 3.06 & 97.86 & 8.86 & 99.31 & 92.21 & 18.36 & 70.62 \\
SpidR 2.0 \textsc{Large} & 4.71 & 2.72 & 97.05 & 8.84 & 98.68 & 88.90 & 22.55 & 65.28 \\
\bottomrule
\end{tabular}%

\end{table}

To complement our zero-shot encoder evaluations in Table~\ref{tab:encoder-eval},
we report downstream results on SUPERB~\citep{yang2021superb}, excluding speaker-related tasks.
The evaluation covers ASR, phone recognition (PR), keyword spotting (KS), spoken-query detection (QbE), intent classification (IC), slot filling (SF), and emotion recognition (ER).
For each supervised probe, we train a task-specific predictor and learn layer-mixing weights while keeping the encoder frozen.

Table~\ref{tab:superb} summarizes the results.
Among DinoSR, SpidR, and SpidR 2.0,
DinoSR achieves the best recognition, semantic, and ER scores,
while SpidR and SpidR 2.0 lead on QbE and KS, respectively.
This pattern is consistent with the divergence between spoken language modeling and supervised downstream performance reported in~\citet{poli2025spidr}.
Nevertheless, SpidR 2.0 outperforms SpidR on ASR and PR while also improving SLM performance.
The reversal in rankings between DiscoPhon's zero-shot evaluation and SUPERB's supervised probing is particularly evident for PR.
On the English subset of DiscoPhon, DinoSR obtains 45.0\% PER,
compared with 35.6\% for SpidR and 35.0\% for SpidR 2.0,
with all three models pretrained exclusively on LibriSpeech.
On SUPERB, DinoSR instead achieves the lowest PER of these models at 3.66\%,
compared with 5.17\% and 4.34\% for SpidR and SpidR 2.0, respectively.

These results suggest that the quality of direct code-to-phone assignments does not necessarily predict performance with a supervised probe.
However, differences in data and evaluation protocols prevent this comparison from isolating the cause of the reversal.
One possible explanation is that SpidR's intermediate prediction heads shape the representation geometry around the code-prediction objective.
This specialization may make phonetic information readily accessible through the pretrained projections
while limiting the flexibility of the learned features for newly trained downstream probes.

Scaling SpidR 2.0 to the \textsc{Large} scale following~\cite{hsu2021hubert} improves all reported SUPERB metrics relative to its \textsc{Base} model.
WavLM remains stronger on most directly comparable tasks at both scales,
but SpidR 2.0 achieves lower PER:
4.34\% versus 4.84\% at \textsc{Base}, and 2.72\% versus 3.06\% at \textsc{Large}.
The latter is the lowest PER among the models reported in Table~\ref{tab:superb}.

\begin{figure}[t]
\centering
\includegraphics[width=0.9\textwidth]{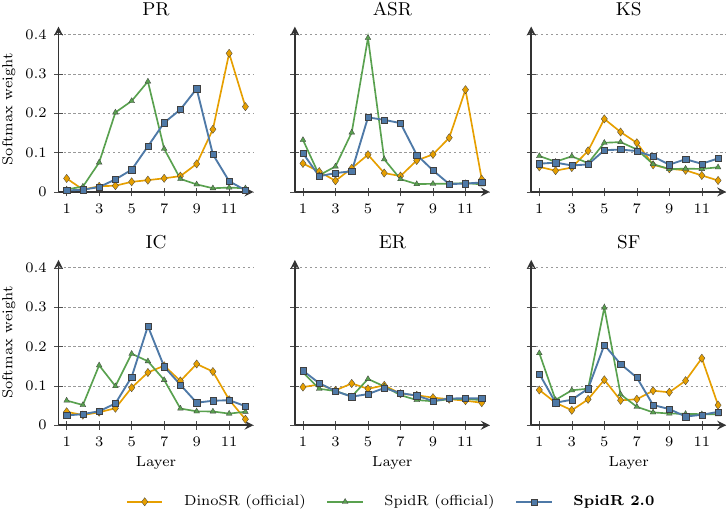}
\caption{Learned layer-mixing weights for the SUPERB probes. Softmax-normalized weights combine intermediate encoder representations into the features used by each downstream probe.}
\label{fig:superb-layer-weights}
\end{figure}

Figure \ref{fig:superb-layer-weights} illustrates the learned softmax weights used to combine encoder layers for the downstream probes.
For DinoSR, the mixtures concentrate on layer~11 for ASR, PR, and SF,
coinciding with its peak MAP scores.
SpidR places the greatest weight on layer~5,
with substantial weights also assigned to layers~4 and~6 for PR.
SpidR 2.0 exhibits a more even distribution across layers:
the largest weight exceeds 20\% on only two tasks (i.e., PR and IC).
This broader distribution motivates combining representations from multiple layers in GSLM2.

\subsubsection{Discrete Unit Quality}
\label{app:sssec:codes-vs-kmeans}

We compare discrete unit quality across encoders, quantization methods, and vocabulary sizes.
For SpidR 2.0, we apply $k$-means to the representations from every layer under two settings:
a fixed vocabulary of $k=256$, matching the total codebook capacity,
and a vocabulary size set separately for each layer to match the number of learned codes activated on the ZeroSpeech'21 \texttt{dev-clean} set.
Table~\ref{tab:codes-vs-kmeans} compares these configurations with the learned codes from SpidR 2.0 and with both $k$-means units and learned codes from SpidR.
We evaluate ABX computed on the discrete units and SLM performance with the 90M-parameter model trained on 6K-hour Libri-Light.
For each configuration, we select the layer that maximizes the mean accuracy across the development splits of sWuggy, sBLIMP, and tSC.

\begin{table}[t]
\centering
\caption{%
Discrete ABX and SLM results for SpidR and SpidR 2.0.
\emph{Codes} denotes units obtained from the prediction heads;
\emph{$k$-means} denotes clustering of continuous encoder representations.
For learned codes, $k$ denotes the number of active codes;
for $k$-means, it denotes the number of clusters.
$k$-means is fitted on LibriSpeech \texttt{train-clean-100},
except for LS-960, which uses the full 960-hour corpus.
Results are reported at the layer with the highest average development score across sWuggy, sBLIMP, and tSC.
ABX errors are averaged over \texttt{dev-clean} and \texttt{dev-other} and over within- and across-speaker conditions.
}
\label{tab:codes-vs-kmeans}
\small
\renewcommand{\arraystretch}{1.12}
\begin{tabular}{l c c c c c c c}
\toprule
\textbf{Model} & \textbf{Unit type} & \textbf{Layer} & $k$ & \textbf{ABX} ($\downarrow$) & \textbf{sWuggy} ($\uparrow$) & \textbf{sBLIMP} ($\uparrow$) & \textbf{tSC} ($\uparrow$) \\
\midrule
\multirow{2}{*}[0pt]{SpidR}
& $k$-means & 6 & 256 & 7.20 & 71.0 & 59.1 & 70.2 \\
& Codes   & 6 & 203 & 6.31 & 69.8 & 58.1 & 70.1 \\
\midrule
\multirow{4}{*}[-2pt]{SpidR 2.0}
& $k$-means & 8 & 256 & 6.63 & 72.5 & 61.6 & 73.6 \\
& $k$-means & 8 & 129 & 6.15 & 71.8 & 62.7 & 73.7 \\
& $k$-means (LS-960) & 7 & 128 & 6.75 & 72.4 & 62.6 & 74.1 \\
\cdashlinelr{2-8}
& \textbf{Codes} & 7 & 128 & \textbf{5.56} & \textbf{73.2} & \textbf{65.6} & \textbf{75.8} \\
\bottomrule
\end{tabular}%
\end{table}

With a fixed vocabulary of $k=256$,
SpidR 2.0 outperforms SpidR on discrete ABX and all three SLM metrics.
These gains extend to an external quantizer,
suggesting that the improvements arise partly from the encoder representations themselves.
Reducing $k$ to match the number of active learned codes lowers SpidR 2.0's discrete ABX error from 6.63\% to 6.15\%,
while leaving mean SLM performance nearly unchanged (69.2\% vs.\ 69.4\%).
The smaller vocabulary appears to improve sBLIMP but reduces sWuggy,
indicating different effects on syntactic and lexical evaluation.
Fitting $k$-means on the full LibriSpeech dataset (LS-960) achieves slightly higher mean SLM performance (69.7\% vs.\ 69.4\%) but higher ABX error,
compared to the default setup using \texttt{train-clean-100}.
However, ABX errors here must be put into context, as the layer with best SLM development scores is not necessarily the best layer for continuous ABX,
as shown in Figure~\ref{fig:layerwise}.

At layer~7 with 128 active units, SpidR 2.0's learned codes outperform the LS-960 $k$-means configuration on discrete ABX and all three SLM metrics,
despite using the same encoder representations, effective vocabulary size, and training corpus.
Thus, for SpidR 2.0, learning the quantization module during encoder pretraining both avoids a post-hoc clustering stage and yields better downstream performance.
We attribute this to the gradients of $\mathcal{L}_{\mathsf{ssl}}$ backpropagated through the prediction heads.
The codebooks' centroids $\mathbf{c}_{l,i}$ are then not only trained to discretize the latent space with coverage constraints as in the $k$-means algorithm,
but also to encourage code assignments to be predictable from masked context in the student branch.

\subsubsection{Stream Similarities}
\label{app:sssec:stream-similarities}

\begin{figure}[t]
\centering
\includegraphics[width=0.9\textwidth]{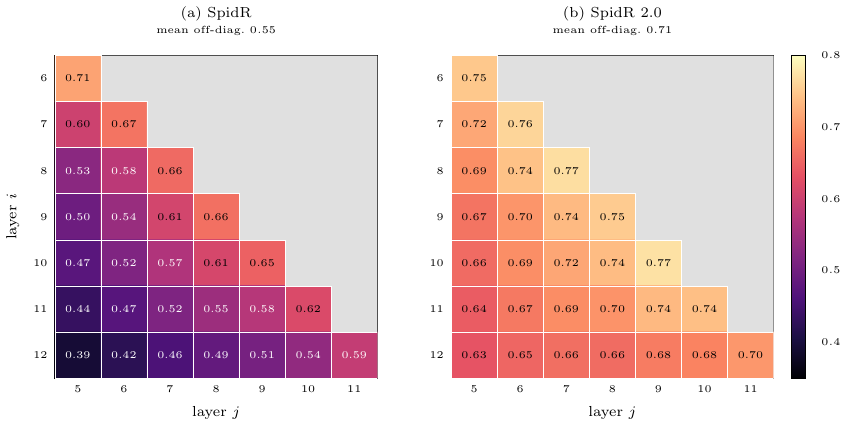}
\caption{Cross-layer unit-stream similarity. Pairwise NMI between unit streams extracted from encoder layers 5-12 of (a) SpidR and (b) SpidR 2.0, computed on LibriSpeech \texttt{dev-clean} and averaged over eight encoder training seeds. Since the matrices are symmetric, only entries below the diagonal are displayed; the diagonal (NMI $=1$) and upper triangle are greyed out.}
\label{fig:codebook-similarity}
\end{figure}

To help interpret the benefits of multi-stream SLM,
we examine the similarity between unit streams from different encoder layers.
Figure~\ref{fig:codebook-similarity} reports pairwise normalized mutual information (NMI) across layers 5-12 on LibriSpeech \texttt{dev-clean},
averaged over eight encoder-training seeds.
Mutual information is normalized by the arithmetic mean of the two marginal entropies.
Adjacent layers generally exhibit higher similarity than more distant layers,
and SpidR 2.0 has a higher mean off-diagonal NMI than SpidR (0.71 vs.\ 0.55).
NMI quantifies statistical overlap,
but does not reveal whether the differences between streams are useful for language modeling.
The layer-diversity ablation (Appendix~\ref{app:sssec:layer_diversity}) provides evidence of their practical value.
These findings support the value of combining representations from different layers despite substantial statistical overlap.

\subsubsection{Comparative Evaluation of Speech Encoders}
\label{app:ssec:literature-encoder-evals}

\begin{table}[t]
\centering
\caption{ABX and MAP (\%) for SpidR 2.0 and publicly available speech encoders on the ZeroSpeech'21 development set. Each metric is reported at the encoder layer $l$ that achieves its best average score. ABX is evaluated within and across speakers; \emph{cln}/\emph{oth} denote the clean/other subsets. Rows are ordered by decreasing average MAP within each group. In the first model group, bold and underlined scores indicate the best and second-best averages for each metric, respectively.}
\label{tab:encoder-benchmark}
{\scriptsize\setlength{\tabcolsep}{3.2pt}
\begin{tabular}{l r c cccc c c cc c}
\toprule
& & & \multicolumn{5}{c}{\textbf{ABX} ($\downarrow$)} & & \multicolumn{3}{c}{\textbf{MAP} ($\uparrow$)} \\
\cmidrule(l{0.3em}r{0.3em}){4-8} \cmidrule(l{0.3em}r{0.3em}){10-12}
& & & \multicolumn{2}{c}{Within} & \multicolumn{2}{c}{Across} & & & & \\
\cmidrule(lr){4-5} \cmidrule(lr){6-7}
\textbf{Model} & \textbf{Params} [M] & $l$ & cln & oth & cln & oth & Avg. & $l$ & cln & oth & Avg. \\
\midrule
\multicolumn{12}{l}{\textit{Self-supervised (masked / predictive)}} \\[1pt]
data2vec 2.0 Large~\citep{baevski2023efficient} & 313 & 2 & 6.20 & 5.92 & 6.92 & 7.75 & 6.70 & 12 & 76.1 & 72.6 & \textbf{74.4} \\
data2vec~\citep{baevski2022data2vec} & 93 & 4 & 4.41 & 5.51 & 5.12 & 7.40 & 5.61 & 10 & 72.5 & 66.9 & \underline{69.7} \\
\textbf{SpidR 2.0} & 93 & 10 & 2.33 & 2.74 & 2.56 & 3.70 & \textbf{2.83} & 7 & 75.6 & 63.6 & 69.6 \\
DinoSR~\citep{liu2023dinosr} & 93 & 5 & 4.05 & 5.11 & 4.74 & 7.30 & 5.30 & 11 & 74.9 & 64.1 & 69.5 \\
MR-HuBERT mono-large 50\,Hz~\citep{shi2024multi} & 320 & 1 & 3.41 & 3.73 & 3.93 & 5.11 & 4.04 & 1 & 73.0 & 64.2 & 68.6 \\
MR-HuBERT mono-base 50\,Hz~\citep{shi2024multi} & 97 & 1 & 3.63 & 3.99 & 4.21 & 5.40 & 4.31 & 1 & 70.7 & 63.1 & 66.9 \\
USAD~\citep{chang2025usad} & 97 & 10 & 3.55 & 4.19 & 4.08 & 5.77 & 4.40 & 9 & 72.5 & 61.2 & 66.8 \\
data2vec 2.0~\citep{baevski2023efficient} & 93 & 1 & 5.13 & 5.77 & 5.71 & 7.57 & 6.04 & 5 & 70.4 & 62.9 & 66.7 \\
MR-HuBERT mono-base 25\,Hz~\citep{shi2024multi} & 97 & 1 & 3.48 & 3.85 & 4.01 & 5.14 & 4.12 & 1 & 68.5 & 61.6 & 65.0 \\
HuBERT Large~\citep{hsu2021hubert} & 315 & 24 & 3.90 & 4.30 & 4.55 & 5.86 & 4.65 & 24 & 69.0 & 59.1 & 64.0 \\
SpidR~\citep{poli2025spidr} & 93 & 6 & 3.32 & 3.74 & 3.66 & 4.95 & \underline{3.92} & 5 & 69.5 & 57.3 & 63.4 \\
data2vec Large~\citep{baevski2022data2vec} & 313 & 7 & 4.51 & 5.46 & 5.24 & 7.18 & 5.60 & 10 & 64.2 & 56.3 & 60.2 \\
HuBERT XL~\citep{hsu2021hubert} & 962 & 48 & 4.04 & 4.38 & 4.80 & 6.16 & 4.84 & 48 & 64.6 & 54.6 & 59.6 \\
XEUS~\citep{chen2024towards} & 574 & 16 & 3.32 & 4.20 & 4.19 & 6.32 & 4.51 & 18 & 61.8 & 52.4 & 57.1 \\
MR-HuBERT mono-large 25\,Hz~\citep{shi2024multi} & 320 & 1 & 4.11 & 4.27 & 4.74 & 5.64 & 4.69 & 1 & 56.5 & 49.9 & 53.2 \\
HuBERT~\citep{hsu2021hubert} & 94 & 11 & 3.38 & 4.26 & 4.04 & 6.45 & 4.53 & 10 & 58.9 & 47.2 & 53.1 \\
wav2vec 2.0 Large~\citep{baevski2020wav2vec} & 315 & 16 & 4.34 & 5.14 & 5.22 & 7.27 & 5.49 & 13 & 52.8 & 42.1 & 47.4 \\
wav2vec 2.0~\citep{baevski2020wav2vec} & 94 & 6 & 4.51 & 5.67 & 5.27 & 7.83 & 5.82 & 7 & 49.8 & 38.6 & 44.2 \\
NEST-L~\citep{huang2025nest} & 115 & 10 & 5.16 & 5.70 & 5.77 & 7.31 & 5.98 & 10 & 43.7 & 39.1 & 41.4 \\
w2v-BERT 2.0~\citep{barrault2023seamless} & 580 & 17 & 5.52 & 7.37 & 8.64 & 14.21 & 8.94 & 11 & 16.0 & 8.9 & 12.5 \\
\midrule
\multicolumn{12}{l}{\textit{SSL $+$ finetuning robustness / speaker-invariance loss}} \\[1pt]
WavLM\,+\,Spin2048~\citep{chang2023spin} & 94 & 12 & 3.09 & 3.48 & 3.52 & 4.45 & 3.63 & 11 & 77.0 & 68.7 & 72.9 \\
WavLM Base+~\citep{chen2022wavlm} & 94 & 12 & 3.54 & 4.08 & 4.10 & 5.83 & 4.39 & 10 & 72.9 & 63.7 & 68.3 \\
WavLM Base~\citep{chen2022wavlm} & 94 & 11 & 3.01 & 3.70 & 3.47 & 5.23 & 3.85 & 10 & 73.6 & 62.1 & 67.9 \\
HuBERT\,+\,Spin2048~\citep{chang2023spin} & 94 & 12 & 2.69 & 3.23 & 3.05 & 4.07 & 3.26 & 11 & 70.8 & 62.8 & 66.8 \\
ContentVec100 ~\citep{qian2022contentvec} & 94 & 12 & 3.30 & 4.04 & 3.83 & 5.49 & 4.17 & 11 & 69.6 & 58.9 & 64.2 \\
WavLM Large~\citep{chen2022wavlm} & 315 & 24 & 3.94 & 4.32 & 4.65 & 5.98 & 4.72 & 24 & 67.6 & 57.5 & 62.5 \\
\midrule
\multicolumn{12}{l}{\textit{Supervised (ASR)}} \\[1pt]
wav2vec 2.0 Large (ASR)~\citep{baevski2020wav2vec} & 315 & 17 & 3.91 & 4.14 & 4.53 & 5.21 & 4.45 & 17 & 76.0 & 69.8 & 72.9 \\
HuBERT Large (ASR)~\citep{hsu2021hubert} & 315 & 14 & 4.69 & 5.38 & 5.48 & 7.25 & 5.70 & 22 & 75.6 & 70.1 & 72.8 \\
OWSM v3.1~\citep{peng2024owsm31} & 29 & 4 & 3.50 & 4.39 & 4.10 & 5.98 & 4.49 & 5 & 72.9 & 61.2 & 67.0 \\
wav2vec 2.0 (ASR)~\citep{baevski2020wav2vec} & 94 & 6 & 3.94 & 4.90 & 4.56 & 6.49 & 4.97 & 9 & 70.8 & 62.7 & 66.8 \\
SenseVoice~\citep{an2024funaudiollm} & 158 & 32 & 5.37 & 6.29 & 6.27 & 8.36 & 6.57 & 50 & 63.2 & 54.5 & 58.8 \\
Moonshine~\citep{jeffries2024moonshine} & 20 & 6 & 3.62 & 4.75 & 4.22 & 6.61 & 4.80 & 7 & 65.1 & 49.1 & 57.1 \\
FastConformer~\citep{rekesh2023fastconformer} & 109 & 9 & 2.92 & 3.63 & 3.21 & 4.68 & 3.61 & 11 & 52.9 & 49.9 & 51.4 \\
Canary~\citep{puvvada2024canary} & 609 & 10 & 3.46 & 4.05 & 3.91 & 5.04 & 4.12 & 11 & 50.3 & 47.8 & 49.1 \\
Whisper-large-v3~\citep{radford2023robust} & 637 & 26 & 6.35 & 6.99 & 7.51 & 9.98 & 7.71 & 27 & 40.6 & 31.6 & 36.1 \\
Whisper small.en~\citep{radford2023robust} & 88 & 8 & 7.01 & 8.28 & 8.28 & 11.75 & 8.83 & 11 & 31.1 & 24.3 & 27.7 \\
\midrule
\multicolumn{12}{l}{\textit{Speech-LLM audio encoders}} \\[1pt]
Kimi-Audio~\citep{kimiteam2025kimiaudio} & 637 & 26 & 6.13 & 6.81 & 7.37 & 9.71 & 7.50 & 32 & 56.8 & 50.8 & 53.8 \\
Audio Flamingo 3~\citep{ghosh2026audio} & 637 & 25 & 6.50 & 7.69 & 7.96 & 10.78 & 8.23 & 27 & 51.2 & 40.9 & 46.0 \\
Qwen2.5-Omni~\citep{xu2025qwen2} & 640 & 25 & 5.54 & 6.52 & 6.27 & 8.93 & 6.81 & 27 & 47.9 & 35.8 & 41.8 \\
Voxtral~\citep{liu2025voxtral} & 637 & 26 & 6.35 & 7.00 & 7.51 & 9.98 & 7.71 & 27 & 40.6 & 31.6 & 36.1 \\
\midrule
\multicolumn{12}{l}{\textit{Neural codecs \& speech tokenizers (pre-quantizer latent)}} \\[1pt]
Mimi~\citep{defossez2024moshi} & 39 & 1 & 4.78 & 5.91 & 5.67 & 8.47 & 6.21 & 1 & 36.5 & 26.9 & 31.7 \\
CosyVoice 2 ~\citep{du2024cosyvoice2} & 124 & 1 & 6.20 & 8.02 & 8.45 & 12.97 & 8.91 & 1 & 15.7 & 9.9 & 12.8 \\
\bottomrule
\end{tabular}}
\end{table}

Table~\ref{tab:encoder-benchmark} compares triphone ABX and word MAP scores across a broad range of speech encoders,
extending the results reported in Table~\ref{tab:encoder-eval}.
We group the models into five categories:
self-supervised encoders, self-supervised encoders with additional robustness objectives, ASR-supervised encoders, audio encoders from speech-capable LLMs, and neural codecs.
Because both metrics evaluate continuous intermediate representations,
we exclude quantizers, prediction heads, and decoders.
We report encoder parameter counts to contextualize the comparison and,
where available, compare our measurements with previously reported scores to identify potential evaluation discrepancies.

SpidR 2.0 achieves the lowest average ABX error among all evaluated encoders (2.83\%).
Four models exceed its word MAP score:
data2vec 2.0 Large, wav2vec 2.0 Large ASR, HuBERT Large ASR, and WavLM + Spin2048.
Its MAP score (69.6\%) is close to those of data2vec (69.7\%) and DinoSR (69.5\%);
we interpret these small differences cautiously given the observed variation across encoder training runs.
Larger models and ASR supervision can yield higher word MAP,
but their effects on ABX are mixed.

\subsubsection{Correlations Between Proxy and Target Metrics}
\label{app:sssec:proxy-target-correlations}

\begin{figure}[p]
\centering
\includegraphics[width=\textwidth]{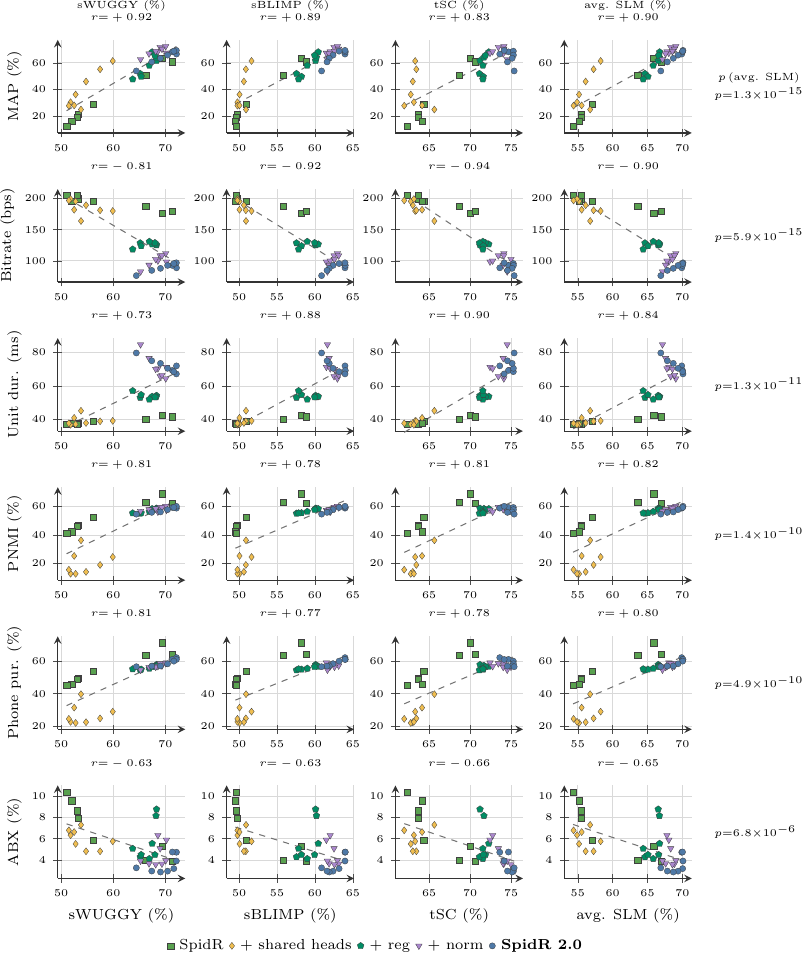}
\caption{Pearson correlations between proxy metrics (rows) and target SLM metrics (columns: sWuggy, sBLIMP, tSC, and their average, denoted avg.\ SLM) \emph{across} the progressive-ablation models from SpidR to SpidR 2.0 (Appendix~\ref{app:sssec:spidr2_ablation}). Each point represents a model-layer pair (layers 5-12), with scores averaged over eight encoder training seeds (i.e., 40 points per panel). Dashed lines show OLS fits, and each panel reports Pearson's $r$. Two-sided $p$-values for correlations with avg.\ SLM are shown to the right of each row, with rows ordered by increasing $p$-value (i.e., the most strongly correlated metric at the top).}
\label{fig:correlations-models}
\end{figure}

\begin{figure}[p]
\centering
\includegraphics[width=\textwidth]{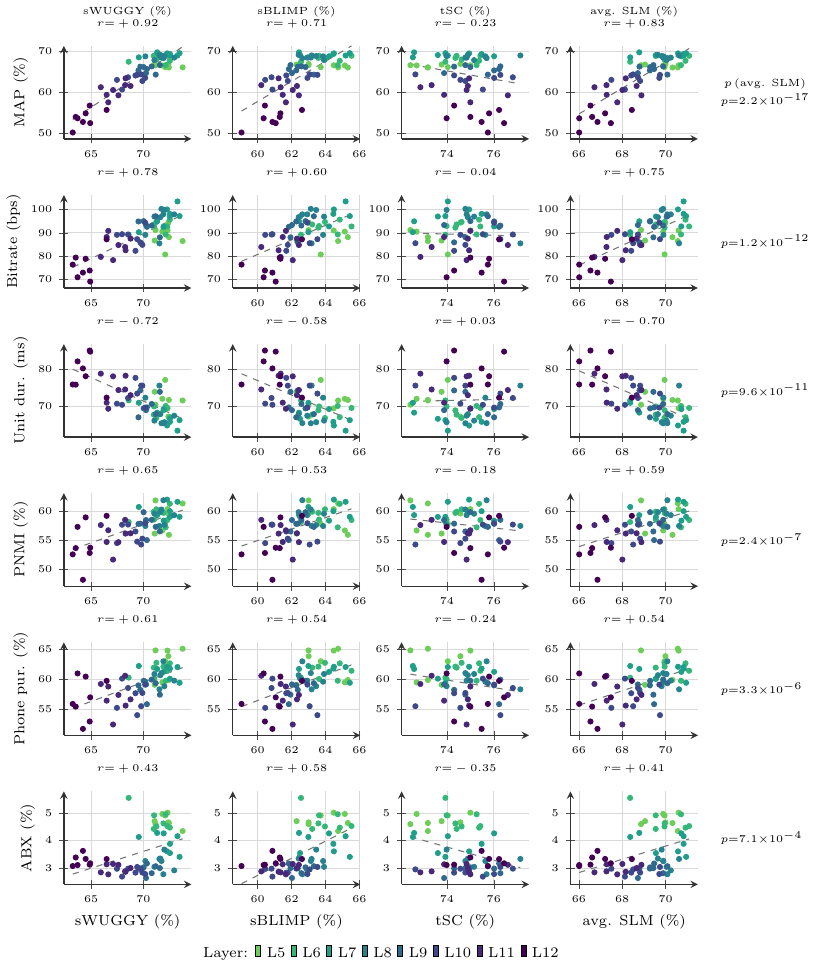}
\caption{Pearson correlations between proxy metrics (rows) and target SLM metrics (columns: sWuggy, sBLIMP, tSC, and their average, denoted avg.\ SLM) \emph{within} SpidR 2.0, across eight training seeds and eight encoder layers (5-12). Each point represents a layer-seed pair (i.e., 64 points per subplot). Dashed lines show OLS fits, and each panel reports Pearson's $r$. Two-sided $p$-values for correlations with avg.\ SLM are shown to the right of each row, with rows ordered by increasing $p$-value (i.e., the most strongly correlated metric at the top). The observed correlations primarily reflect variation across layers rather than across seeds.}
\label{fig:correlations-scatter}
\end{figure}

We report Pearson correlation coefficients and corresponding $p$-values between six proxy metrics (continuous ABX, MAP, phone purity, PNMI, unit duration, and bitrate) and three target SLM metrics (sWuggy, sBLIMP, and tSC) in two settings.
Figure \ref{fig:correlations-models} presents correlations across model variants from the progressive development of SpidR into SpidR 2.0,
as described in Appendix~\ref{app:sssec:spidr2_ablation}.
Each point represents a model-layer pair,
with layers ranging from 5 to 12 and scores averaged over eight encoder training seeds.
Figure \ref{fig:correlations-scatter} displays correlations across layers and seeds within SpidR 2.0.
For the latter analysis, we use a nominal sample size of 64.

Across model variants, MAP and bitrate show the strongest correlations with SLM performance,
whereas ABX has the highest $p$-values.
Within SpidR 2.0, tSC shows the weakest associations with the proxies,
with $R^2$ exceeding $0.10$ only for ABX.
The ranking of proxies by the $p$-values of their correlations with the average SLM score remains unchanged between the two settings.
The correlations for ABX, bitrate, and unit duration nevertheless reverse sign between settings.
Across model variants, longer units (lower bitrate) and lower ABX error rates are associated with better SLM performance.
Within SpidR 2.0, earlier layers generally have higher ABX error rates and produce shorter units,
yet achieve better sWuggy and sBLIMP scores.
This pattern results in positive correlations between averaged SLM metrics and both bitrate and ABX error rate.

Overall, our findings support reporting word MAP as a primary proxy rather than ABX~\citep{poli2025spidr} or PNMI~\citep{chang2025dc}.

\newpage

\subsection{Speech Resynthesis Evaluation}
\label{app:ssec:speech_resynthesis}

\begin{table}[t]
    \centering
    \caption{Speech resynthesis quality using oracle and predicted unit durations. Oracle durations are extracted from the ground-truth utterance, while predicted durations are estimated from the deduplicated unit sequence using the corresponding duration predictor.}
    \label{tab:speech_resynthesis}

    \small
    \renewcommand{\arraystretch}{1.12}
    \setlength{\tabcolsep}{6pt}

    \begin{tabular}{lcccc}
        \toprule
        & \multicolumn{2}{c}{\textbf{Oracle duration}}
        & \multicolumn{2}{c}{\textbf{Predicted duration}} \\
        \cmidrule(l{0.3em}r{0.3em}){2-3}
        \cmidrule(l{0.3em}r{0.3em}){4-5}
        \textbf{Model}
        & WER ($\downarrow$)
        & UTMOS ($\uparrow$)
        & WER ($\downarrow$)
        & UTMOS ($\uparrow$) \\
        \midrule
        Ground truth & 1.11\% & 4.15 & -- & -- \\
        \midrule
        SpidR     & 1.91\% & 2.99 & 2.08\% & 3.02 \\
        SpidR 2.0 & 3.35\% & 3.00 & 3.76\% & 3.08 \\
        GSLM2     & 2.00\% & 3.49 & 2.13\% & 3.51 \\
        \bottomrule
    \end{tabular}
\end{table}

We evaluate the resynthesis quality of the unit vocoder trained for each GSLM system.
Using the speech-continuation test data constructed from LibriSpeech (see Appendix~\ref{app:sssec:asr_ppl}),
we extract speech units from each utterance and resynthesize the speech with the corresponding unit HiFi-GAN vocoder.
We consider two duration conditions:
oracle durations extracted from the source utterance
and predicted durations generated from deduplicated speech units by the corresponding duration predictor.
We evaluate intelligibility using word error rate (WER), computed with Qwen3-Omni,
and naturalness with UTMOS~\citep{saeki2022utmos}.
Table~\ref{tab:speech_resynthesis} reports the results,
including performance on the unmodified test utterances as a ground-truth reference.

SpidR achieves lower WER than SpidR 2.0 under both oracle durations and predicted durations,
while the two models obtain similar UTMOS scores.
The increase in WER when replacing oracle durations with predicted durations is also smaller for SpidR than for SpidR 2.0 (0.17 vs.\ 0.41 percentage points),
suggesting that the durations of SpidR 2.0's coarser units are more difficult to predict.
Together with the ABX results from Table~\ref{tab:encoder-eval} (3.92\% for SpidR vs.\ 2.83\% for SpidR 2.0),
these findings suggest a trade-off between phonemic discriminability and resynthesis fidelity:
SpidR retains finer-grained acoustic detail that benefits resynthesis,
whereas SpidR 2.0 achieves better ABX performance but produces coarser units with fewer active codes (see Appendix~\ref{app:sssec:codebook-metrics}),
potentially discarding acoustic information needed for faithful resynthesis.
However, adjusting the regularization weights can produce shorter units and increase codebook usage (see Appendix~\ref{app:sssec:regularization}),
and we have not fully explored how these adjustments affect resynthesis quality.

GSLM2, by contrast, combines multiple unit levels in its vocoder.
It achieves substantially lower WER than SpidR 2.0 under both oracle durations and predicted durations,
while obtaining the highest UTMOS scores among the three systems.
Moreover, replacing oracle durations with predicted durations increases its WER by only 0.13 percentage points,
suggesting that cross-stream deduplication preserves sufficient temporal resolution for accurate duration prediction.
These results indicate that multi-level units help preserve fine-grained speech variation captured by the SpidR 2.0 encoder,
thereby improving resynthesis quality.
This interpretation is consistent with prior resynthesis findings~\citep{shi2024mmm,chang2025dc}.
In GSLM2, these improvements occur alongside gains in SLM performance.

\subsection{Statistical Significance Analysis}
\label{app:ssec:statistical_study}

\subsubsection{Confidence Intervals}
\label{app:sssec:confidence_intervals}

\begin{table}
    \centering

    \caption{Accuracy (\%) and corresponding 95\% bootstrap confidence intervals for the main results reported in Table~\ref{tab:main_results}.}
    \label{tab:confidence_intervals}

    \scriptsize
    \setlength{\tabcolsep}{3pt}
    \renewcommand{\arraystretch}{1.12}

    \newcommand{\accCI}[3]{%
      \makecell{#1\\[-1pt]{\tiny[#2, #3]}}%
    }

    \begin{tabular}{llccccccc}
    \toprule
    & & \multicolumn{2}{c}{\textbf{ZeroSpeech'21} ($\uparrow$)} & \multicolumn{2}{c}{\textbf{StoryCloze} ($\uparrow$)} & \multicolumn{3}{c}{\textbf{LongTail-Swap} ($\uparrow$)} \\
    \cmidrule(l{0.3em}r{0.3em}){3-4}
    \cmidrule(l{0.3em}r{0.3em}){5-6}
    \cmidrule(l{0.3em}r{0.3em}){7-9}
    \multicolumn{2}{l}{\textbf{Model}} & sWuggy & sBLIMP & tSC & sSC & WS & IS & AS \\
    \midrule

    \multirow{4}{*}[-12pt]{\rotatebox{90}{600\,h / 90M}}
    & Text LM & \accCI{73.4}{72.9}{73.8} & \accCI{61.7}{61.3}{62.0} & \accCI{75.4}{73.4}{77.3} & \accCI{53.4}{51.2}{55.7} & \accCI{73.5}{72.9}{74.0} & \accCI{80.6}{79.9}{81.3} & \accCI{62.2}{61.3}{63.2} \\
    \cmidrule(l{0.2em}r{0.2em}){2-9}
    & SpidR SLM & \accCI{67.4}{67.0}{67.7} & \accCI{54.2}{53.9}{54.6} & \accCI{65.4}{63.2}{67.7} & \accCI{52.4}{50.1}{54.6} & \accCI{54.3}{53.6}{55.0} & \accCI{57.9}{57.0}{59.1} & \accCI{54.2}{53.4}{55.3} \\
    & SpidR 2.0 SLM & \accCI{69.2}{68.9}{69.6} & \accCI{57.7}{57.4}{58.1} & \accCI{68.1}{66.0}{70.1} & \accCI{52.7}{50.4}{54.9} & \accCI{58.0}{57.3}{58.6} & \accCI{63.5}{62.8}{64.8} & \accCI{55.0}{54.1}{56.1} \\
    & GSLM2 & \accCI{71.5}{71.1}{71.8} & \accCI{59.8}{59.4}{60.1} & \accCI{74.1}{72.1}{76.1} & \accCI{52.0}{49.9}{54.4} & \accCI{62.4}{61.8}{63.1} & \accCI{68.2}{67.5}{69.4} & \accCI{59.1}{58.1}{60.1} \\
    \cmidrule(l{0.2em}r{0.2em}){1-9}

    \multirow{4}{*}[-12pt]{\rotatebox{90}{6K\,h / 300M}}
    & Text LM & \accCI{79.5}{79.1}{79.9} & \accCI{66.8}{66.4}{67.1} & \accCI{84.8}{83.2}{86.4} & \accCI{52.2}{49.9}{54.4} & \accCI{80.1}{79.7}{80.5} & \accCI{90.9}{90.7}{91.1} & \accCI{79.1}{78.8}{79.4} \\
    \cmidrule(l{0.2em}r{0.2em}){2-9}
    & SpidR SLM & \accCI{71.1}{70.8}{71.4} & \accCI{59.3}{58.9}{59.6} & \accCI{71.4}{69.3}{73.4} & \accCI{50.7}{48.4}{52.9} & \accCI{62.0}{61.5}{62.4}  & \accCI{63.0}{62.6}{63.4} & \accCI{58.5}{58.1}{58.8} \\
    & SpidR 2.0 SLM & \accCI{73.9}{73.6}{74.3} & \accCI{65.1}{64.7}{65.4} & \accCI{74.7}{72.6}{76.6} & \accCI{52.2}{50.0}{54.5} & \accCI{67.5}{67.0}{67.9} & \accCI{69.0}{68.7}{69.4} & \accCI{64.9}{64.5}{65.3} \\
    & GSLM2 & \accCI{76.3}{76.0}{76.6} & \accCI{67.2}{66.9}{67.5} & \accCI{78.9}{77.1}{80.8} & \accCI{52.5}{50.2}{54.8} & \accCI{75.6}{75.2}{76.1} & \accCI{75.5}{75.2}{75.9} & \accCI{69.8}{69.4}{70.2} \\
    \cmidrule(l{0.2em}r{0.2em}){1-9}

    \multirow{4}{*}[-12pt]{\rotatebox{90}{60K\,h / 1.2B}}
    & Text LM & \accCI{83.8}{83.5}{84.1} & \accCI{71.5}{71.2}{71.8} & \accCI{89.2}{87.8}{90.6} & \accCI{54.6}{52.3}{56.8} & \accCI{87.3}{87.0}{87.7} & \accCI{94.2}{94.0}{94.3} & \accCI{85.4}{85.2}{85.7} \\
    \cmidrule(l{0.2em}r{0.2em}){2-9}
    & SpidR SLM & \accCI{73.9}{73.6}{74.2} & \accCI{62.4}{62.1}{62.8} & \accCI{75.1}{73.1}{77.0} & \accCI{52.8}{50.7}{55.1} & \accCI{66.4}{66.0}{66.9} & \accCI{65.5}{65.2}{65.8} & \accCI{60.7}{60.3}{61.0} \\
    & SpidR 2.0 SLM & \accCI{76.7}{76.4}{77.0} & \accCI{69.6}{69.2}{69.9} & \accCI{81.0}{79.2}{82.8} & \accCI{56.2}{54.0}{58.4} & \accCI{74.8}{74.3}{75.2} & \accCI{72.5}{72.2}{72.8} & \accCI{66.6}{66.2}{66.9} \\
    & GSLM2 & \accCI{79.3}{79.0}{79.6} & \accCI{71.8}{71.5}{72.1} & \accCI{84.5}{83.0}{86.2} & \accCI{57.6}{55.2}{59.7} & \accCI{79.5}{79.1}{79.9} & \accCI{78.5}{78.3}{78.8} & \accCI{69.9}{69.6}{70.3} \\
    \bottomrule
    \end{tabular}
\end{table}

Table~\ref{tab:confidence_intervals} reports marginal 95\% bootstrap confidence intervals (CIs) for the main results in Table~\ref{tab:main_results}.
We use 10,000 item-level bootstrap resamples.
For sWuggy, we first average scores across the four voice realizations of each lexical item and then resample in-vocabulary and out-of-vocabulary items separately.
For sBLIMP, items are resampled within each linguistic subtype.
For sSC and tSC, story items are resampled.
For each LT-Swap task, we resample items separately within each frequency bin.
In each bootstrap draw, we compute bin-level accuracies and macro-average over frequency bins.
These intervals quantify uncertainty arising from the finite size of the evaluation set for a fixed checkpoint.
Appendix~\ref{app:sssec:training_seed_variance} further investigates training seed variance,
focusing on SpidR 2.0.

We perform pairwise significance tests among the three SLMs using matched evaluation items.
We use two-sided label-swap tests with $10^6$ random swaps for sWuggy and exact McNemar tests for sBLIMP, sSC, and tSC.
For LT-Swap, we use two-sided paired label-swap tests with $10^6$ random swaps,
computing accuracy within each frequency bin and macro-averaging equally across bins for each swap type.
For each model pair and training scale,
we apply Holm correction across the overall results for sWuggy, sBLIMP, sSC, and tSC and, separately, across the three LT-Swap types.
Across all training scales, 
the three SLMs differ significantly from one another on both sWuggy and sBLIMP (all $p\leq 3\times10^{-6}$).
GSLM2 also significantly outperforms both SpidR and SpidR 2.0 on tSC at every scale (all $p<10^{-5}$).
On sSC, no pairwise difference is significant at 600\,h or 6K\,h.
At 60K\,h, SpidR 2.0 and GSLM2 both significantly outperform SpidR ($p\leq0.0065$),
whereas they do not differ significantly from each other (56.2\% vs.\ 57.6\%; $p=0.20$).
On LT-Swap,
all observed improvements are statistically significant (all $p \le 3 \times 10^{-6}$),
except for the AS comparison between SpidR SLM and SpidR 2.0 SLM at 600\,h (54.2\% vs.\ 55.0\%),
which is not significant ($p=0.27$).

\subsubsection{Training Seed Variance}
\label{app:sssec:training_seed_variance}

The confidence intervals in Appendix~\ref{app:sssec:confidence_intervals} quantify uncertainty due to the finite evaluation set for a fixed checkpoint.
Here, we examine the complementary variation introduced by retraining the speech encoder and the SLM.
Training variability can affect model comparisons even when evaluation-set confidence intervals are narrow,
as highlighted in other work across deep learning fields~\citep{henderson2018deep,bouthillier2021accounting,dufour2026fid}.
We quantify this variability for SpidR 2.0 SLM on the ZeroSpeech'21 and StoryCloze benchmarks,
for which spoken language modeling studies typically report results from a single training run.

\paragraph{Experimental setup.}
We pretrain SpidR 2.0 encoders on LibriSpeech and train 90M-parameter single-stream SLMs on 6K hours of Libri-Light.
We conduct three experiments.
(1) We train eight encoders with seeds $\{0,\ldots,7\}$.
For each encoder, we train a separate SLM on the speech units extracted from each of layers~5--12, keeping the SLM seed fixed at 0.
(2) We use the encoder trained with seed 0 and vary the SLM seed over $\{0,\ldots,7\}$ for each of the same layers.
(3) We repeat SLM training eight times with its seed fixed at 0,
using speech units from layer 7 of the seed-0 encoder.
This final experiment measures residual variation between nominally identical training runs.

\paragraph{Variance estimates.}
For each metric, we first compute the sample variance across runs separately at each layer.
Let $s_{1,l}^2$ and $s_{2,l}^2$ denote this variance for Exps.~(1) and~(2), respectively,
at layer $l \in \{5,\ldots,12\}$.
We pool these within-layer variances across the eight layers:
\begin{equation}
\sigma_{\mathsf{enc,obs}}^2
= \frac{1}{8}\sum_{l=5}^{12}s_{1,l}^2,
\qquad
\sigma_{\mathsf{slm}}^2
= \frac{1}{8}\sum_{l=5}^{12}s_{2,l}^2.
\label{eq:layer_pooled_estimates}
\end{equation}
The resulting estimates summarize variability across runs within layers.

Fixing the SLM seed controls seeded random choices,
such as weight initialization and data ordering,
but does not necessarily guarantee full determinism during SLM training.
We estimate $\sigma_{\mathsf{res}}$ as the sample standard deviation across the eight fixed-seed repetitions in Exp.~(3),
obtaining a nonzero estimate for every reported metric ($\sigma_{\mathsf{res}} > 0$).
The observed variance $\sigma_{\mathsf{enc,obs}}^2$ in Exp.~(1) therefore includes both encoder variation and residual variation from SLM retraining.
Under an additive variance model,
we estimate the encoder component by subtracting the residual variance:
\begin{equation}
    \sigma_\mathsf{enc}^2 = \sigma_{\mathsf{enc,obs}}^2 - \sigma_\mathsf{res}^2.
\end{equation}
The SLM-training variance $\sigma_\mathsf{slm}^2$ observed in Exp.~(2) also includes both variation across SLM seeds and residual variation.
Retaining both contributions, we estimate the pooled standard deviation for independent encoder and SLM retraining as
\begin{equation}
    \label{eq:sigma_tot}
    \sigma_\mathsf{tot} = \left(
        \sigma_{\mathsf{enc}}^2 + \sigma_{\mathsf{slm}}^2
    \right)^{1/2},
\end{equation}
which counts residual variation once.
This decomposition assumes independent, additive encoder and SLM effects with negligible interactions.
Because residual variance is measured only at layer~7 of the seed-0 encoder,
applying the same correction to other layers and encoder checkpoints assumes that their residual variances are comparable.
For every reported metric, we observe that the estimated residual variance $\sigma_{\mathsf{res}}^2$ is smaller than the estimated SLM-training variance $\sigma_{\mathsf{slm}}^2$ by more than an order of magnitude.

\begin{table*}[t]
    \centering
    \renewcommand{\arraystretch}{1.08}

    \begin{minipage}{0.4\textwidth}
        \centering
        \caption{Estimated standard deviations for SpidR 2.0 SLM on ZeroSpeech'21 and StoryCloze. $\sigma_{\mathsf{tot}}$ (Eq.~(\ref{eq:sigma_tot})) estimates the pooled standard deviation when the encoder and SLM are independently retrained. For Average, the variance estimates are computed from the per-run mean of sWuggy, sBLIMP, and tSC accuracies, excluding sSC in gray. All values are in accuracy percentage points.}
        \label{tab:slm-noise-floor}
        {\small
        \begin{tabular}{l cccc}
        \toprule
        \textbf{Metric} & $\sigma_{\mathsf{slm}}$ & $\sigma_{\mathsf{enc}}$ & $\sigma_{\mathsf{res}}$ & $\sigma_{\mathsf{tot}}$ \\
        \midrule
        sWuggy & 0.16 & 0.99 & 0.02 & 1.00 \\
        sBLIMP & 0.38 & 0.97 & 0.09 & 1.04 \\
        tSC  & 0.60 & 1.05 & 0.15 & 1.21 \\
        \textcolor{black!50}{sSC} & \textcolor{black!50}{0.62} & \textcolor{black!50}{0.88} & \textcolor{black!50}{0.07} & \textcolor{black!50}{1.08} \\
        \cdashlinelr{1-5}
        Average & 0.27 & 0.75 & 0.03 & 0.79 \\
        \bottomrule
        \end{tabular}
        }
    \end{minipage}
    \hfill
    \begin{minipage}{0.57\textwidth}
    \centering
    \caption{Layer-pooled estimates of $\sigma_{\mathsf{enc}}$ across five training recipes, each evaluated with eight encoder seeds and one SLM run per encoder seed and layer. Recipe columns follow the cumulative changes from SpidR to SpidR 2.0, as studied in Appendix~\ref{app:sssec:spidr2_ablation}. Average excludes sSC (gray) and is defined as in Table~\ref{tab:slm-noise-floor}. All values are in accuracy percentage points.}
    \label{tab:slm-sigma-recipe}
    {\small\setlength{\tabcolsep}{4pt}
    \begin{tabular}{lccccc}
    \toprule
    \textbf{Metric} & SpidR & $+$heads & $+$reg. & $+$norm. & \textbf{SpidR 2.0} \\
    \midrule
    sWuggy & 2.50 & 1.69 & 2.53 & 1.49 & 0.99 \\
    sBLIMP & 1.23 & 0.59 & 1.36 & 1.16 & 0.97 \\
    tSC & 1.77 & 1.90 & 1.28 & 0.86 & 1.05 \\
    \textcolor{black!50}{sSC} & \textcolor{black!50}{1.08} & \textcolor{black!50}{1.05} & \textcolor{black!50}{0.91} & \textcolor{black!50}{0.73} & \textcolor{black!50}{0.88} \\
    \cdashlinelr{1-6}
    Average & 1.63 & 1.21 & 1.42 & 0.73 & 0.75 \\
    \bottomrule
    \end{tabular}
    }
    \end{minipage}
\end{table*}

\paragraph{Encoder pretraining accounts for most training variance.}
Table~\ref{tab:slm-noise-floor} shows that the estimated encoder component exceeds the SLM-training component for every reported metric.
Under the additive variance model,
the encoder share of total variance, $\sigma_{\mathsf{enc}}^2/\sigma_{\mathsf{tot}}^2$, 
ranges from approximately 67\% for sSC to 98\% for sWuggy.
These findings highlight the importance of including encoder retraining when estimating training variability in this setting.

\paragraph{Training variability depends on the model recipe.}
We estimate $\sigma_{\mathsf{enc}}$ for each configuration in the progressive ablation study in Appendix~\ref{app:sssec:spidr2_ablation},
using the same pooling procedure.
For each recipe, we subtract the residual variance estimated for SpidR 2.0,
assuming comparable residual variance across recipes.
Table~\ref{tab:slm-sigma-recipe} shows substantial differences in encoder variability across recipes, 
indicating that a single estimate of $\sigma_{\mathsf{enc}}$ may not generalize to another.
SpidR 2.0 has lower estimated standard deviations than SpidR,
particularly for sWuggy and the average score,
although the reductions are not monotonic across intermediate configurations.
The differences for sSC are smaller and should be interpreted in light of both models' near-chance performance in this setting,
as also observed for the 300M-parameter models in Table~\ref{tab:confidence_intervals}.

\textbf{Limitations.} This study attempts to give order-of-magnitude estimates of retraining variability rather than point out statistical weaknesses.
We acknowledge that, given resource constraints, conducting large-scale variance analyses might not be feasible.
We therefore limit each experimental setting to eight runs.
The additive model underlying $\sigma_{\mathsf{tot}}$ assumes independent encoder and SLM effects with negligible interactions.
The full variance decomposition is evaluated in a single setting:
SpidR 2.0 encoders pretrained on LibriSpeech and 90M-parameter SLMs trained on 6K-hour Libri-Light.
These estimates may not generalize to larger training datasets.
The closest related work~\citep{dufour2026fid} reports that their variance estimates are scale-invariant, though the setting differs substantially from ours.
We do not isolate separate sources of variance, as was done in that work.
In addition, this study only focuses on ZeroSpeech'21 and StoryCloze,
although we also report ABX and MAP spreads in Tables~\ref{tab:ablation-abx} and~\ref{tab:ablation-map}.

\end{document}